\documentclass{article}

\usepackage{arxiv}

\usepackage[utf8]{inputenc} % allow utf-8 input
\usepackage[T1]{fontenc}    % use 8-bit T1 fonts
\usepackage{hyperref}       % hyperlinks
\usepackage{url}            % simple URL typesetting
\usepackage{booktabs}       % professional-quality tables
\usepackage{array}
\usepackage{amsfonts}       % blackboard math symbols
\usepackage{nicefrac}       % compact symbols for 1/2, etc.
\usepackage{graphicx}
\usepackage{subcaption}% for subfigs
\usepackage{adjustbox}
\usepackage[numbers]{natbib}
\usepackage{doi}
\usepackage{amsmath}
\usepackage{amssymb}
\usepackage{algorithm}
\usepackage{algpseudocode}
\usepackage{xcolor}
\usepackage{fancyhdr}
\usepackage{amsthm}
\usepackage{cleveref}

\newtheoremstyle{remarkstyle}  % Name
  {5pt}                        % Space above
  {5pt}                        % Space below
  {}                           % Body font (upright)
  {}                           % Indent amount
  {\bfseries}                  % Theorem head font (bold)
  {.}                          % Punctuation after theorem head
  { }                          % Space after theorem head
  {}                           % Theorem head spec (empty = `normal`)
\newtheorem{theorem}{Theorem}[section]

\newtheorem{definition}[theorem]{Definition}
\theoremstyle{remarkstyle}
\newtheorem{remark}[theorem]{Remark}

\crefname{figure}{Figure}{Figures}

\title{A Resolution Independent Neural Operator
}

\author{Bahador Bahmani\thanks{Corresponding authors.}\\
	Hopkins Extreme Materials Institute\\
        Dept. of Civil and Systems Engineering\\
	Johns Hopkins University\\
	Baltimore, USA\\
	\texttt{bbahman2@jh.edu} \\
	\And
        Somdatta Goswami\\
        Dept. of Civil and Systems Engineering\\
	Johns Hopkins University\\
	Baltimore, USA\\
	\texttt{sgoswam4@jhu.edu} \\
	\And
        Ioannis G. Kevrekidis\\
        Dept. of Chemical and Biomolecular Engineering\\
        Dept. of Applied Mathematics and Statistics\\
	Johns Hopkins University\\
	Baltimore, USA\\
	\texttt{yannisk@jhu.edu} \\
	\And
        Michael D. Shields\textsuperscript{*}\\
        Dept. of Civil and Systems Engineering\\
	Johns Hopkins University\\
	Baltimore, USA\\
	\texttt{michael.shields@jhu.edu} \\
}

\hypersetup{
    colorlinks=true,      % Enable colored links
    linkcolor=green,      % Color for internal links
    urlcolor=blue,        % Color for URLs
    citecolor=blue        % Color for citations
}

\begin{document}
\maketitle

\begin{abstract}
    The Deep operator network (DeepONet) is a powerful yet simple neural operator architecture that utilizes two deep neural networks to learn mappings between infinite-dimensional function spaces. This architecture is highly flexible, allowing the evaluation of the solution field at any location within the desired domain. However, it imposes a strict constraint on the input space, requiring all input functions to be discretized at the same locations; this limits its practical applications. In this work, we introduce a general framework for operator learning from input-output data with arbitrary number and locations of sensors. This begins by introducing a resolution-independent DeepONet (RI-DeepONet), enabling it to handle input functions that are arbitrarily, but sufficiently finely, discretized. %across samples both during training and inference. 
    To this end, we propose two dictionary learning algorithms to adaptively learn a set of appropriate continuous basis functions, parameterized as implicit neural representations (INRs), from correlated signals defined on arbitrary point cloud data. %These bases establish a coordinate system to represent point-cloud data, allowing the coordinates to be directly used in DeepONet without any architectural changes.% To adress YK comment
    These basis functions are then used to project arbitrary input function data as a point cloud onto an embedding space (i.e., a vector space of finite dimensions) with dimensionality equal to the dictionary size, which can be directly used by DeepONet without any architectural changes. In particular, we utilize sinusoidal representation networks (SIRENs) as trainable INR basis functions. The introduced dictionary learning algorithms are then used in a similar way to learn an appropriate dictionary of basis functions for the output function data, which defines a new neural operator architecture referred to as the \textbf{R}esolution \textbf{I}ndependent \textbf{N}eural \textbf{O}perator (RINO). In the RINO, the operator learning task simplifies to learning a mapping from the coefficients of input basis functions to the coefficients of output basis functions.
    %This approach can be seen as an extension of POD-DeepONet for cases where the realizations of the output functions have different discretizations, making the Proper Orthogonal Decomposition (POD) approach inapplicable.
    We demonstrate the robustness and applicability of RINO in handling arbitrarily (but sufficiently richly) sampled input and output functions during both training and inference through several numerical examples.
\end{abstract}

% keywords can be removed

\section{Introduction}
\label{sec:intro}

\begin{figure}[h]
  \centering
    \includegraphics[width=0.9\textwidth]{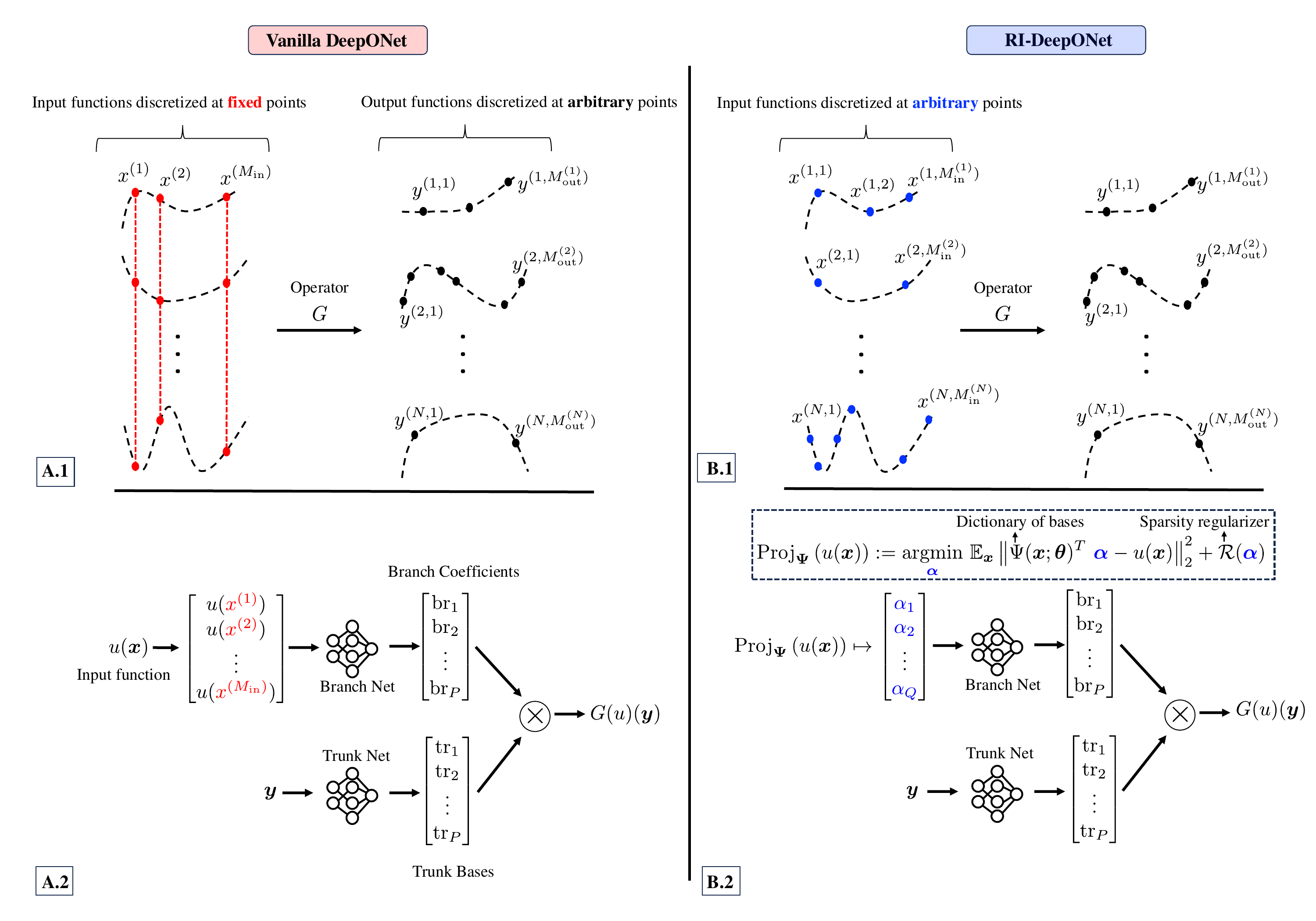}
  \caption{
  Vanilla DeepONet and RI-DeepONet in a nutshell. Vanilla DeepONet requires input functions to be sampled at a fixed set of points (\textbf{A.1}), while RI-DeepONet allows for arbitrary discretization of the input functions (\textbf{B.1}). The branch network in Vanilla DeepONet operates directly on the discretized input functions (\textbf{A.2}), whereas in RI-DeepONet, the DeepONet operates on the embeddings obtained from projecting the input signal onto a dictionary learned \textit{a priori} (\textbf{B.2}).
  }
  \label{fig:summary-methods}
\end{figure}

Partial Differential Equations (PDEs) serve as fundamental mathematical models for a vast array of phenomena in science and engineering. Solving a PDE amounts to computing the underlying solution operator, which maps given input functions such as initial and boundary conditions, source terms, coefficients, etc. to the solution. With the advent of modern machine learning models, new methodologies have emerged for creating fast data-driven emulators to solve parametric PDEs. 
Since operators are crucial in the context of PDEs, machine learning (ML) architectures that are designed to learn these operators and map functions to functions are increasingly viewed as a suitable paradigm for applying ML techniques to PDEs.

Modern operator learning frameworks employing deep neural networks (DNNs) have emerged as a powerful method for constructing emulators for PDEs that describe physical systems. Neural operators, first introduced in $2019$ with the deep operator network (DeepONet) \cite{lu2021learning}, utilize two DNNs to create a surrogate model, enabling fast inference and high generalization accuracy. In this architecture one DNN, known as the branch network, encodes the input functions at fixed sensor points (see Figure \ref{fig:summary-methods}), while another DNN, the trunk network,
%encodes the information related to the spatio-temporal coordinates of the output function. % to address YK comment
encodes information related to the output function at various spatiotemporal coordinates. 
%The solution operator is defined as the element-wise product of the embeddings of the branch network (coefficients) and the trunk network (basis functions) summed over the total number of output embeddings (see Figure~\ref{fig:OL-format}(a)). % to address YK comment
The solution operator is defined as a linear combination of basis functions learned by the trunk network, with their coefficients learned by the branch network.
The design of DeepONet is inspired by the universal approximation theorem for operators proposed by Chen \& Chen \cite{chen1995universal}. The generalized theorem for DeepONet essentially replaces shallow networks used for the branch and trunk net in \cite{chen1995universal} with DNNs to gain expressivity. Since its introduction, DeepONet has been employed to address challenging problems involving complex high-dimensional dynamical systems \cite{di2021deeponet,borrel2024sound,goswami2022physics,taccari2024developing,cao2024deep}. The framework is highly versatile, allowing the two DNNs to adopt various architectures such as convolutional neural networks (CNNs) or multilayer perceptrons (MLPs), depending on the specific problem requirements. 

%Despite its significant success, a notable limitation of the vanilla DeepONet framework is the need to discretize all input functions in the training and test sets at fixed sensor locations, which can be a major bottleneck in some applications -- either in cases where different computational meshes are used or a single computational mesh is adaptively refined in time or space-- for example when different samples in the training dataset correspond to solutions of a finite element model with different meshes.
Despite its significant success, a notable limitation of the vanilla DeepONet framework is the need to discretize all input functions in the training and test sets at fixed sensor locations. This requirement can be a major bottleneck in some applications; for example in cases where different computational meshes are used, multi-fidelity training data are provided, or where a single computational mesh is adaptively refined in time or space.
In this work, we develop efficient strategies for creating unique continuous representations of input and output functions without constraints on their discretization, such as the number and location of sensors for both the labeled training dataset and the testing samples that can be consistently parameterized and applied directly to existing DeepONet architectures (Resolution-Independent DeepONet, RI-DeepONet in Figure~\ref{fig:summary-methods}) or used to define a new neural operator in the embedding space (Figure~\ref{fig:func-space-mapping}). This results in the new \textbf{R}esolution \textbf{I}ndependent \textbf{N}eural \textbf{O}perator (RINO). Note, however, that although they do not require fixed sensor locations, these strategies of course require sufficient resolution to capture the characteristic length or time scales of the underlying function. For example, discretization must still satisfy the classical restriction on the minimal sampling rate due to the Nyquist–Shannon sampling theorem \cite{shannon1949communication}.

To achieve resolution independence, we develop a data-driven algorithm that maps an arbitrarily discretized input function realization to its corresponding embedding coordinates. We construct this embedding space by projecting these input function realizations onto a set of fixed, continuous, and fully differentiable basis functions shared across all realizations, where the projection coefficients serve as the embedding coordinates. This approach is motivated by and rooted in classical numerical analysis concepts, such as the Stone–Weierstrass theorem \citep{stone1948generalized}, providing interpretability for the learned representation in terms of the identified basis functions.
%Given the availability of the embedding coordinates, they can be utilized as the input to the branch network in RINO as shown in Figure~\ref{fig:summary-methods}.
For flexibility, these basis functions are parameterized by neural networks and are learnable, although they do not have to be and could, for example, be specified \textit{a priori}. Specifically, the proposed algorithm discovers appropriate neural network basis functions from a collection of correlated signals, i.e., input/output function realizations. This algorithm extends standard dictionary learning techniques from finite-dimensional vector spaces to function spaces. Dictionary learning is an unsupervised machine learning method that identifies a set of basis vectors to represent data as a sparse combination of these basis vectors. It has various applications, including sparse coding \citep{olshausen1997sparse}, image denoising \cite{mairal2007sparse}, and face recognition \cite{wright2008robust}. Unlike some classical ML methods such as principal component analysis (PCA) \cite{pearson1901liii,hotelling1933analysis} or Proper Orthogonal Decomposition (POD) \cite{berkooz1993proper,karhunen1946spektraltheorie}, dictionary learning can learn overcomplete basis sets, where the number of basis vectors exceeds the dimensionality of the input space. This may provide more flexibility for signal modeling and representation.

Here, the dictionary construction is performed in an \textit{offline} manner. At the construction stage, we iteratively add new trainable basis functions (i.e., atoms) and their corresponding coefficients (i.e., codes) to reduce the expected residual of the signal reconstruction error across all realizations. Similar to other standard dictionary learning algorithms, we additionally promote the learning towards per-realization sparse coefficients. Once the dictionary is learned, we apply it to unseen cases by reusing the dictionary and solving only for the basis coeficients of the queried realization \textit{online}. This is achieved by projecting the queried signal onto the learned dictionary atoms.

We leverage recent advances in \textit{implicit neural representations (INRs)} \citep{mescheder2019occupancy,sitzmann2020implicit} to model the dictionary basis functions. These representations are continuous, differentiable functions implicitly defined and parameterized by MLPs.
%Instead of parametrizing the signal of interest with discrete representations such as 2D grids, voxels, point clouds, and meshes, an MLP can be learned to continuously represent the signal of interest as an implicit function, $\Phi:\boldsymbol{x} \rightarrow \Phi(\boldsymbol x)$, mapping spatiotemporal coordinates $\boldsymbol{x} \in \mathbb{R}^M$ from an $M$-dimensional space to their corresponding $N$-dimensional value, $\Phi(\boldsymbol x) \in \mathbb{R}^N$.
%
Instead of the \textit{explicit} parameterization of the signal of interest, where the discrete signal values are the parameters (for example, signal values at 2D grids, voxels, point clouds, and meshes), INRs represent the signal values \textit{implicitly} (or indirectly) as a function parameterized by a neural network. This network $\Phi:\boldsymbol{x} \rightarrow \Phi(\boldsymbol x)$ maps spatiotemporal coordinates $\boldsymbol{x} \in \mathbb{R}^M$ from an $M$-dimensional space to their corresponding $N$-dimensional signal values, $\Phi(\boldsymbol{x}) \in \mathbb{R}^N$.
INRs find extensive applications in tasks such as image generation, super-resolution, object reconstruction, and modeling complex signals \cite{sitzmann2020implicit,sitzmann2019deepvoxels}. 
INRs offer several advantages. First, the functions are defined on the continuous domain of $\boldsymbol{x}$ rather than on a discrete grid, enabling the representation to adapt flexibly to various resolutions. Second, they ensure that the function is fully continuous and differentiable with respect to $\boldsymbol{x}$, facilitating the use of automatic differentiation, which is crucial for problems requiring access to the function’s gradient. Lastly, they have shown greater memory efficiency compared to grid-based representations \cite{mescheder2019occupancy}, as their capacity to model fine details depends on the architecture of the model rather than the grid resolution.
The application of INRs as basis functions in RINO is motivated by their resolution invariance, ability to capture fine details, and differentiability. In particular, we utilize sinusoidal representation networks (SIRENs) \cite{sitzmann2020implicit}.

\textbf{Connection with ReNO:}
The problem of interest in this study is to establish a neural operator learning method that can operate with input-output data sampled irregularly from one realization to another while maintaining relatively consistent accuracy.
% , particularly from the input function perspective. 
We assume that the function realizations are bandwidth-limited signals and that the density of sensor locations is sufficiently fine, exceeding the minimal Nyquist-Shannon sampling rate. Therefore, we do not explore potential aliasing issues related to sampling below the Nyquist-Shannon rate during training. Representation equivalent Neural Operators (ReNO) \cite{bartolucci2024representation} were developed to investigate aliasing issues in the context of operator learning and proposes a general formalism based on \textit{frame theory} to determine the conditions under which a discrete representation of an operator is equivalent to its continuous version. Since the proposed dictionary learning algorithm identifies basis functions that are (weakly) orthogonal, they may form a \textit{frame}; Bartolucci et al. \cite{bartolucci2024representation} show that orthogonality is a sufficient condition for defining a frame.
% see why orthogonality guarantees this in Equation 2.3 in the reference. 
The connection between discrete and continuous representations is established by these frames (basis functions) via the \textit{synthesis operator} (reconstruction operator) and its adjoint, known as the \textit{analysis operator} (projection operator). Therefore, the RINO framework aligns with the ReNO frame theory, but the proposed RINO further provides an algorithm to adaptively construct appropriate basis functions (frames) to enable consistent mapping between discrete and continuous representations of signals of interest.

\textbf{Other Relevant Work:} 
Perhaps the closest work to the proposed RINO is the recent work using neural fields for operator learning \citep{serrano2023operator}, but there are several fundamental differences. First, the neural fields approach relies on hypernetwork INRs \citep{dupont2022data}, which consist of a single large INR network augmented with additional modulation parameters. These modulation parameters are assumed to be realization-dependent and serve as embedding coordinates. As a result, the neural field embedding becomes a highly nonlinear function of the INR hypernetwork basis, making the method less interpretable than the proposed RINO. In contrast, we construct orthogonal basis functions sequentially, ensuring they can linearly reconstruct the input or output field. This approach is deeply motivated by and connected to classical ideas in numerical analysis for function approximation, such as the Stone–Weierstrass theorem \citep{stone1948generalized}, providing a principled and interpretable framework. Each basis in the RINO is parameterized by a simple vanilla INR, which further enhances modularity and interpretability. Additionally, the neural field embedding is more prone to non-uniqueness, as multiple distinct modulations can produce the same reconstruction. This increases the likelihood that the neural field method requires more data for effective training. Moreover, the neural field dependence on network parameters may result in high-dimensional embeddings, further escalating the data requirements for both training and achieving generalization. 
Recently, Ingebrand et al.~\citep{ingebrand2024zero} used neural network basis functions to encode signals of interest for reinforcement learning applications, but again their algorithm does not enforce orthogonality. As a result, it may lack the uniqueness, compactness, and interpretability of the embeddings. 
%Their functional encoders further use simple MLPs for basis parameterization, which are known to struggle with capturing high-frequency information, particularly when the input is low-dimensional. This limitation is especially pronounced in PDE-based problems, where the basis functions are typically supported on, at most, a 4-dimensional space-time domain.
%
Other works have explored adapting neural network architectures that are well-suited for handling unstructured data, such as transformers/attention mechanisms \citep{li2022transformer,li2024scalable,huang2024resolution}, graph neural networks \citep{morrison2024gfn}, and generative adversarial networks \citep{jiang2024resolution}. 

A related approach that has been applied in the recent literature is to learn neural operators on dimension-reduced latent spaces. 
Hesthaven et al.~\cite{hesthaven2018non} introduced POD-NN where POD is first employed to span the output function space using data snapshots, followed by the use of neural networks to predict the corresponding POD coefficients.
Bhattacharya et al.~\cite{bhattacharya2021model} proposed PCA-based operator learning, where PCA serves as an autoencoder for both input and output function spaces, and a neural network is used to interpolate within the resulting latent spaces. However, due to the reliance of these methods on PCA or POD in the initial step, these approaches may not handle input or output data that is unaligned.
Oommen et al.~\cite{oommen2022learning} introduced a framework for efficient operator learning in latent spaces by employing offline nonlinear dimension reduction through autoencoders and constructing a DeepONet on the resulting latent space, enabling operator learning for functions with sharp gradients. Building on this work, Kontolati et al.~\cite{kontolati2024learning} conducted an extensive study of latent operator networks and demonstrated that training on latent spaces (Latent DeepONet or L-DeepONet) significantly accelerates operator learning while improving computational efficiency. 
%Additionally, Zhang et al.~\cite{zhang2022multiauto} explored training multi-resolution high-dimensional data using latent spaces for efficient handling of varying resolutions.
%
%Although all of this research has been carried out in a two-step training process (training a reduced order model followed by training of a neural operator), \cite{wang2024latent_time, wang2024latent_forward_inverse} introduced the idea of training an operator in the latent space in a single training session employing a transformer block in the DeepONet output space. However, the computational efficiency, memory requirements, and convergence characteristics regarding the single training session need further validation through direct comparisons with the two-step training process. 
%
Meanwhile, Seidman et al.~\cite{seidman2022nomad} introduced NOMAD, which directly builds the neural operator on a nonlinear manifold of reduced dimension. While the proposed RINO is related to these works in the sense that it builds a neural operator on a latent space, its primary objective differs in an important way. Instead of prioritizing dimension reduction, which other latent neural operators do, the RINO prioritizes learning on a space of continuous and orthogonal basis functions (for both input and output functions). While the associated embedding in RINO typically reduces dimension (it does not have to), this continuous and orthogonal basis provides a unique representation for each input/output function that is resolution independent. This allows it to handle multi-fidelity data with arbitrary number and locations of sensor points.

The paper is organized as follows. In Section \ref{sec:form}, we describe the components of the proposed framework, namely the operator learning architecture, the concept of learning a dictionary of basis functions, and the parametrization approach via the INRs. In Section \ref{sec:exam}, we present six numerical examples to test the RINO framework. For each numerical example, we conducted a comprehensive study by varying the input function, the number of sensors, and their locations. Details about the data generation process are provided in each respective example. Finally, we summarize our observations and provide concluding remarks in Section \ref{sec:conclusion}. An important discussion on the connection between the proposed function-space dictionary learning algorithm and classical Gappy POD (GPOD), as well as alternative possibilities using random projections to achieve vector representations of input or output function data, is presented in Appendix~\ref{appx:gpod}.

% \begin{figure}[ht]
%   \centering
%   \begin{subfigure}[b]{0.49\textwidth}
%     \includegraphics[width=\textwidth]{figs/data_classic_deeponet.pdf}
%     \caption{vanilla DeepONet \cite{lu2021learning}}
%     %\label{subfig:one}
%   \end{subfigure}
%   \hfill % optional: add some separation between the subfigures
%   \begin{subfigure}[b]{0.49\textwidth}
%     \includegraphics[width=\textwidth]{figs/data_rino.pdf}
%     \caption{RINO}
%     %\label{subfig:two}
%   \end{subfigure}
%   \caption{Discretization of the input function space in the (a) vanilla DeepONet and (b) RINO. The number and location of input sensors can vary arbitrarily from one realization to another in both the training and testing datasets in RINO.}
%   \label{fig:data-format}
% \end{figure}

% \begin{figure}[ht]
%   \centering
%   \begin{subfigure}[b]{0.45\textwidth}
%     \includegraphics[width=\textwidth]{figs/deeponet_classic_3.pdf}
%     \caption{vanilla DeepONet \cite{lu2021learning}}
%     %\label{subfig:one}
%   \end{subfigure}
%   \hfill % optional: add some separation between the subfigures
%   \begin{subfigure}[b]{0.45\textwidth}
%     \includegraphics[width=\textwidth]{figs/rino_method.pdf}
%     \caption{RINO}
%     %\label{subfig:two}
%   \end{subfigure}
%   \caption{Architectures in (a) vanilla DeepONet and (b) RINO. The branch network in RINO accepts the codes (embeddings) obtained from the projection of the input signal onto a dictionary found a priori.}
%   \label{fig:OL-format}
% \end{figure}

\section{Formulation}
\label{sec:form}
In Section~\ref{sec:OL}, we revisit the operator learning problem statement that is the focus of this work. We begin by utilizing the DeepONet formalism to learn operators in a data-driven manner and develop a formulation that accepts a resolution-independent embedding of the input functions (e.g., spatially and/or temporally dependent source terms, initial conditions, material properties, etc.) instead of directly working with input function data in the original space. This Resolution-Independent DeepONet (RI-DeepONet) approach eliminates the resolution dependency of the vanilla DeepONet from the perspective of the input function.
We then present a more general resolution independent operator learning formulation, the RINO, that bypasses the need for an unknown trunk during the operator learning training and enables operator learning directly in the embedded function space. This formulation extends the POD-DeepONet approach into a fully resolution-independent framework.
In Section~\ref{sec:DL}, we introduce two algorithms to learn appropriate parametrized basis functions to represent input and/or output function data in a resolution-independent manner. Finally, in Section~\ref{sec:inr}, we specifically introduce the basis function parametrization as INRs.

\subsection{Operator Learning Between Function Spaces}
\label{sec:OL}
% In this section, we provide the details of the operator learning problem setup. Specifically, we reformulate the vanilla DeepONet to be learned via a resolution-independent embedding of the input function, rather than a discretized version of the input function.

% Our goal is to learn a function between two infinite-dimensional spaces, given a finite number of input-output pairs. 

Let $\mathcal{U}$ and $\mathcal{S}$ be Banach spaces of vector-valued functions, as follows:
\begin{align}
    &\mathcal{U} = \{ u: \mathcal{X} \to \mathbb{R}^{d_u}\}, \quad \mathcal{X}\subseteq \mathbb{R}^{d_x}\\
    &\mathcal{S} = \{ s: \mathcal{Y} \to \mathbb{R}^{d_s}\}, \quad \mathcal{Y}\subseteq \mathbb{R}^{d_y}.
\end{align}
where $\mathcal{U}$ and $\mathcal{S}$ denote the \textit{input functions} and \textit{output functions}, respectively for the operator learning problem. Assuming there is a ground-truth operator $\mathcal{G}: \mathcal{U} \to \mathcal{S}$, the operator learning task is to approximate $\mathcal{G}$ with a parameterized functional $\mathcal{F}_{\theta}: \mathcal{U}\times \Theta \mapsto \mathcal{S}$ where $\Theta \subseteq \mathbb{R}^{\text{dim}(\theta)}$ is the parameter space. The optimal parameters can be found, in the empirical risk minimization sense, as follows:
\begin{equation}
    \underset{\boldsymbol{\theta}}{\text{argmin}} \ \mathbb{E}_{u\sim \mu_u}
    \left\|
    \mathcal{G}(u) - \mathcal{F}_{\theta}(u)
    \right\|_{\mathcal{S}}^2
    \approx 
    \frac{1}{N} \sum_{i=1}^{N}
    \left\| s^{(i)} - \mathcal{F}_{\theta}(u^{(i)})
    \right\|_{\mathcal{S}}^2,
\end{equation}
where the dataset $\mathcal{D}_{\text{OL}} = \left\{(u^{(i)}, s^{(i)})\right\}_{i=1}^N$ contains $N$ pairs of input and output functions with $u^{(i)} \sim \mu_{u}$ are i.i.d (independent and identically distributed) samples from a probability measure $\mu_{u}$ supported on $\mathcal{U}$ and $s^{(i)} = \mathcal{G}(u^{(i)})$. 

In practice, we only have access to finite-dimensional observations of input-output functions, which correspond to their discretized versions. Thus, in the most general setting, the $i$-th input function $u^{(i)}(x)$ is arbitrarily discretized at $M_{\text{in}}^{(i)}$ input sensor locations and stored as the point cloud $\mathcal{D}^{(i)}_{u} = \left\{x^{(i, j)}, \bar{u}^{(i, j)}\right\}_{j=1}^{M^{(i)}_{\text{in}}}$, where $x^{(i, j)} \in \mathcal{X}$ and $\bar{u}^{(i, j)}(x^{(i, j)})\in \mathcal{U}$. Similarly, its corresponding output function $s^{(i)}(y)$ is arbitrarily discretized at $M^{(i)}_{\text{out}}$ output sensor locations and stored as a point cloud $\mathcal{D}^{(i)}_{s} = \left\{y^{(i, j)}, \bar{s}^{(i, j)}\right\}_{j=1}^{M^{(i)}_{\text{out}}}$, where $y^{(i, j)} \in \mathcal{Y}$ and $\bar{s}^{(i, j)}(y^{(i, j)})\in \mathcal{S}$. This general data configuration is schematically illustrated in Figure~\ref{fig:summary-methods}(B.1). 

\begin{remark}
In this manuscript, superscript indices are used to denote data-related information. Specifically, a single superscript ($\cdot^{(i)}$) indicates the $i$-th function realization, which can represent a sample. A double superscript ($\cdot^{(i,j)}$) is used to indicate the $j$-th observation point (or sensor) within the $i$-th function realization. For example, $u^{(i)}$ denotes the $i$-th realization of the function $u$, and $x^{(i,j)}$ denotes the $j$-th observation point in the $i$-th function realization. We use the bar notation $\bar{u}$ to represent the discretized form of the continuous field $u$, which may include measurement and/or discretization errors.
\end{remark}

Various approaches have been developed to approximate the ground truth operator $\mathcal{G}$ using parameterized functions, with neural networks being particularly notable. Two prominent methods are DeepONet and Fourier Neural Operator (FNO) introduced by \cite{lu2021learning} and \cite{li2020fourier}, respectively. The FNO leverages Fourier transforms to efficiently learn operators on grids, though it can be computationally intensive due to the need for these Fourier transforms. In contrast, DeepONet offers greater architectural flexibility, making it particularly suitable for operations on point cloud data. Within the DeepONet framework, the parameterized functional $\mathcal{F}_{\theta}(u)(y)$ has the following structure:
\begin{equation}
    \mathcal{G}\left(u\right)(y)
    \approx
    \mathcal{F}_{\theta}(u)(y)
    \mathrel{\stackrel{\text{DeepONet}}{=}}
    {\mathcal{F}_{\boldsymbol{\theta}}^{\text{br}}}^T(\bar{u})\ {\mathcal{F}_{\boldsymbol{\theta}}^{\text{tr}}}(y)
    =
    \sum_{k=1}^{P} \text{br}_k(\bar{u}; \boldsymbol{\theta}) \ \text{tr}_k(y; \boldsymbol{\theta}),
\end{equation}
where $\mathcal{F}^{\text{tr}}_{\boldsymbol{\theta}}$ and $\mathcal{F}^{\text{br}}_{\boldsymbol{\theta}}$ are parameterized functions modeled by suitable neural networks, referred to as the \textit{trunk} and \textit{branch} nets, respectively. The trunk net defines a set of basis functions over the domain of the output function, while the branch net defines the corresponding coefficients, which depend on the input function $u$. Within this formalism, a finite-dimensional vector representation (i.e., embedding) of the \textit{input function} $u$ is necessary for use in standard neural network architectures. Indeed, the observed data $\bar{u}$ can serve as an effective representation of the input function, as used in the original DeepONet. However, this requires that the input sensor locations remain consistent across all input data, as illustrated in Figure~\ref{fig:summary-methods}(A). This has two drawbacks: first, it makes the resulting approximate operator dependent on the resolution, and second, it may cause the number of parameters in the branch network to become unnecessarily large (and therefore expensive to train), depending on how finely the input functions are discretized.

One approach to mitigate this issue is to find a vector representation of the input data that is resolution-independent. If such a representation exists, we hypothesize that a Resolution-Independent DeepONet (RI-DeepONet) can be formulated in this embedding space as follows:
\begin{equation}
    \mathcal{G}\left(u\right)(y)
    \approx
    \sum_{k=1}^{P} \text{br}_k(\boldsymbol{\alpha}(u); \boldsymbol{\theta}) \ \text{tr}_k(y; \boldsymbol{\theta}),
    \label{eqn:RINO_representation}
\end{equation}
where $\boldsymbol{\alpha} \in \mathbb{R}^Q$ is an embedding of the input function $u(x)$ in a well-defined embedding space; see Figure~\ref{fig:summary-methods}(B.2) for a schematic of the architecture and Figure~\ref{fig:func-space-mapping}(a) for a schematic of the function space mappings of this RI-DeepONet. Note that the selection of the embedding can be defined arbitrarily by projecting the input onto a set of continuous orthogonal basis functions (e.g. orthogonal polynomials). However, we aim to identify an optimally compact basis that both improves accuracy and reduces the number of necessary parameters in the operator. In the next section, we will introduce a method to find this desired representation.

The resulting optimization statement for the operator learning task with this embedding, in the discrete sense, solved in this work is as follows:
\begin{equation}
    \underset{\boldsymbol{\theta}}{\text{argmin}}
    \
    \mathbb{E}_{i}
    \mathbb{E}_{(y, s)\sim\mathcal{D}_s^{(i)}}
    \left\| s - \mathcal{F}_{\theta}(\boldsymbol{\alpha}^{(i)})(y)
    \right\|_{2}^2.
\end{equation}

\begin{remark}
    In this paper, the symbol $\theta$ is used to represent neural network parameters. Its specific meaning may vary by section, and it denotes the parameters of different functions or models. Readers should refer to the context within each section for the precise definition.
\end{remark}

\subsubsection{Operator Learning Between Embedded Function Spaces (RINO)}
\label{sec:OL-BrOnly}

In the vanilla DeepONet, the trunk network essentially tries to find appropriate basis functions that approximate the output functions. So, if a suitable set of bases already exists for the output function data, it can be directly used. Using this intuition, Lu et al.~\cite{lu2022comprehensive} propose the POD-DeepONet, where the trunk network is replaced by fixed, predefined POD modes associated with the output functions, resulting in only the branch network being trainable. This approach may facilitate training and improve the accuracy of the operator learning task. However, due to the use of POD, which works on fixed grids, similar discretization dependence issues arise, as with the input function process in the vanilla DeepONet. For example, if the discretization of output functions varies from realization to realization, POD cannot be directly used.

\begin{figure}[h]
  \centering
    \includegraphics[width=0.8\textwidth]{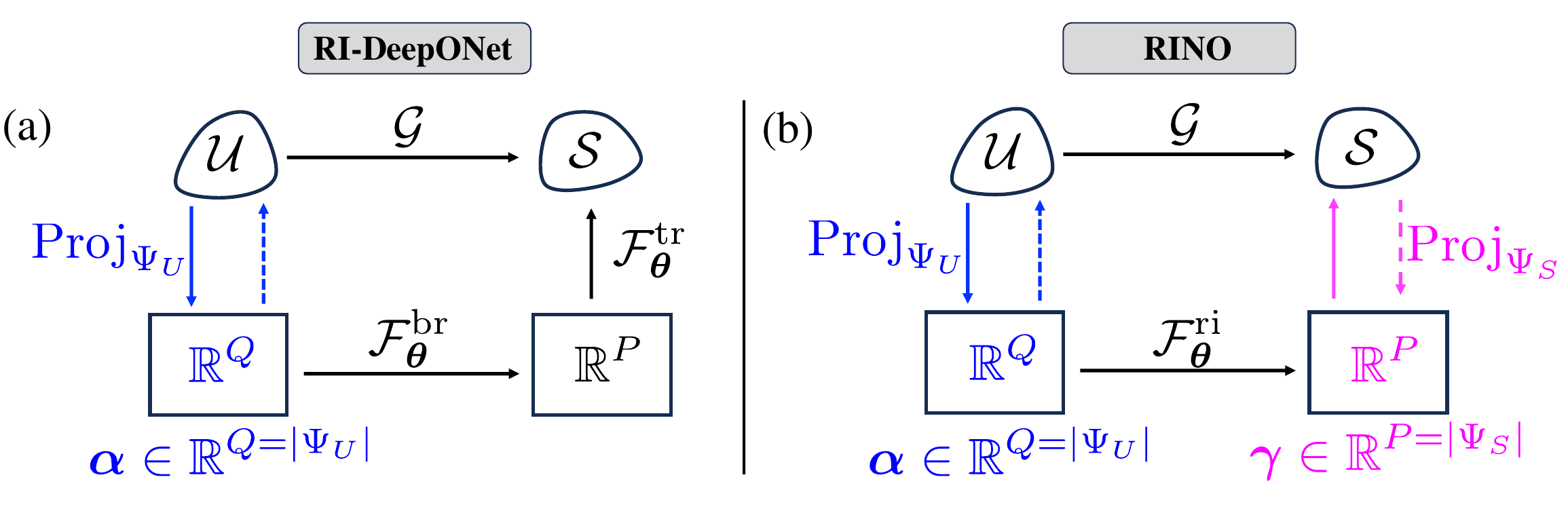}
  \caption{
  Operator learning with RINO when the output basis functions used to span the output field are either (a) unknown or (b) fixed during the operator learning stage. In (a), only the input functions are embedded using a learned set of basis function and the operator is learned between the embedding space and the output directly by training both branch ($\mathcal{F}_{\boldsymbol{\theta}}^{\text{br}}$) and trunk ($\mathcal{F}_{\boldsymbol{\theta}}^{\text{br}}$) networks. In (b), is functions are learned separately for both output and input function data. Consequently, the operator learning task effectively becomes a mapping between these two constructed subspaces via a single nueral network $\mathcal{F}_{\boldsymbol{\theta}}^{\text{ri}}$.
  }
  \label{fig:func-space-mapping}
\end{figure}

In the proposed RINO, we develop dictionary learning algorithms that adaptively learn basis functions to approximate signals defined on arbitrary point clouds. This allows us to project arbitrary signals onto these basis functions and use their corresponding coefficients as appropriate representations in the embedding space. This is the nature of the embedding of the input function in the RI-DeepONet from Eq.~\eqref{eqn:RINO_representation}. But, if such a method exists (as discussed in the next section), it can also be applied \textit{separately} to the output function data. This process would build an appropriate dictionary of basis functions for the output data, which can subsequently be used in the operator learning task as a fixed, predefined basis (analogous to the trunk in POD-DeepONet) as showed in Figure~\ref{fig:func-space-mapping}(b). The RINO is then built to approximate the operator that maps between these embedded functions spaces as illustrated by $\mathcal{F}_{\boldsymbol{\theta}}^{\text{ri}}$ in Figure~\ref{fig:func-space-mapping}(b).

For now, let us additionally assume that the output dictionary $\boldsymbol{\Psi}_S(y) = \{\psi_{l}^{S}(y)\}_{l=1}^{P}$ is known and available for use where $P=|\Psi_S|$. The availability of the input dictionary, denoted by $\boldsymbol{\Psi}_U(x) = \{\psi_{l}^{U}(x)\}_{l=1}^{Q}$ where $Q=|\Psi_U|$, was already assumed in the previous section. Under this setup, which is schematically shown in Figure~\ref{fig:func-space-mapping}(b), the objective function for operator learning is defined as:
\begin{equation}
    \underset{\boldsymbol{\theta}}{\text{argmin}}
    \
    \mathbb{E}_{i}
    \mathbb{E}_{(y, s)\sim\mathcal{D}_s^{(i)}}
    \underbrace{
    \left\|
    s
    -
    \text{Recon}_{\Psi_S}
    \left(
        y; \bar{\boldsymbol{\gamma}}^{(i)}_{\boldsymbol{\theta}}
    \right)
    \right\|_{2}^2
    }_
    {\text{prediction loss}}
    +
    \tau
    \underbrace{
    \left\|
    \boldsymbol{\gamma}^{(i)}
    -
    \bar{\boldsymbol{\gamma}}^{(i)}_{\boldsymbol{\theta}}
    \right\|_{2}^2
    }_
    {\text{embedding consistency loss}}
    ,
\label{eq:loss-two-basis}
\end{equation}
where
$
\bar{\boldsymbol{\gamma}}^{(i)}_{\boldsymbol{\theta}} = \mathcal{F}_{\boldsymbol{\theta}}^{\text{ri}}(\boldsymbol{\alpha}^{(i)})
$, 
$
\boldsymbol{\gamma}^{(i)}
=
\text{Proj}_{\Psi_S}(\mathcal{D}^{(i)}_s)
$,
$
\boldsymbol{\alpha}^{(i)}
=
\text{Proj}_{\Psi_U}(\mathcal{D}^{(i)}_u)
$, and
$\tau$ is a penalty factor that controls the extent to which the second term is enforced. 
%
%Note that we retain the superscript $(\cdot)^{\text{br}}$ in the learned operator to acknowledge the analogy to the branch net in a DeepONet, but point out that the architecture in Figure~\ref{fig:func-space-mapping}(b) is, in fact, new in the sense that the neural operator in this setting is comprised of a single neural network operating between embedded function spaces.  
%
Note that $\mathcal{F}^{\text{ri}}$ can be viewed as analogous to the branch net in DeepONet, however the architecture in Figure~\ref{fig:func-space-mapping}(b) is different in the sense that the neural operator in this setting is comprised of a single neural network operating between embedded function spaces.
The first term in this objective function corresponds to the prediction part of the operator network, while the second term helps the branch network produce a representation consistent with the coefficients of the output basis functions.
The reconstruction operator for a dictionary $\boldsymbol{\Psi}$ is defined as follows:
\begin{equation}
    \text{Recon}_{\Psi}(\cdot;\boldsymbol{\gamma})
    \triangleq
    \sum_{l=1}^{|\Psi|}
    \ 
    \psi_l(\cdot) \ \gamma_l.
\end{equation}
The dictionary projection operator will be specified in the next section.

\textbf{A Discussion:}
If the second term of the training loss reaches zero, the first term will also achieve its minimum. This is because, prior to the operator learning step, the output basis functions are learned in a data-driven manner to ensure they can represent the output function within a specified tolerance. However, since the basis functions are not guaranteed to be perfectly orthogonal (as orthogonality is weakly enforced on finite sample points in a data-driven manner), there is a slight possibility of multiple sets of basis coefficients for a given realization producing the same minimal output reconstruction error. This underscores the importance of enforcing orthogonality, as it ensures the uniqueness of the embedding representation, thereby improving both stability and interpretability. In such cases, while the second term of the loss may have a higher error, the first term remains minimal. Consequently, the first term is the most critical, and the second term can be ignored. However, this simplification may lead to slower convergence during training. To address this (computational) challenge, one can pre-condition the operator network by initially enforcing the second term more strictly and gradually reducing its penalty weight as training progresses. Alternatively, one could bypass the first term entirely throughout the optimization process, depending on the specific training strategy, application requirements, and, importantly, the size of the embedding. This is because, in very high-dimensional spaces, the Euclidean distance between any two random vectors tends to become negligible, reducing the practical significance of the second term. However, in this work, since the learned basis functions are (weakly) orthogonal and the embedding dimensions do not exceed 100, we focus solely on the second loss term to expedite the operator learning training process.

\subsection{Dictionary Learning}
\label{sec:DL}
In this section, we introduce two new dictionary learning algorithms to identify a set of basis functions parameterized by neural networks, which are used to approximate signals defined on point cloud data. 
The first algorithm aims to adaptively find basis functions that capture the \textit{expected} value of the residual across batches of signal realizations, while the second algorithm aims to find a basis each time one is needed to approximate only one realization at a time.

\subsubsection{Batch-wise Learning}

\begin{figure}[h]
  \centering
    \includegraphics[width=0.6\textwidth]{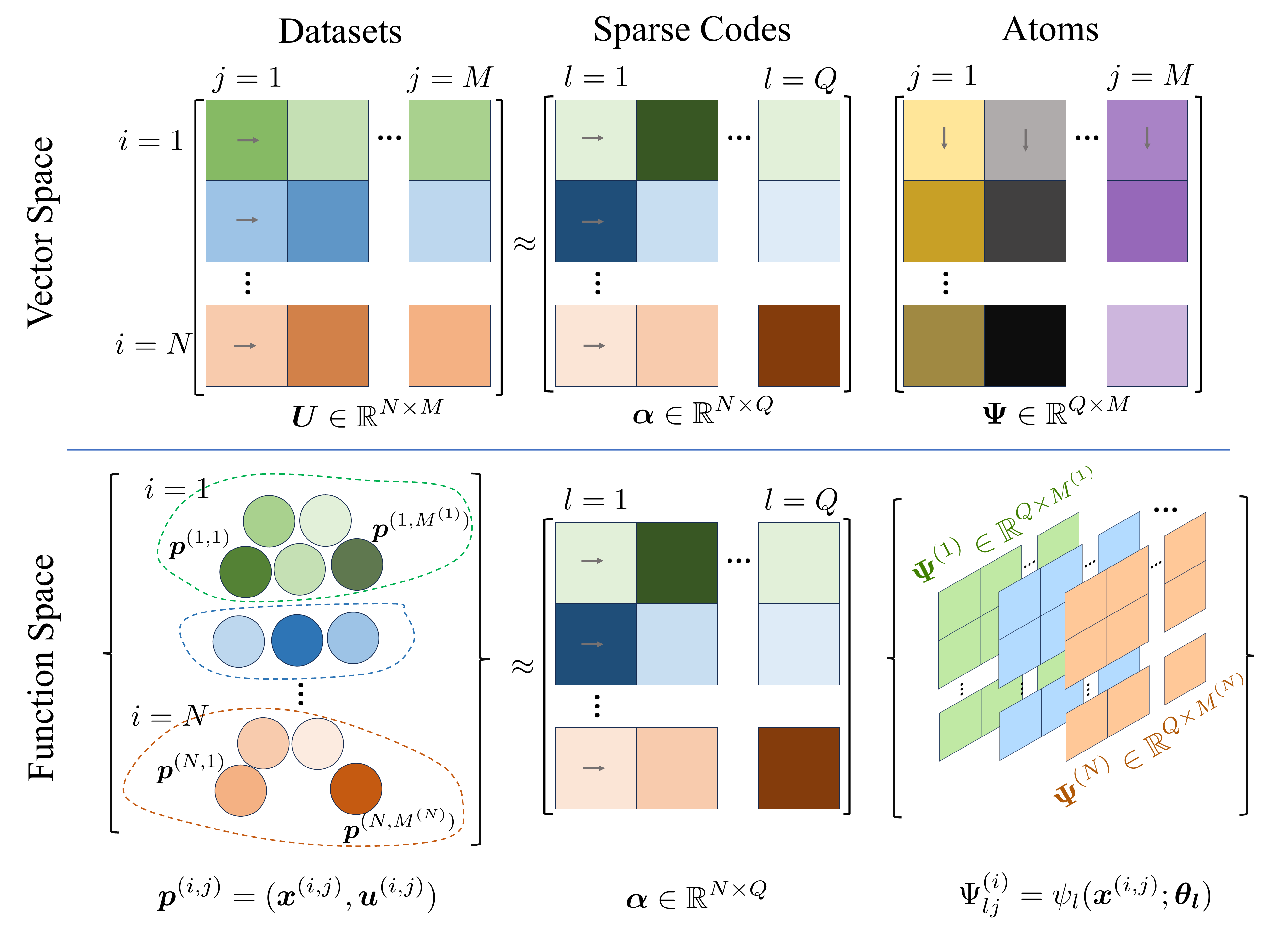}
  \caption{
  Dictionary learning (\textbf{top}) in the classical vector space setup and (\textbf{bottom}) in the proposed function space setup. The proposed method can be directly applied to point cloud signals that are sampled arbitrarily (but sufficiently richly) in terms of the number and location of discretized points. In the proposed method, classical discrete atoms (basis vectors) are replaced by continuous, fully differentiable functions parameterized by neural networks.
  }
  \label{fig:dict-learn}
\end{figure}

Given a collection of $N$ signals $\boldsymbol{U} \in \mathbb{R}^{N\times M}$, which are discretized at $M$ locations and assumed to be correlated and share a common structure, the classical dictionary learning goal is to learn a set of $Q$ basis vectors (or \textit{atoms}), $\boldsymbol{\Psi}\in \mathbb{R}^{Q\times M}$, such that all signals in the set can be sufficiently approximated using a sparse linear combination of these basis vectors (see top of Figure~\ref{fig:dict-learn}). A classical example of dictionary learning is PCA, where the goal is to approximate high-dimensional, correlated signals using a few principal components, which serve as basis vectors. 
Here, we go beyond discovering dictionaries that exist in finite-dimensional vector spaces and instead focus on finding those that reside in continuous function spaces, making them suitable for processing signals defined on irregular domains such as point clouds (see bottom of Figure~\ref{fig:dict-learn}).

Throughout the following sections, we describe the procedure by learning a dictionary that can represent the input function $u(x)$; however, the same procedure is applied to the output function as well in the RINO. In a general setup, we are given a dataset of $N$ recorded signals, denoted by $\mathcal{D}_{u}=\{
\mathcal{D}_{u}^{(i)} \}_{i=1}^{N}$, where each element $\mathcal{D}_{u}^{(i)}$ contains an arbitrarily discretized version of the $i$-th function realization $u^{(i)}$, as described in the previous section and depicted in the bottom of Figure~\ref{fig:dict-learn}. If the number and locations of sensors (discretization points) are identical for all signals, the data setup resembles the classical PCA setup.

We aim to find a common set of $Q$ basis functions $\boldsymbol{\Psi}(x) = \left\{\psi_l(x)\right\}_{l=1}^{Q} \equiv [\psi_1(x), \dots, \psi_Q(x)]^T$, \textit{continuously} defined over the domain of $u$, such that the signal realizations can be represented sufficiently well by a linear combination of these basis functions, as follows:
\begin{equation}
    u^{(i)}(x) \approx
    \sum_{l=1}^Q \alpha^{(i)}_l \ \psi_{l}(x) =
     \boldsymbol{\Psi}^T(x) \ \boldsymbol{\alpha}^{(i)},
\end{equation}
where $\alpha_l^{(i)}$ is the coefficient corresponding to the $l$-th basis function for the $i$-th signal realization.
Naturally, one would be interested in orthogonal basis functions to eliminate redundancy. This property also enables faster computation at inference since the projection onto orthogonal basis functions can be formulated as a convex optimization problem. Additionally, certain statistical analyses become more straightforward; for instance, the total variance of the random field can be decomposed into the sum of the variances of the individual coefficients. Similarly, this reduction can be applied to statistical moments of higher orders.
%Most importantly, since we will use these bases to introduce a coordinate system, orthogonality is crucial. This ensures that each coordinate (code or embedding) is independent of the others, which is beneficial for the downstream ML task.

\begin{definition}[Orthogonal Functions]
Two functions $\psi_i(x)$ and $\psi_j(x)$ are orthogonal over the domain $x$ if their inner product $\langle \psi_i, \psi_j\rangle $ is zero for $i\neq j$, i.e.,
\begin{equation}
    \langle \psi_i, \psi_j\rangle \triangleq \int_x \psi_i(x) \ \psi_j(x) \ dx = 0, \quad \text{if } i \neq j.
\end{equation}

\end{definition}

These basis functions can be predefined using well-known orthogonal bases, such as Fourier bases or those used in polynomial chaos expansion (PCE) \citep{novak2024physics,blatman2011adaptive,ghanem2003stochastic,xiu2002wiener,wiener1938homogeneous}. However, similar to PCA \textbf{basis vectors}, it would be more flexible and potentially more efficient if the \textbf{basis functions} could be data-dependent.
%In fact classical dictionary learning goal is to find such data-dependent bases for \bold{bases vector} defined on predifiemd discrete support
To this end, we parameterize these functions as neural networks $\psi_l(x;\boldsymbol{\theta}_l)$ where $\boldsymbol{\theta}_l$ concatenates all parameters associated with the $l$-th neural network basis function, as follows:
\begin{equation}
    u^{(i)}(x) \approx
     \boldsymbol{\Psi}^T(x; \boldsymbol{\theta}) \ \boldsymbol{\alpha}^{(i)},
     \label{eq:dict-frwd}
\end{equation}
where $\boldsymbol{\theta}$ concatenates the parameters of all neural network basis functions. Now, the challenge is to learn the parameterized dictionary $\boldsymbol{\Psi}(x; \boldsymbol{\theta})$ such that its atoms are orthogonal to one another. 

\begin{remark}
    In this section, the symbol $\theta$ is reused to denote the parameters of the neural networks $\psi_l$. It is important to note that these parameters are distinct from those used for $\mathcal{F}_{\theta}$ in Section~\ref{sec:OL}, despite the notation being the same.
\end{remark}

The projection operator $\text{Proj}_{\Psi}(\cdot): \mathcal{U} \mapsto \mathbb{R}^{|\Psi|}$ is defined with respect to the dictionary $\boldsymbol{\Psi}$ with $|\Psi|=Q$ number of atoms, which 
returns the coordinates of the projection of the input function $u(x)$ onto the subspace spanned by the dictionary’s basis functions, such that an error measure (in the empirical risk minimization sense) between the original signal and its projection is minimized, as follows,
\begin{equation}
    \boldsymbol{\alpha} =
    \text{Proj}_{\Psi}[u(x)] \triangleq \underset{\boldsymbol{\alpha}}{\text{argmin}}\
    ||
    u(x)-
    \boldsymbol{\Psi}^T(x; \boldsymbol{\theta}) \ \boldsymbol{\alpha}
    ||_2^2
    + \mathcal{R}(\mathcal{\boldsymbol{\alpha}}).
    \label{eq:dl-rec-loss}
\end{equation}
Here, the $L_2$-norm is chosen as the error measure with respect to a well-defined inner product $||r(x)||_2^2=\langle r(x), r(x)\rangle$, and $\mathcal{R(\boldsymbol{\alpha})}$ is a regularization term, which can promote sparsity in the solution and/or improve the condition number of the Hessian matrix for optimization, depending on its form. We apply $L_2$ regularization $\mathcal{R(\boldsymbol{\alpha})} = \lambda || \boldsymbol{\alpha}||_2^2$, where $\lambda \in \mathbb{R}^+$ is a penalty parameter that controls the influence of the regularizer. Hence, one can easily show that the coordinates can be found via,
\begin{equation}
    \alpha_l =
    \text{Proj}_{\psi_l}[u(x)]
    =
    \frac{\langle u(x), \psi_l(x)\rangle}{\langle \psi_l(x), \psi_l(x)\rangle + \lambda}; \quad 1\le l \le |\boldsymbol{\Psi}|.
    \label{eq:continious-alpha}
\end{equation}
For the optimal values, and when $\lambda=0$, one can easily show that the residual, $r(x) = u(x) -  \boldsymbol{\Psi}^T(x; \boldsymbol{\theta}) \ \boldsymbol{\alpha}$, is orthogonal to the basis functions, hence to the subspace spanned by the dictionary. Therefore, for small values of $\lambda$ (i.e., $\lambda \to 0^+$), one can expect the residual to be ``approximately'' orthogonal to this subspace.
%In other words, the residual cannot be expressed linearly by other basis functions and hence introduces a new functional direction (or axis).
In other words, the residual cannot be expressed linearly by the other basis functions. Therefore, the residual can serve as an effective choice for a new basis function to enhance the representation of the data, while maintaining (approximate) orthogonality. This is similar to the vector space viewpoint, where the residual vector is used as a new direction for data representation.
This simple yet powerful concept has also been employed in diffusion maps (an unsupervised ML method \citep{nadler2006diffusion,coifman2006diffusion}) to identify unique eigendirections through local linear regression \cite{dsilva2018parsimonious}. We utilize this intuitive fact to derive an algorithm that aims to iteratively construct an approximately orthogonal dictionary of atoms as descibed in Algorithm \ref{algo:DL}.

\begin{algorithm}
\caption{Dictionary Learning (Batch-wise)}
\label{algo:DL}
\begin{algorithmic}[1] % The number tells where the line numbering should start

\State \textbf{Input:} 
\State $
\mathcal{D}_{u}=\left\{
\mathcal{D}_{u}^{(i)} \right\}_{i=1}^{N}, \mathcal{D}_{u}^{(i)} = \left\{
\left(
x^{(i, j)}, \bar{u}^{(i, j)}
\right)
\right\}_{j=1}^{M^{(i)}}$ \Comment{$N$ realizations of the input function $\bar{u}(x)$}
\State $\boldsymbol{\Psi} = \{1\}$ \Comment{A trivial dictionary}
\State $\eta\in \mathbb{R}^{+}$ \Comment{Optimizer learning rate}
\State $0 < \text{Tol.} \ll 1$ \Comment{A small target tolerance to be achieved for the reconstruction}
\State \textbf{Output:}  $\boldsymbol{\Psi} = \left\{ \psi_l(\cdot; \boldsymbol{\theta}_l) \right\}_{l=1}^{|\boldsymbol{\Psi}|}$ \Comment{A parameterized dictionary by $|\boldsymbol{\Psi}|$ neural networks}
%
%\State $max \gets a$ \Comment{Initialize max as a}
\While{$\text{Recon. Err.} \ge \text{Tol.}$}
    \State Add a randomly initialized neural network basis $\psi^{\text{new}}(x;\boldsymbol{\theta}^{\text{new}})$ to the dictionary $\boldsymbol{\Psi}(x;\boldsymbol{\theta}, \boldsymbol{\theta}^{\text{new}})$
    \For{$e \gets 1$ to $N_{\text{epoch}}$}
        \For{$i \gets 1$ to $N$} \Comment{Loop over realizations}
        \State $\boldsymbol{\alpha}^{(i)} = \text{Proj}_{\Psi}[u^{(i)}(x)] \approx \underset{\boldsymbol{\alpha}}{\text{argmin}}\
    \underset{(x, \bar{u})\sim\mathcal{D}_{u}^{(i)}}{\mathbb{E}}\left\|
    \bar{u} -
    \boldsymbol{\Psi}^T(x; \boldsymbol{\theta}, \boldsymbol{\theta}^{\text{new}}) \ \boldsymbol{\alpha}
    \right\|_2^2
    + \mathcal{R}(\mathcal{\boldsymbol{\alpha}})$
        \EndFor
        \State $\boldsymbol{\theta}^{\text{new}} \gets \boldsymbol{\theta}^{\text{new}} - \eta \nabla_{\boldsymbol{\theta}^{\text{new}}} \underset{i}{\mathbb{E}}\left[ \underset{(x,\bar{u})\sim\mathcal{D}_{u}^{(i)}}{\mathbb{E}}\left\|
    \bar{u}(x) -
    \boldsymbol{\Psi}^T(x; \boldsymbol{\theta}, \boldsymbol{\theta}^{\text{new}}) \ \boldsymbol{\alpha}^{(i)}
    \right\|_2^2\right]$ \Comment{Batch-wise gradient descent}
    \EndFor
    \State Calculate reconstruction error.
\EndWhile
\end{algorithmic}
\end{algorithm}

To promote orthogonality, each time we add a new basis function $\psi_{\text{new}}$, this function is used to capture the expected value of the residual function over the entire dataset. This can be observed in line 13 of the algorithm, as follows:
\begin{equation}
    \mathcal{L}(\boldsymbol{\theta}_{\text{new}}) : 
    \bar{u} - \boldsymbol{\Psi}^T(x; \boldsymbol{\theta}, \boldsymbol{\theta}_{\text{new}})\ \boldsymbol{\alpha}
    =
    \underbrace{
    \bar{u} - \sum_{l=1}^{|\Psi| - 1} \psi_l(x; \boldsymbol{\theta}_l)\ \alpha_l
    }_{\text{current residual}}
     - 
    \psi_{\text{new}}(x; \boldsymbol{\theta}_{\text{new}})\ \alpha_{\text{new}}.
\end{equation}
Note that here the basis functions and their sparse coefficients are unknown, making the optimization challenging. We adopt ideas from the Alternating Direction Method of Multipliers (ADMM) \cite{boyd2011distributed} by separating these two unknowns. In the first step (line 11 in Algorithm~\ref{algo:DL}), we assume the basis functions are fixed and find the corresponding coefficients. Then, in the second step (line 13 in Algorithm~\ref{algo:DL}), we assume the coefficients are fixed and optimize the basis functions. This approach bypasses the need for backpropagation through an optimization step in the first stage, which, while possible, can be computationally expensive. Additionally, performing backpropagation through a linear solver with a poorly conditioned kernel (due to inaccurate basis functions) can lead to unstable optimization.

As previously mentioned, the first ADMM step (line 13 in Algorithm \ref{algo:DL}) can be executed efficiently by appropriately choosing the regularizer and ensuring the orthogonality of the dictionary. Although the $L_1$-norm regularizer may be more effective in promoting sparsity, the $L_2$-norm regularizer facilitates faster computation and implementation because the optimization can be formulated as a linear system. Consequently, the dictionary coefficient $\boldsymbol{\alpha}^{i} \in \mathbb{R}^{|\Psi|}$ for the $i$-th realization can be found via:
\begin{equation}
    \boldsymbol{\alpha}^{(i)} = 
    \left(
         \boldsymbol{\Psi}^{(i)}
         {\boldsymbol{\Psi}^{(i)}}^T
        +
        \lambda \boldsymbol{I}
    \right)^{-1}
    {\boldsymbol{\Psi}^{(i)}} \boldsymbol{U}^{(i)};
    \quad
    \boldsymbol{\Psi}^{(i)}\in \mathbb{R}^{|\Psi|\times M^{(i)}}
    \quad
    \boldsymbol{U}^{(i)}\in \mathbb{R}^{M^{(i)}},
\end{equation}
where $\boldsymbol{\Psi}^{(i)}(\boldsymbol{x}; \boldsymbol{\theta})$ represents the basis function values at the discretization points of the $i$-th realization $\{\boldsymbol{x}^{(i, j)}\}_{j=1}^{M^{(i)}}$, and $\boldsymbol{U}^{(i)}$ are the $i$-th realization values at those corresponding points $\{\boldsymbol{u}^{(i, j)}\}_{j=1}^{M^{(i)}}$. Notice that the kernel of the linear system scales with the dictionary size and not the discretization resolution. Therefore, this implementation is usually more computationally favorable than Eq.~\eqref{eq:continious-alpha}, which requires numerical integration and scales with the resolution size, particularly in cases where the number of basis functions is much less than the discretization resolution, which is the case in the numerical examples.

We have studied the ability of the proposed algorithm to find appropriate orthogonal basis functions in Appendix~\ref{appx:gpod}, where we compare the proposed algorithm with classical Gappy PCA or GPOD \cite{everson1995karhunen} in identifying basis functions from masked, incomplete data. Moreover, the possibility of using random basis functions is also explored and discussed.

\subsubsection{Sample-wise Learning}
By assuming the existence of a set of orthogonal basis functions $\psi_{i}$ whose linear combinations can represent the signals, and since the signals themselves belong to the subspace spanned by these basis functions, it follows that each basis function can be represented as a linear combination of the signals. This is essentially a generalization of the Gram-Schmidt orthogonalization to function spaces, where a new basis function $\psi_{\text{new}}^{\text{gs}}(x)$ is constructed according to a new signal $u_{\text{new}}(x)$ as follows:
\begin{equation}
    \psi_{\text{new}}^{\text{gs}}(x)
    =
    u_{\text{new}}(x)
    -
    \sum_{j=1}^{|\Psi^{\text{gs}}|}
    \alpha^{\text{new}}_j
    \psi_{j}^{\text{gs}}(x),
\end{equation}
where the right-hand side is the residual between the new signal and its best reconstruction with the current basis functions available in the dictionary. Hence, the coefficients $\alpha^{\text{new}}_j$ are the projections of the new signal onto the previously found basis functions $\psi_{j}^{\text{gs}}(x)$, see Eq.~\eqref{eq:continious-alpha}. It is straightforward to show that if $u_{\text{new}}(x)$ does not belong to the dictionary subspace, the residual signal $\psi_{\text{new}}^{\text{gs}}(x)$ is orthogonal to the subspace. The complete procedure of constructing an appropriate dictionary with (weakly) orthogonal basis functions parameterized by neural networks is presented in Algorithm~\ref{algo:DL-sbs}.

\begin{algorithm}
\caption{Dictionary Learning (Sample-wise)}
\label{algo:DL-sbs}
\begin{algorithmic}[1] % The number tells where the line numbering should start

\State \textbf{Input:} 
\State $
\mathcal{D}_{u}=\left\{
\mathcal{D}_{u}^{(i)} \right\}_{i=1}^{N}, \mathcal{D}_{u}^{(i)} = \left\{
\left(
\boldsymbol{x}^{(i, j)}, \bar{u}^{(i, j)}
\right)
\right\}_{j=1}^{M^{(i)}}$ \Comment{$N$ realization of the input function $\bar{u}(\boldsymbol{x})$}
\State $\boldsymbol{\Psi}^{\text{gs}} = \emptyset$ \Comment{A trivial dictionary}
\State $\eta\in \mathbb{R}^{+}$ \Comment{Optimizer learning rate}
\State $0 < \text{Tol.} \ll 1$ \Comment{A small target tolerance to be achieved for the reconstruction}
\State \textbf{Output:}  $\boldsymbol{\Psi}^{\text{gs}} = \left\{ \psi_l^{\text{gs}}(\cdot; \boldsymbol{\theta}_l) \right\}_{l=1}^{|\boldsymbol{\Psi}^{\text{gs}}|}$ \Comment{A parameterized dictionary by $|\boldsymbol{\Psi}^{\text{gs}}|$ neural networks}
\For{$i \gets 1$ to $N$}
    \State $r^{(i)}(\boldsymbol{x}; \boldsymbol{\alpha}, \boldsymbol{\theta}) = u^{(i)}(\boldsymbol{x}) -
    {\boldsymbol{\Psi}^{\text{gs}}}^T(\boldsymbol{x}; \boldsymbol{\theta})\ \boldsymbol{\alpha}$ \Comment{The residual function associated with the $i$-th signal}
    \State $\boldsymbol{\alpha}^{(i)} =  \underset{\boldsymbol{\alpha}}{\text{argmin}}\
    \underset{\boldsymbol{x}\sim\mathcal{D}_{u}^{(i)}}{\mathbb{E}}\left\|
    r^{(i)}(\boldsymbol{x}; \boldsymbol{\alpha}, \boldsymbol{\theta})
    \right\|_2^2
    + \mathcal{R}(\mathcal{\boldsymbol{\alpha}})$ \Comment{The best estimator of basis coefficients}
    \State $\text{Err} = \underset{\boldsymbol{x}\sim\mathcal{D}_{u}^{(i)}}{\mathbb{E}}\left\|
    r^{(i)}(\boldsymbol{x}; \boldsymbol{\alpha}^{(i)}, \boldsymbol{\theta})
    \right\|_2^2$ \Comment{The estimated residual norm of the best estimator}
    \If{$\text{Err.}\le\text{Tol.}$} \Comment{Check if the reconstruction error is satisfactory}
        \State \textbf{continue}
    \EndIf
    \State $\boldsymbol{\theta}_{\text{new}} =  \underset{\bar{\boldsymbol{\theta}}}{\text{argmin}}\
    \underset{(\boldsymbol{x}, \bar{u})\sim\mathcal{D}_{u}^{(i)}}{\mathbb{E}}\left\|
    \psi^{\text{gs}}_{\text{new}} (\boldsymbol{x}; \bar{\boldsymbol{\theta}})-
    r^{(i)}(\boldsymbol{x}; \boldsymbol{\alpha}^{(i)}, \boldsymbol{\theta})
    \right\|_2^2
    $ \Comment{Train a new basis to capture the residual signal}
    \State $\boldsymbol{\Psi}^{\text{gs}} \gets \boldsymbol{\Psi}^{\text{gs}} \cup \{\psi^{\text{gs}}_{\text{new}} (\boldsymbol{x}; \boldsymbol{\theta}_{\text{new}})\}$ \Comment{Update the dictionary}
\EndFor
%
%
%
%\State $max \gets a$ \Comment{Initialize max as a}
\end{algorithmic}
\end{algorithm}

\textbf{Discussion: Computational Complexity and Efficiency --} 
Although these two algorithms share similarities, they may have different computational costs and training stability. For example, time complexity of Algorithm~\ref{algo:DL-sbs} scales linearly with the number of realizations, which can be mitigated by the batch-wise training feature of Algorithm~\ref{algo:DL}. Additionally, the basis functions identified by Algorithm~\ref{algo:DL-sbs} may be more sensitive to the order of the data stream since the first basis is derived from the first data it encounters. Moreover, when the data per realization is severely sparse, this algorithm may produce initial basis functions that are overfitted and not representative of the entire set of realizations. Consequently, under such scenarios, the initial basis functions may introduce highly oscillatory components, leading to noisy information during the basis identification for other realizations, which do not reflect the statistical properties of the entire dataset. The answer to all these questions requires a separate study and comparison between these two algorithms under various scenarios, which is beyond the main focus of this research. Hence, in this work, we primarily focus on the application of Algorithm~\ref{algo:DL} and only showcase the application of Algorithm~\ref{algo:DL-sbs} in the final numerical example.

\textbf{Discussion: Normalization --}
From a theoretical perspective, the introduced algorithms aim to (weakly) enforce orthogonality of the identified basis functions. However, for several reasons, one might be interested in basis functions that are normalized as well, i.e., orthonormal basis functions. 
One reason could be attributed to potential identifiability issues associated with Eq.~\eqref{eq:dict-frwd}, where both the basis functions and coefficients are unknown. If one scales up the coefficients by a factor and simultaneously scales down the basis functions by the same factor, the same representation can be achieved, although the unknowns themselves have changed. The ADMM method mitigates this issue by solving for each unknown separately while fixing the others, thereby preventing simultaneous changes. However, controlling the scale or norm of either set of unknowns may further suppress such potential issues. 
Also, the relative scale of each basis function compared to others may impact the scale of its corresponding optimal coefficient relative to the coefficients of other basis functions. Since we may later use the basis coefficients in other downstream tasks (here, operator learning), it may be helpful to have learned bases with relatively the same scaling or norm. This way, the scale of their corresponding coefficients will have a particular meaning in terms of their relative importance in the prediction (coefficients with higher values indicate that their corresponding bases have more impact in the prediction) which further enhances the interpretability of the proposed method.
Several strategies are possible for the normalization of the learned basis functions: (1) one may use a normalization layer in the output layer of the neural network \textit{during} the training to adjust the basis functions according to the batch statistics. In this study, we used an $L_2$-norm normalization layer of this type, where the normalized basis is $\psi_{\text{norm}}(\boldsymbol{x}; \boldsymbol{\theta}) = \psi(\boldsymbol{x}; \boldsymbol{\theta}) / \sqrt{\mathbb{E}_x[\psi^2(\boldsymbol{x}; \boldsymbol{\theta})]}$. This type of normalization shares similarities with other batch-wise normalization layers commonly used in ML \cite{ba2016layer,zhang2019root}. (2) If the batch statistics of the given data do not accurately represent the actual statistics, one may normalize the learned basis \textit{after} its training. To do this, a sufficiently large number of points can be artificially generated to empirically estimate the norm of the current un-normalized basis function, using a Monte Carlo estimate. This scaling factor can then be saved for later use before proceeding to learn the next basis function. In this work, we used the first approach in the numerical examples unless otherwise specified.

\subsection{Parameterization via Implicit Neural Representation (Neural Field)}
\label{sec:inr}
In this section, we provide details on how we chose the neural network parameterization for the unknown basis functions introduced in the previous section.

%As outlined in the previous section, each basis function $\psi_l(\boldsymbol{x})$ acts as a template, or mode, for representing an arbitrary field $u(\boldsymbol{x})$ through a linear combination of these templates.
%%To achieve a continuous and differentiable representation of $\psi_l(x)$, we utilize the implicit neural representation (INR) approach \cite{sitzmann2020implicit}. 
%%This approach continuously parameterizes an arbitrary function over its domain using multilayer perceptrons (MLPs). It has been primarily applied to image data, modeling images as functions defined over their spatial domain.
%
%%INR methods offer several advantages. First, the functions are defined on the continuous domain of $x$ rather than on a discrete grid, enabling the representation to adapt flexibly to various resolutions. Second, this approach ensures that the function is fully continuous and differentiable with respect to $x$, facilitating the use of automatic differentiation, which is crucial for problems requiring access to the function’s gradient. Lastly, they have shown greater memory efficiency compared to grid-based representations \cite{mescheder2019occupancy}, as their capacity to model fine details depends on the architecture of the model rather than the grid resolution.
Use cases of INRs or Neural fields in the literature mostly deal with modeling the signal itself directly. However, in this work, instead of applying the INRs directly to the signal $u(\boldsymbol{x})$, we apply them to the basis functions $\psi_l(\boldsymbol{x})$, which serve as templates for modeling $u(\boldsymbol{x})$. One challenge in utilizing INRs is that conventional MLPs, despite their universal approximation property \cite{hornik1989multilayer}, struggle to learn fine details or high-frequency content due to a phenomenon known as spectral bias \cite{rahaman2019spectral}.
Several recent works aim to address this issue by modifying the conventional MLP architecture. Two prominent approaches include using sinusoidal activation functions, known as the SIREN approach \cite{sitzmann2020implicit}, and incorporating random Fourier features of different frequencies in the first layer of the MLP, known also as positional encoding for language modeling applications \citep{mildenhall2021nerf,tancik2020fourier, rahimi2007random}. Here, we use the SIREN approach to parameterize each newly added basis function $\psi_l(x;\boldsymbol{\theta}_l)$ in Algorithm~\ref{algo:DL} due to its simplicity and the extensive success reported in various computer vision tasks \citep{sitzmann2020implicit,chan2021pi,xie2022neural}.

We parameterize each basis function $\psi_l(x)$ using a SIREN MLP with $n$ hidden layers as follows:
\begin{equation}
    \psi_l(\boldsymbol{x}; \boldsymbol{\theta}_l) = \boldsymbol{W}^{(l)}_{n+1}
    \left(
    \phi_{n}\circ \phi_{n-1} \cdots \phi_{1}
    \right)(\boldsymbol{x}) + \boldsymbol{\beta}^{(l)}_{n+1}; \quad \phi_m(\boldsymbol{h}^{(l)}_{m-1}) \triangleq 
    \sin
    \left(
    \boldsymbol{W}^{(l)}_{m} \boldsymbol{h}^{(l)}_{m-1} + \boldsymbol{\beta}^{(l)}_m
    \right), 
\end{equation}
where $\boldsymbol{h}^{(l)}_m$ denotes the $m$-th hidden state, with $h_{0}\triangleq \boldsymbol{x}$ and $0\le m \le n$. $\boldsymbol{W}^{(l)}_{m}$ and $\boldsymbol{\beta}^{(l)}_{m}$ represent the typical weights and biases of the MLP, respectively. The vector $\boldsymbol{\theta}_l$ concatenates all the trainable parameters of the SIREN, i.e., $\boldsymbol{\theta}_l \equiv \left\{\boldsymbol{W}^{(l)}_{m}, \boldsymbol{\beta}^{(l)}_{m}\right\}_{m=1}^{n+1}$.
The convergence and accuracy of the SIREN depend on the appropriate initialization of the trainable parameters. Following the method proposed by \cite{sitzmann2020implicit}, we initialize the weights drawn from a uniform distribution $\boldsymbol{W}_m \sim \text{Uniform}[-\omega_{0}\sqrt{6/\text{dim}(\boldsymbol{h}_{m-1})}, \omega_{0}\sqrt{6/\text{dim}(\boldsymbol{h}_{m-1})}]$ where the frequency parameter $\omega_{0}=1$ for all layers except the first layer, in which it is a hyperparameter that depends on the problem at hand. This ensures that the pre-activations are normally distributed with unit variance, leading to more stable training by preventing vanishing or exploding gradients. The biases are set to zero during the initialization stage.

\begin{remark}
Although INRs could be more memory efficient per instance signal compression (e.g., an image), their vanilla use case may not seem ideal when dealing with many instances. This is because each instance requires a separate INR with different parameters to be trained from scratch, without leveraging any information obtained from other instances or any potential common structures across instances. Notice that each training of an INR involves solving a \textbf{non-convex} optimization problem, which is typical for deep neural network training. One approach currently under exploration is to use meta-learning for better initialization of each INR, leading to fewer iterations, or some form of parameter sharing across instances \citep{sitzmann2020metasdf, tancik2021learned,dupont2022data}. However, as explained, in the proposed approach we do not directly model signals with INRs. Instead, we learn a dictionary of basis functions as INRs in an \textit{offline} manner, which indirectly captures the common structure across instances. Then, at inference or for an unseen signal, we only solve a \textbf{convex} optimization problem to find the coefficients of the INR basis. This eliminates the need to introduce or train a new INR from scratch. In that regard, using INRs as basis functions may offer new ways to reduce the computational burden during inference when dealing with implicit representations of signals; however, this is not the focus of the current work.

%This contrasts with typical use cases of INRs, where a separate non-convex optimization is required for each instance at inference, which can be computationally costly. 
%However, there have been attempts to reduce this computational burden through meta-learning \citep{sitzmann2020metasdf, tancik2021learned}. 
%Using INRs as basis functions may offer new ways to reduce the computational burden during inference when dealing with implicit representations of signals; however, this is not the focus of the current work.
\end{remark}

\section{Numerical Examples}
\label{sec:exam}
%loss functions and training setup
In this section, we showcase the resolution independence of RINO by solving several numerical examples, where the data is synthesized from PDE solutions under different setups, ranging from 1D operators to operators with different supports in their input and output functions. The setup for the data generation and operator learning task is specified for each problem separately in their respective sections.
Examples in \Cref{sec:antider,sec:darcy1d,sec:darcy2d,sec:burger} are chosen from the literature or designed to verify the resolution independence of the RI-DeepONet with respect to the input function data, since these simple functions do not require an advanced embedding of the output. Examples in Section~\ref{sec:darcy2dMesh} are designed to demonstrate the application of RINO in more challenging scenarios where the data realizations are obtained from finite element simulations of varying resolutions. In these cases, an embedding of the output is also beneficial. The compactness and consistency of the vectorial representation achieved through the proposed dictionary learning algorithm are discussed and studied in comparison with random projections in Appendix~~\ref{appx:gpod}. Moreover, the generalization, interpolation, and extrapolation capabilities of the proposed dictionary learning approach are discussed there. Additional potential advantages of RINO over RI-DeepONet in terms of generalization are examined in Appendix~\ref{appx:ri-vs-ri}.

%\subsection{Comment on data generation}
In cases where the PDE solver for data generation works on a fixed discretization, we randomly select only a subset of input function sensors for training to showcase the proposed method's ability to handle irregular data. The locations and numbers of these sensors vary from sample to sample. Assuming an input function realization is discretized at $M$ points, a corresponding ``random'' realization with random discretization is identified by randomly selecting $M_{\text{rand}}$ points without replacement, where 
$M_{\text{rand}}$ is independently drawn for each realization from the uniform distribution $M_{\text{rand}}\sim \text{Uniform}[M_{\text{min}}, M_{\text{max}}]$, and $0<M_{\text{min}}<M_{\text{max}}\le M$. %In the remainder of this section, when we label results as ``random'', we refer to this type of data. 

Since the input and output functions may have different scales across realizations, we use the relative mean square error (RMSE) as the training loss function. This approach is suggested to facilitate training and improve accuracy in operator learning problems \citep{wang2021learning,lu2021learning}. Consequently, all reported errors are based on the relative MSE measure, as follows:
\begin{equation}
    \text{Err}(v, v^{\text{pred}}) = 
    \mathbb{E}_{v^{}\sim\mu_{v}} \frac{\|v- v^{\text{pred}}\|_2^2}{\|v\|_{\infty}^2}.
\end{equation}

%\subsection{Comment on results}

\subsection{Verification Examples}
\subsubsection{ Example 1: Antiderivative}
\label{sec:antider}
The first verification example is taken from \cite{lu2021learning}. In this example, the data is generated according to the following antiderivative operator:
\begin{equation}
    \frac{ds}{dx} = u(x); \quad x\in [0, 1]
\end{equation}
where $u(x)$ is modeled as a Gaussian random process with zero mean and a covariance function defined by the radial basis function (RBF) with length scale $l=0.2$, i.e., $u(x)\sim \mathcal{GP}(0, \text{Cov}(x_1, x_2))$; details are provided in Appendix~\ref{appx:RFG}. We use the open-source dataset from \cite{lu2021learning}, which includes 150 pairs of input-output functions for the training set and 1000 pairs for the test set. In the original dataset, the functions are discretized at 100 equally spaced sensor points. Following the procedure described earlier, we set $M_{\text{min}} = 10$ and $M_{\text{max}} = 60$ to generate random point clouds for the training and testing cases. In other words, each training data function has a randomly subsampled discretization of between 10 and 60 points.

\begin{figure}[!ht]
  \centering
  \begin{subfigure}[b]{0.27\textwidth}
    \includegraphics[width=\textwidth]{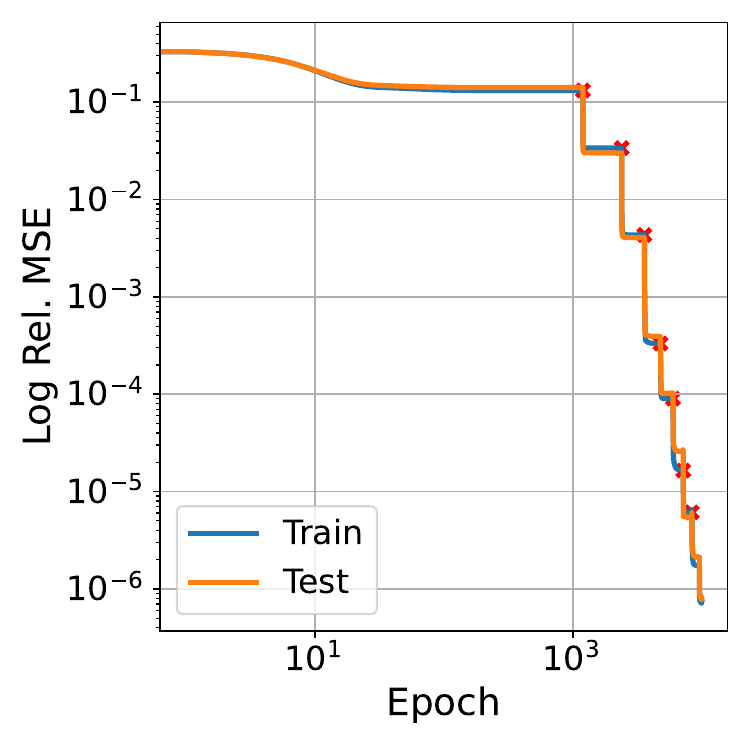}
    \caption{}
    %\label{subfig:one}
  \end{subfigure}
    \begin{subfigure}[b]{0.27\textwidth}
    \includegraphics[width=\textwidth]{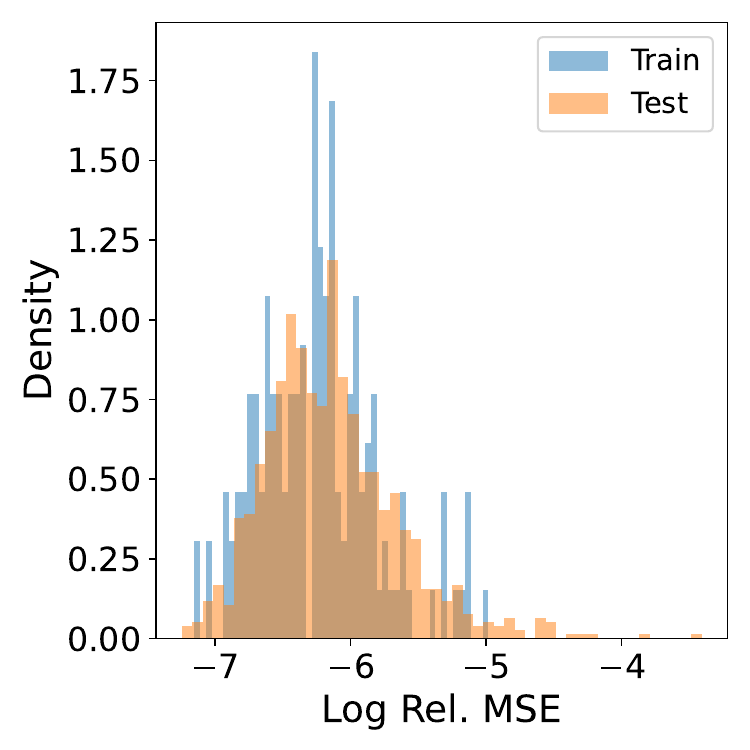}
    \caption{}
    %\label{subfig:one}
  \end{subfigure}
    \begin{subfigure}[b]{0.375\textwidth}
    \includegraphics[width=\textwidth]{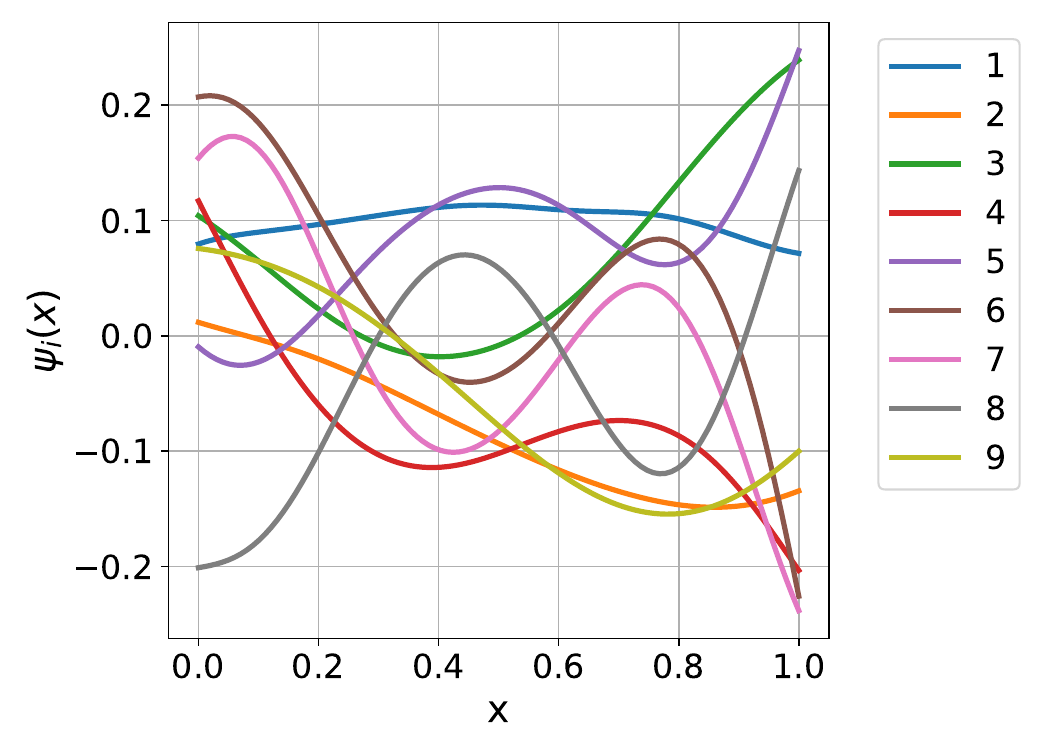}
    \caption{}
    %\label{subfig:one}
  \end{subfigure}
  \caption{Antiderivate Example: (a) Loss function during optimization iterations for dictionary learning of the input function bases. (b) Distribution of reconstruction errors for the train and test datasets after training. (c) The learned basis functions in the dictionary.
  % from rand 4
  }
  \label{fig:antider-loss-in}
\end{figure}

As described in the formulation section, the first step in the proposed approach is to learn a dictionary to represent the input functions (again, here we only encode the input). This dictionary will be used to find embedding coordinates for the operator learning task, specifically for use in the branch network of the RI-DeepONet. Figure~\ref{fig:antider-loss-in}(a) shows the reconstruction errors in Algorithm~\ref{algo:DL} during the ADAM \cite{kingma2014adam} iterations. Each red cross on the curve indicates the step at which a new basis function is added (line 8 in Algorithm~\ref{algo:DL}). With a total of 9 learned basis functions, the data can be reconstructed with high accuracy, as shown by the distribution of errors in Figure~\ref{fig:antider-loss-in}(b) for both training and testing datasets. The learned basis functions are plotted in Figure~\ref{fig:antider-loss-in}(c).

\begin{figure}[!ht]
  \centering
  \begin{subfigure}[b]{0.3\textwidth}
    \includegraphics[width=\textwidth]{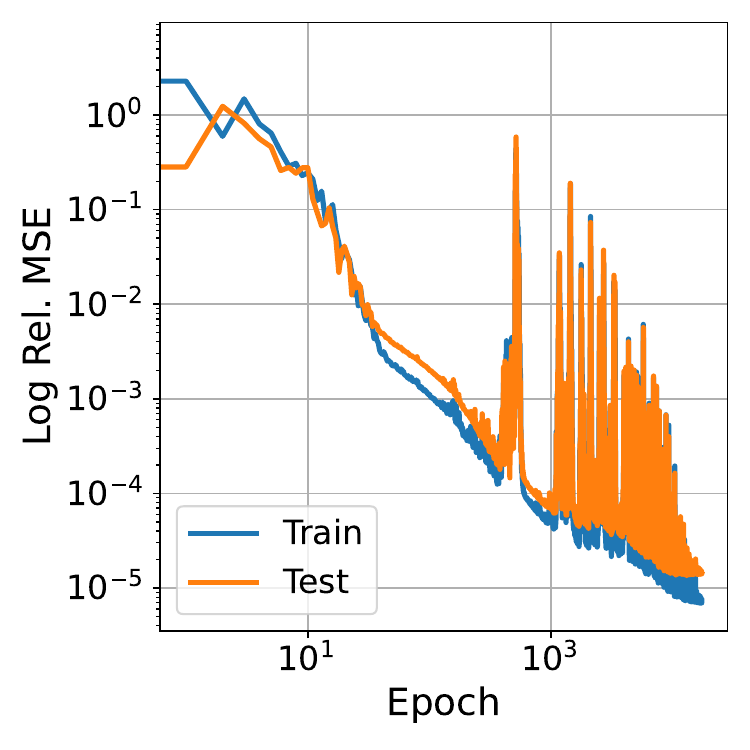}
    \caption{}
    %\label{subfig:one}
  \end{subfigure}
    \begin{subfigure}[b]{0.3\textwidth}
    \includegraphics[width=\textwidth]{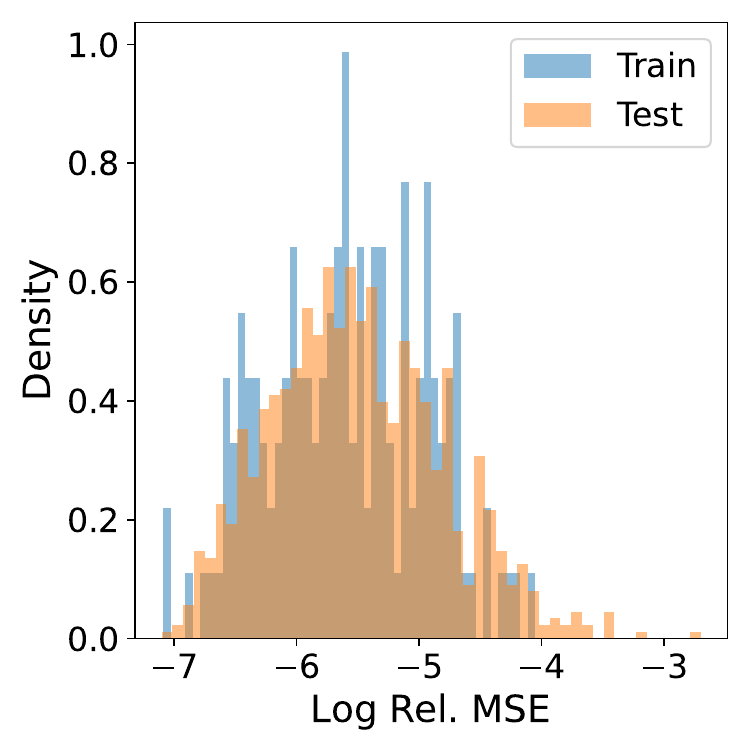}
    \caption{}
    %\label{subfig:one}
  \end{subfigure}
  \caption{Antiderivative Example: (a) Loss function during optimization iterations for operator learning. (b) Distribution of output function prediction errors for the training and testing datasets after training.
  % from rand 4
  }
  \label{fig:antider-loss-out}
\end{figure}

The ADAM iterations for operator learning using the input function embeddings obtained from the learned dictionary are shown in Figure~\ref{fig:antider-loss-out}(a). The close gap between the training and testing errors during the iterations indicates good generalization behavior, suggesting that the input function representation is informative. In Figure~\ref{fig:antider-loss-out}(b), the distribution of output function prediction errors after training is shown for both the training and testing sets.

\begin{figure}[!ht]
  \centering
  \begin{subfigure}[b]{0.9\textwidth}
    \includegraphics[width=\textwidth]{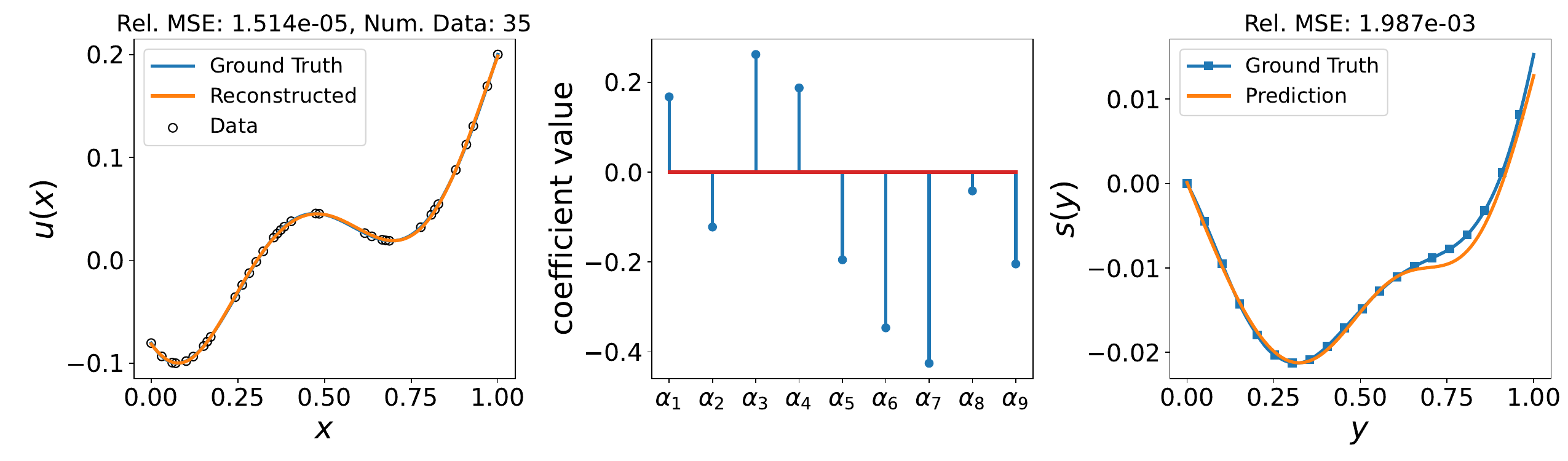}
    \caption{}
    %\label{subfig:one}
  \end{subfigure}
  %\hfill
  \begin{subfigure}[b]{0.9\textwidth}
    \includegraphics[width=\textwidth]{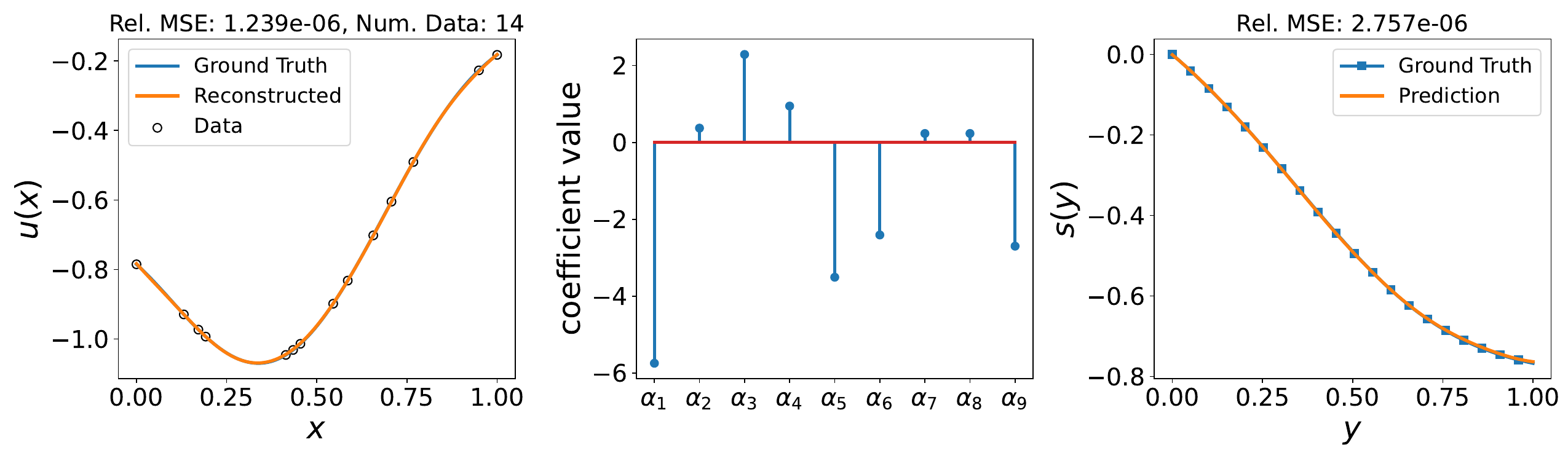}
    \caption{}
    %\label{subfig:two}
  \end{subfigure}
  \caption{Antiderivative Example: RI-DeepONet prediction for two queried input functions at inference and for the testing dataset: (a) Worst-case sample with the highest error in output function prediction, (b) Sample at the 25th percentile of the error distribution (i.e., 75\% of cases have less error than this value). The first column shows the queried input function data as `o' 
  % and the masked points (not provided to the dictionary) as `x', 
  along with the reconstruction from dictionary learning. The second column shows the projection coefficients of the queried input function data onto the learned dictionary of basis function from Figure~\ref{fig:antider-loss-in}(c). The last column shows the output function prediction along with the true solution.
  % from rand 4
  }
  \label{fig:antider-pred}
\end{figure}

In Figure~\ref{fig:antider-pred}, the prediction of RI-DeepONet at inference is shown for two queries from the test dataset. Figure~\ref{fig:antider-pred}(a) presents the results for the worst-case output function prediction. The first column shows the queried input function, which includes a point cloud with 35 random data points shown as `o' along with the reconstructed input from dictionary learning and the true function.
% along with the masked data points shown as `x' that are not provided to the dictionary but are used to assess the accuracy of the reconstruction. 
The second column shows the projection coefficients of the input function data onto the learned dictionary, which has a total of 9 basis functions. 
% The reconstruction using these coefficients is also plotted in the first figure. 
The last column shows the prediction of the output function along with the true model solution. Similarly, Figure~\ref{fig:antider-pred}(b) presents the same results for the test sample at the 25th percentile of the error distribution, meaning 75\% of test cases have lower error than this sample. As the results show, the RI-DeepONet method demonstrates good performance in providing predictions at inference for input function data recorded at an arbitrary (but sufficiently rich) number of sensors and locations.

\subsubsection{ Example 2: Nonlinear 1D Darcy's Equation}
\label{sec:darcy1d}
In this example, we increase the complexity of the previous problem by introducing a nonlinear operator. A variant of the nonlinear 1D Darcy's equation is given by the following form:
\begin{equation}
    \frac{ds}{dx}
    \left(
    -\kappa(s(x)) \frac{d s}{dx}
    \right)
     = u(x); \quad x\in[0, 1], 
\end{equation}
where the solution-dependent permeability is $\kappa(s(x)) = 0.2 + s^2(x)$ and the source term is a random field $u(x)\sim \mathcal{GP}$ with length scale $l=0.05$. Homogeneous Dirichlet boundary conditions $s=0$ are defined at the domain boundaries. The FEniCS finite element solver \cite{BarattaEtal2023} is used to generate data by discretizing the domain into 50 uniformly spaced nodal points. Following the procedure described earlier, we set $M_{\text{min}} = 20$ and $M_{\text{max}} = 35$ to generate random point clouds for the training and testing cases. The sizes of the training and test datasets are 800 and 200, respectively.

\begin{figure}[ht]
  \centering
  \begin{subfigure}[b]{0.9\textwidth}
    \includegraphics[width=\textwidth]{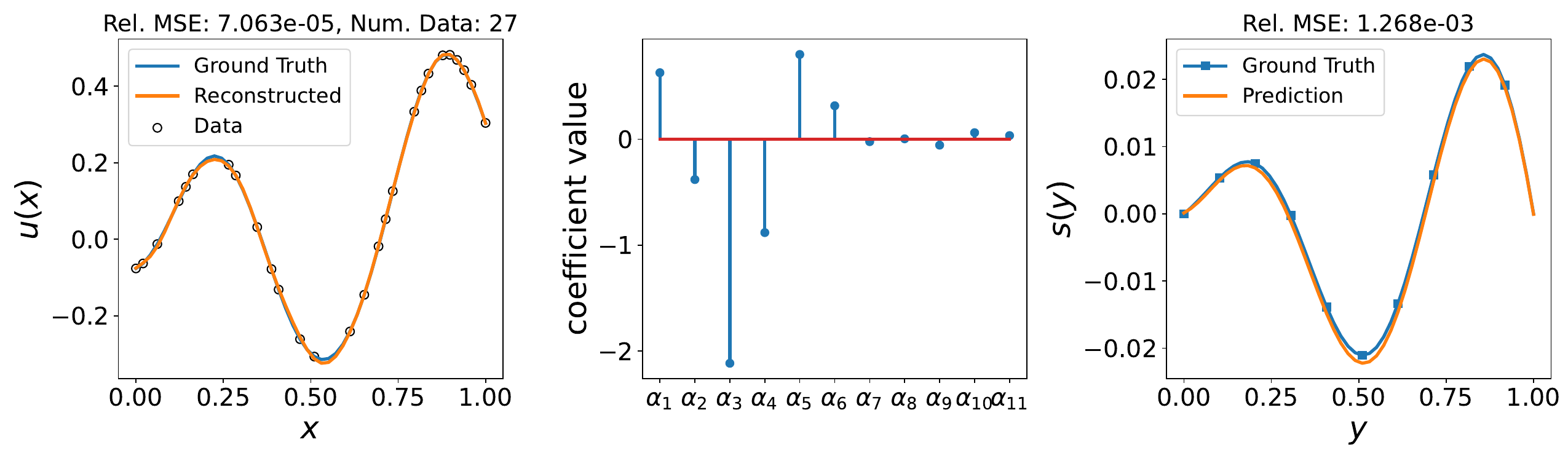}
    \caption{}
    %\label{subfig:one}
  \end{subfigure}
  %\hfill
  \begin{subfigure}[b]{0.9\textwidth}
    \includegraphics[width=\textwidth]{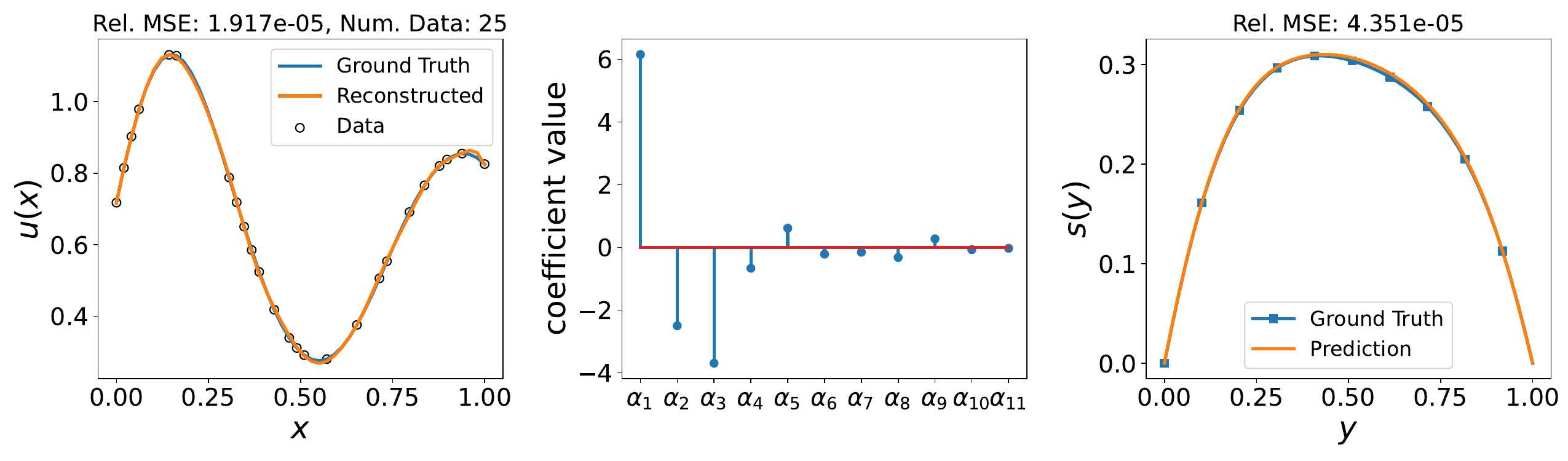}
    \caption{}
    %\label{subfig:two}
  \end{subfigure}
  \caption{1D Nonlinear Darcy's Equation: RI-DeepONet Prediction for two queried input functions at inference and for the testing dataset: (a) Worst-case sample with the highest error in output function prediction, (b) Sample at the 25th percentile of the error distribution (i.e., 75\% of cases have less error than this value). The first column shows the queried input function data as `o' 
  % and the masked points (not provided to the dictionary) as `x', 
  along with the reconstruction from dictionary learning. The second column shows the projection coefficients of the queried input function data onto the learned dictionary of basis function from Figure~\ref{fig:antider-loss-in}(c). The last column shows the output function prediction along with the true solution.
  % from rand 3
  }
  \label{fig:darcy-nonlin-pred}
\end{figure}

The worst-case RI-DeepONet predictions are shown in Figure~\ref{fig:darcy-nonlin-pred}(a), demonstrating visually acceptable accuracy for both the output function prediction and the input function reconstruction. Similarly the test sample at the 25th percentile of the error distribution is shown in Figure~\ref{fig:darcy-nonlin-pred}(b).

\subsubsection{ Example 3: Nonlinear 2D Darcy's Equation}
\label{sec:darcy2d}
In this example, we extend the previous problem to a two-dimensional setting. The nonlinear Darcy's equation in two dimensions is given by:
\begin{equation}
    \nabla \cdot\left(-\kappa(s(\boldsymbol{x})) \nabla s\right)
     = u(\boldsymbol{x}); \quad \boldsymbol{x}\in[0, 1]^2, 
\end{equation}
where the solution-dependent permeability is $\kappa(s(\boldsymbol{x})) = 0.2 + s^2(\boldsymbol{x})$ and the source term is a random field $u(\boldsymbol{x})\sim \mathcal{GP}$ with length scale $l=0.25$. Homogeneous Dirichlet boundary conditions $s=0$ are defined at the domain boundaries. The solution field is obtained via the finite element method. Input function random fields are discretized on a $20\times20$ uniform grid, while the triangulation of the finite element domain does not necessarily coincide with these input sensor point locations. Following the procedure described earlier, we set $M_{\text{min}} = 100$ and $M_{\text{max}} = 280$ to generate random point clouds over a subset of the 400 grid points for the training and testing cases. The sizes of the training and test datasets are 800 and 200, respectively.

\begin{figure}[!ht]
  \centering
    \begin{subfigure}[b]{0.75\textwidth}
    \includegraphics[width=\textwidth]{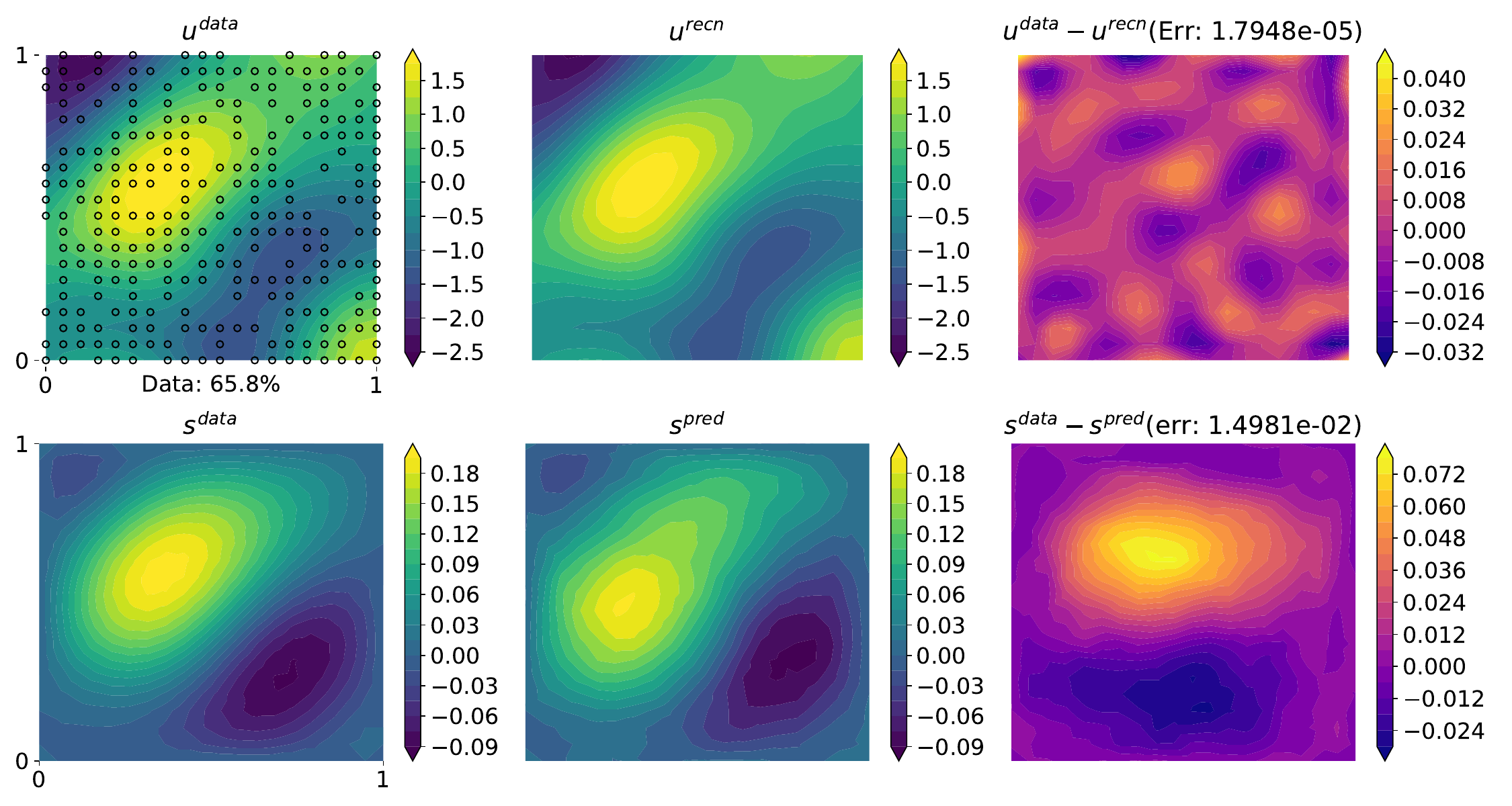}
    %\caption{}
    %\label{subfig:one}
  \end{subfigure}
  \caption{
  2D Nonlinear Darcy's Equation: RI-DeepONet prediction for the worst-case output function  $s(\boldsymbol{y})$ in the test dataset. The top left figure shows the queried input function data $u(\boldsymbol{x})$, with black circle symbols indicating the queried data. The middle top figure shows the input field reconstruction via the learned dictionary. The top right figure shows the error between the reconstructed input field and the ground truth. The bottom figures show the output field $s(\boldsymbol{y})$, the RI-DeepONet prediction, and its corresponding errors.
  % from rand 0
  }
  \label{fig:darcy2d-nonlin-pred-Worst}
\end{figure}

The RI-DeepONet predictions corresponding to the worst-case test error for the output field is shown in  \cref{fig:darcy2d-nonlin-pred-Worst}, demonstrating satisfactory error. 
% While RINO's accuracy is relatively satisfactory, it could be improved by conducting a rigorous hyperparameter search to find a better neural network configuration.

\subsubsection{ Example 4: Nonlinear Burgers' Equation}
\label{sec:burger}
In this numerical example, we examined an initial boundary value problem in which the input and output functions of the operator have different domains and dimensions. Specifically, we looked at the parametric nonlinear Burgers' equation, as introduced in \cite{wang2021learning}, with the random field initial condition $s(x)|_{t=0}$ as the input function and the space-time solution $s(x, t)$ as the output function:
\begin{align}
    &\frac{\partial s}{\partial t} + s \frac{\partial s}{\partial x}
    =
    \nu \frac{\partial^2  s}{\partial x^2}; \quad (x,t)\in [0, 1]^2,\\
    & s|_{t=0} = u(x),
\end{align}
where periodic boundary conditions are assumed, and the periodic initial condition is a Gaussian random process with parameters similar to \cite{wang2021learning}, and 
$\nu=0.01$. A spectral solver is used to obtain the solution field on a $101\times100$ grid in the space-time domain for each input function, which is discretized on 101 uniformly spaced points. Following the procedure described earlier, we set $M_{\text{min}} = 40$ and $M_{\text{max}} = 70$ to generate random point clouds for the training and testing cases. The sizes of the training and test datasets are 1500 and 500, respectively. Note that the input function samples are randomly discretized; however, the predictions are on the full solution grid for the results in this section, similar to other examples.

\begin{figure}[!ht]
  \centering
  \begin{subfigure}[b]{0.75\textwidth}
    \includegraphics[width=\textwidth]{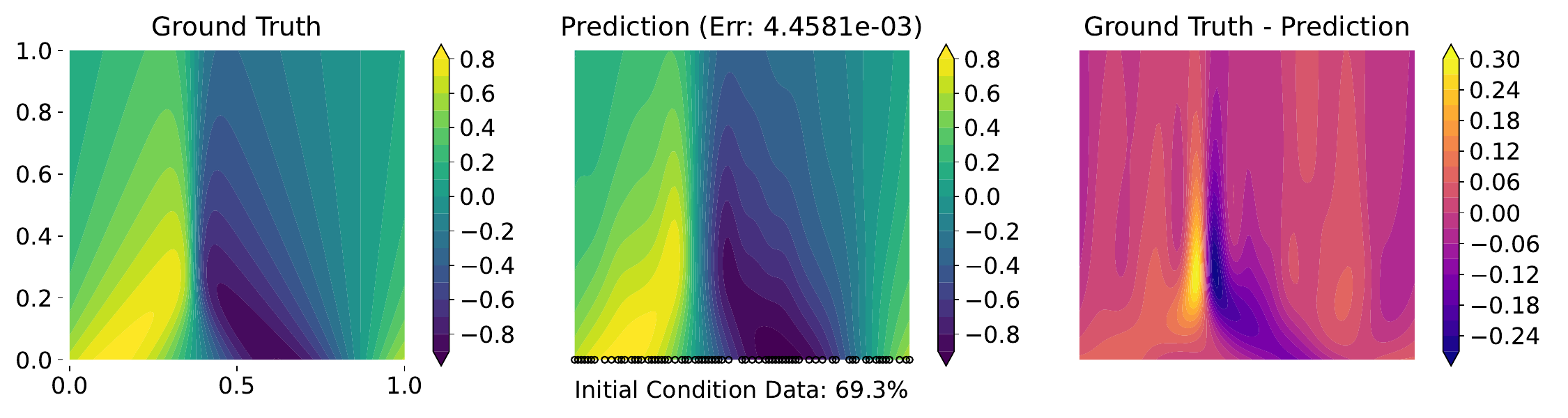}
    \caption{}
    %\label{subfig:one}
  \end{subfigure}
  %\hfill
  \begin{subfigure}[b]{0.75\textwidth}
    \includegraphics[width=\textwidth]{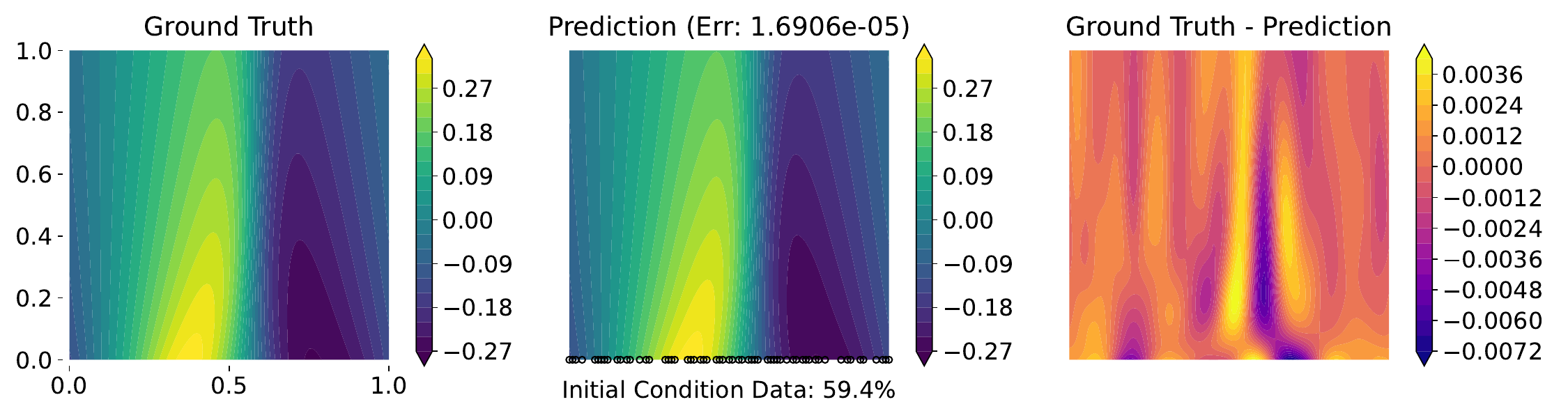}
    \caption{}
    %\label{subfig:two}
  \end{subfigure}
  \caption{2D Burger's Equation: RI-DeepONet prediction for two queried initial condition fields at inference and for the testing dataset: (a) Sample with the highest error in output function prediction $s(\boldsymbol{y})$, and (b) Sample at the 25th percentile of the error distribution (i.e., 75\% of cases have less error than this value). The first column shows the ground truth solution. The second column shows the RI-DeepONet prediction with the queried input initial condition points shown along the x-axis. The last column shows the point-wise error. The output domain $\boldsymbol{y}=(x,t)$ has spatio-temporal coordinates, where the horizontal axis represents space and the vertical axis represents time. 
  % The red cross symbols in the middle column indicate the masked locations of the queried initial condition.
  % from rand 1
  }
  \label{fig:Burgers-pred}
\end{figure}

To showcase the flexibility of the RI-DeepONet framework, we use predefined trunk basis vectors as the first 70 POD bases obtained from the solution output snapshots of the training dataset (making it effectively RI-POD-DeepONet). It has been shown previously that using predefined optimal bases can facilitate the training of DeepONet and achieve faster convergence \cite{lu2022comprehensive}. Two samples from the testing set are shown in Figure~\ref{fig:Burgers-pred}.

\subsubsection{Discussion}
After training each of the models listed above on the defined randomly sub-sampled data, we can perform inference during testing using any combination of input function sensors and at any desired points, thanks to the fully continuous nature of the RI-DeepONet for processing both input and output functions. To examine the influence of discretization, we perform additional tests on regularly subsampled sensors for the input function at inference. A summary of the results is provided in Table~\ref{tab:all-result}. Rows in this table with a numeric value for the number of sensors (e.g., $M=100$) indicate performance during inference for input functions queried at structured and uniformly spaced sensor locations. Rows in the table denoted `random,' correspond to test sets on the randomly sub-sampled discretizations described in Sections 3.1.1 -- 3.1.4. The RI-DeepONet used for predictions is trained using a data set of randomly sub-sampled points as described for each problem in Sections 3.1.1 -- 3.1.4. Therefore, training error is reported only for the `random' case.
%Note that we train only on randomly created point clouds, which is why we do not report training error at regular points in Table \ref{tab:all-result}.

The error statistics shown in Table~\ref{tab:all-result} are computed for the trained models described above. Each neural network model is randomly initialized before training, and the sensor locations for the input functions are also randomly generated for each training run.
%Rows in the tables with a numeric value for the number of sensors indicate performance during inference for input functions queried at structured and uniformly spaced sensor locations. For example, in the forth row of Table~\ref{tab:all-result}, $M=100$ sensor locations are used. 

As the results suggest, the output function prediction is nearly insensitive to the discretization resolution of the queried input function for all PDE cases. However, in the 1D nonlinear Darcy's and nonlinear Burgers' equations, we observe a degradation in accuracy after some level of resolution coarsening (see the last row for each case). This occurs because, as anticipated, when the resolution falls below a certain threshold—particularly when it is significantly less than the problem's characteristic length scale—the predictions become erroneous and thus resolution-dependent. In other words, these correspond to cases where a ReNO cannot be accurately identified due to under-sampling.

\begin{table}[h]
\centering
\begin{tabular}{@{}lccc@{}}
\toprule
     & Number of Sensors ($M$) & Test Error (mean $\pm$ std)  & Train Error \\ 
\midrule
 \textbf{Antiderivative}\\
    & random & 1.751e-05 $\pm$ 3.398e-06 & 7.044e-06 $\pm$ 4.122e-07 \\
    & 100 & 1.456e-05 $\pm$ 3.126e-06 & -  \\
    & 51 & 1.198e-05 $\pm$ 1.439e-06 & -  \\
    & 26 & 1.107e-05 $\pm$ 4.244e-07 & -  \\
    & 21 & 1.104e-05 $\pm$ 3.037e-07 & -  \\
    & 11 & 1.238e-05 $\pm$ 6.151e-07 & - \\
  \midrule
     \textbf{Nonlinear Darcy (1D)}\\
     & random & 3.630e-05 $\pm$ 7.164e-06 & 1.390e-05 $\pm$ 1.628e-06 \\
     & 50 & 2.989e-05 $\pm$ 5.290e-06 & -  \\
     & 26 & 3.018e-05 $\pm$ 5.431e-06 & -  \\
     & 11 & 4.721e-05 $\pm$ 7.422e-06 & -  \\
     & 6 & 8.347e-03 $\pm$ 5.425e-03 & - \\
  \midrule
   \textbf{Nonlinear Darcy (2D)}\\
    & random & 1.121e-03 $\pm$ 1.456e-04 & 9.261e-04 $\pm$ 2.344e-04 \\
    & $20\times20$ & 1.135e-03 $\pm$ 1.423e-04 & -  \\
    & $10\times10$ & 1.135e-03 $\pm$ 1.394e-04 & -  \\
    & $4\times4$ & 1.368e-03 $\pm$ 2.039e-04 & -  \\
  \midrule
   \textbf{Burgers}\\
    & random & 3.795e-05 $\pm$ 1.986e-06 & 1.361e-05 $\pm$ 9.999e-07 \\
    & 101 & 3.789e-05 $\pm$ 1.989e-06 & -  \\
    & 51 & 3.788e-05 $\pm$ 1.991e-06 & -  \\
    & 21 & 3.790e-05 $\pm$ 1.986e-06 & -  \\
    & 11 & 7.011e-04 $\pm$ 3.537e-04 & -  \\
  \bottomrule                          
\end{tabular}
\caption{Resolution dependence study for each verification problem. RI-DeepONet output function prediction errors are reported for queried input functions at various resolutions denoted by the Number of Sensors ($M$). Cases where a numerical value is prescribed correspond to fixed regular discretizations of the input. Cases denoted `random' indicate a random subsampling of the input discretization as defined in Sections 3.1.1 -- 3.1.4. Mean training/test errors and standard deviations are computed from five independent trials.}
\label{tab:all-result}
\end{table}

\subsection{Demonstration Example 1: Nonlinear 2D Darcy’s Equation on Unstructured Meshes}
\label{sec:darcy2dMesh}
In this problem, we revisit the 2D Darcy flow problem previously introduced in Section~\ref{sec:darcy2d}. The main difference is that, in this example, the data belongs to unstructured finite element meshes that differ from realization to realization. The goal here is not to showcase the method's consistency across varying resolutions (from low to high or vice versa), which was studied in previous problems. Instead, we aim to demonstrate that the RINO can perform operator learning for problems of practical interest involving arbitrary finite element meshes.
% the problem under the configuration where there are many different finite element solutions with varying meshes, and we are interested in conducting an operator learning task under such data.

To generate training data, we purposely generate unstructured triangular meshes with varying density per realization. The areas with higher mesh density are arbitrary and do not correspond, for example, to areas where the mesh might be refined based on an error estimator. We then interpolate the previously generated random input fields (see Section~\ref{sec:darcy2d}) onto the new unstructured meshes and obtain the solution field for each realization using a nonlinear finite element solver written in FEniCS \cite{BarattaEtal2023}.

\begin{figure}[!ht]
  \centering
  \begin{subfigure}[b]{0.3\textwidth}
    \includegraphics[width=\textwidth]{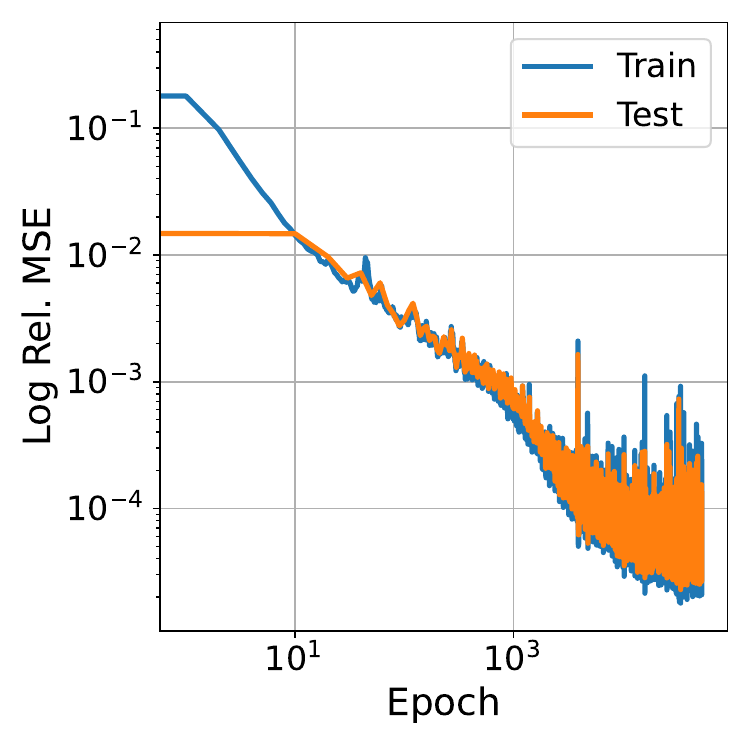}
    \caption{}
    %\label{subfig:one}
  \end{subfigure}
    \begin{subfigure}[b]{0.3\textwidth}
    \includegraphics[width=\textwidth]{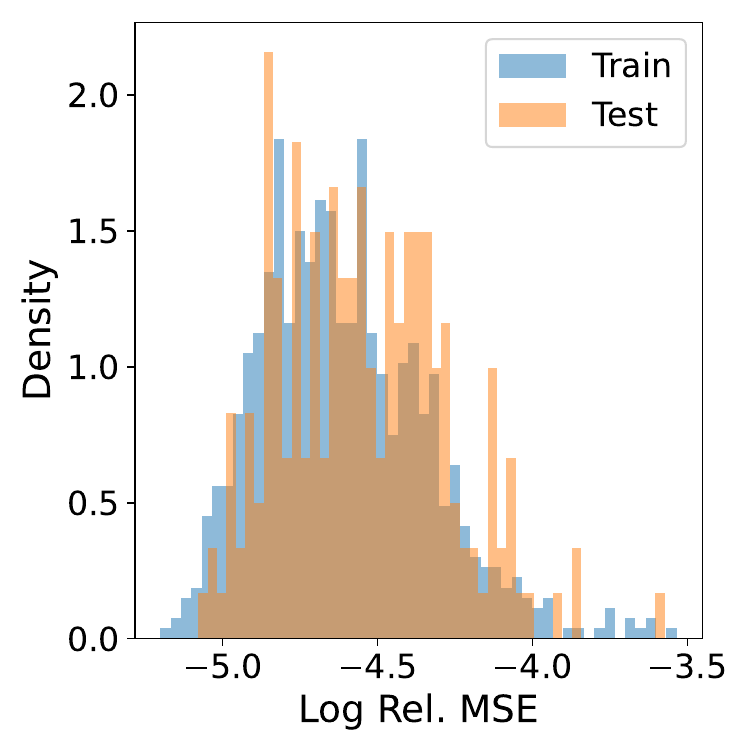}
    \caption{}
    %\label{subfig:one}
  \end{subfigure}
  \caption{2D Darcy's Equation on Unstructured Meshes with RI-DeepONet: (a) Loss function during optimization iterations for operator learning. (b) Distribution of output function prediction errors for the training and testing datasets after training.
  % from rand 3
  }
  \label{fig:darcy2dMesh-loss-out}
\end{figure}

\paragraph{Operator Learning with RI-DeepONet.}
Similar to the previous examples, using Algorithm~\ref{algo:DL}, we learn 40 INR basis functions to effectively project an arbitrary input function point cloud onto the dictionary subspace with a satisfactory reconstruction error (see Appendix~\ref{appdx:darcy2dMesh}). We then use this 40-dimensional latent representation for operator learning where the trunk and branch neural networks are unknown \textit{a priori}.  The evolution of the loss function during the operator learning stage is depicted in Figure~\ref{fig:darcy2dMesh-loss-out}(a). The distribution of output function prediction errors after training is plotted in Figure~\ref{fig:darcy2dMesh-loss-out}(b).

\begin{figure}[!ht]
  \centering
    \begin{subfigure}[b]{0.8\textwidth}
    \includegraphics[width=\textwidth]{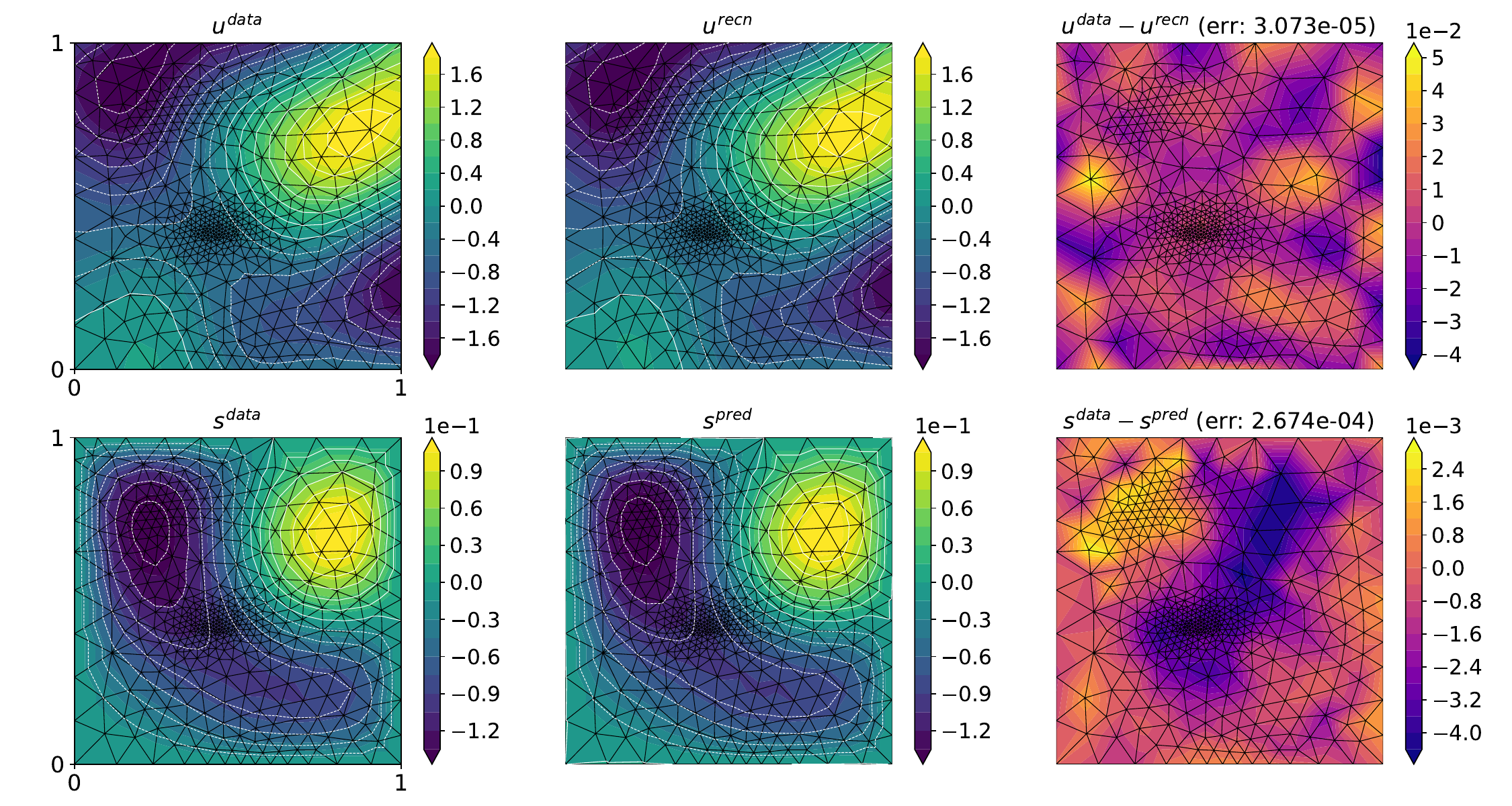}
    %\caption{}
    %\label{subfig:one}
  \end{subfigure}
  \caption{
    2D Nonlinear Darcy's Equation (Unstructured Mesh): RI-DeepONet prediction with the worst-case output function, $s(\boldsymbol{y})$, in the test dataset. The top left figure shows the queried input function data $u(\boldsymbol{x})$. The middle top figure shows the input field reconstruction via learned dictionary. The top right figure shows the error between the reconstructed input field and the ground truth. The bottom figures show the true output field $s(\boldsymbol{y})$, the RI-DeepONet prediction, and its corresponding errors.
  }
  \label{fig:darcy2dMesh-nonlin-pred-Worst}
\end{figure}

\begin{figure}[!ht]
  \centering
    \begin{subfigure}[b]{0.8\textwidth}
    \includegraphics[width=\textwidth]{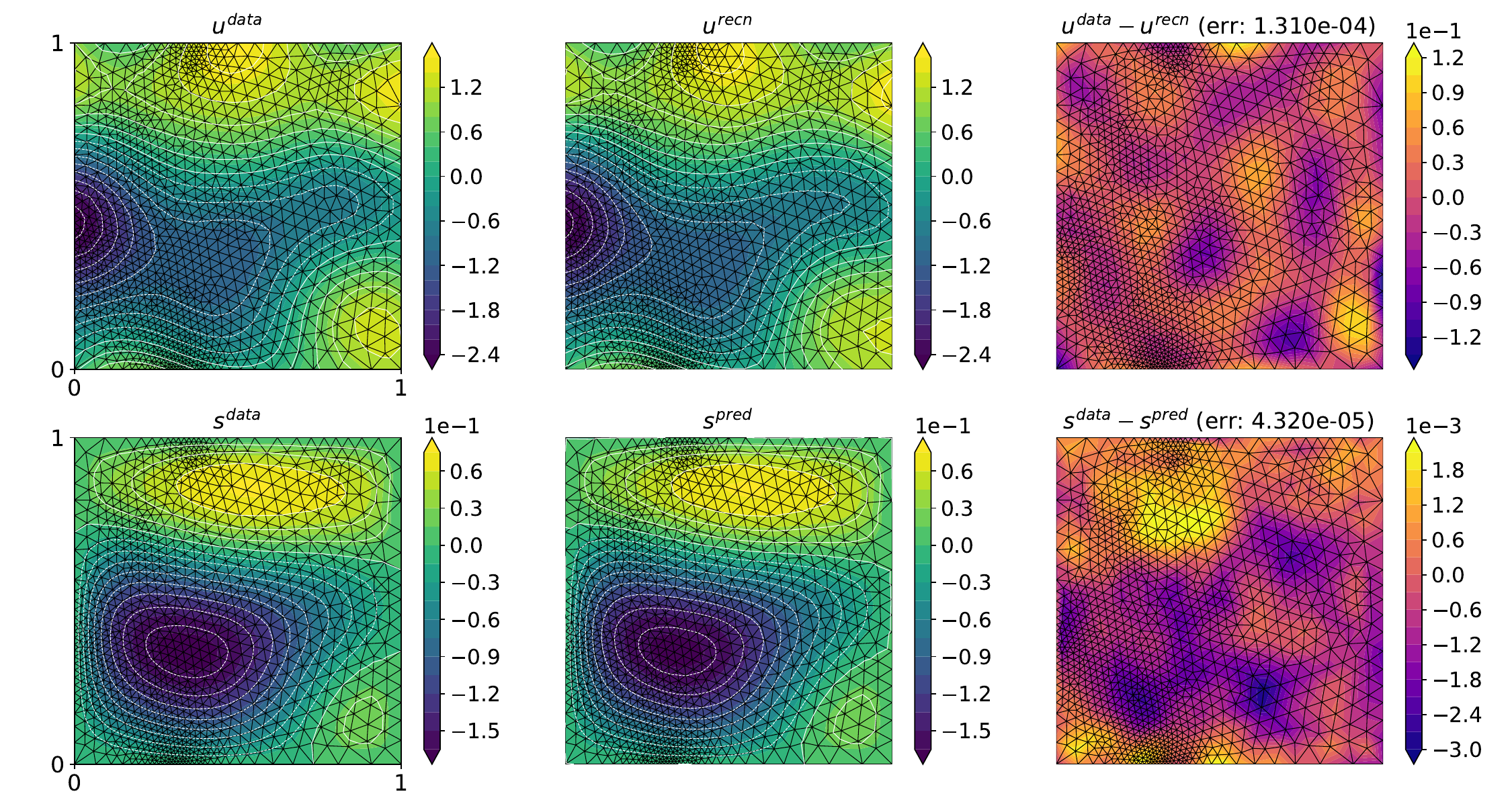}
    %\caption{}
    %\label{subfig:one}
  \end{subfigure}
  \caption{
    2D Nonlinear Darcy's Equation (Unstructured Mesh): RI-DeepONet results for the test sample with an output function prediction error in the 25th percentile of the error distribution.
    The top left figure shows the queried input function data $u(\boldsymbol{x})$. The middle top figure shows the input field reconstruction via learned dictionary. The top right figure shows the error between the reconstructed input field and the ground truth. The bottom figures show the true output field $s(\boldsymbol{y})$, the RI-DeepONet prediction, and its corresponding errors.
  }
  \label{fig:darcy2dMesh-nonlin-pred-q25}
\end{figure}

The worst-case output function prediction and the case corresponding to the 25th percentile of the error distribution from the unseen test data are shown in ~\cref{fig:darcy2dMesh-nonlin-pred-Worst,fig:darcy2dMesh-nonlin-pred-q25}, respectively. The results indicate a satisfactory training process and prediction capability.

\paragraph{Operator Learning with }
%As mentioned in the previous problem, one can improve the training process of the DeepONet architecture by bypassing the need for learning the trunk basis functions, instead of jointly learning branch and trunk neural networks \cite{lu2022comprehensive}. The early idea utilized POD basis functions learned from the output functions and used them as fixed, predefined trunk basis functions. However, using vanilla POD only works when the output functions are discretized at fixed locations, hence bringing resolution dependency issues again.
As discussed in Section~\ref{sec:OL-BrOnly}, the proposed dictionary learning algorithms can be applied in a similar manner for learning appropriate basis functions for the output function data. Once the dictionary is learned (\textit{offline}), it can be utilized as a predefined basis for the output. Under such a setup, we only need to train a single neural network between embedding spaces as a neural operator. In this problem, we utilize Algorithm~\ref{algo:DL-sbs} to learn a dictionary of 37 INR basis functions that can sufficiently approximate the output function data. The full error distributions of the learned output basis functions for the training and test data are provided in Appendix~\ref{appdx:darcy2dMesh}.
%; the error distribution of this dictionary learning task is shown in Appendix~\ref{appdx:darcy2dMesh}.

\begin{figure}[!ht]
  \centering
  \begin{subfigure}[b]{0.3\textwidth}
    \includegraphics[width=\textwidth]{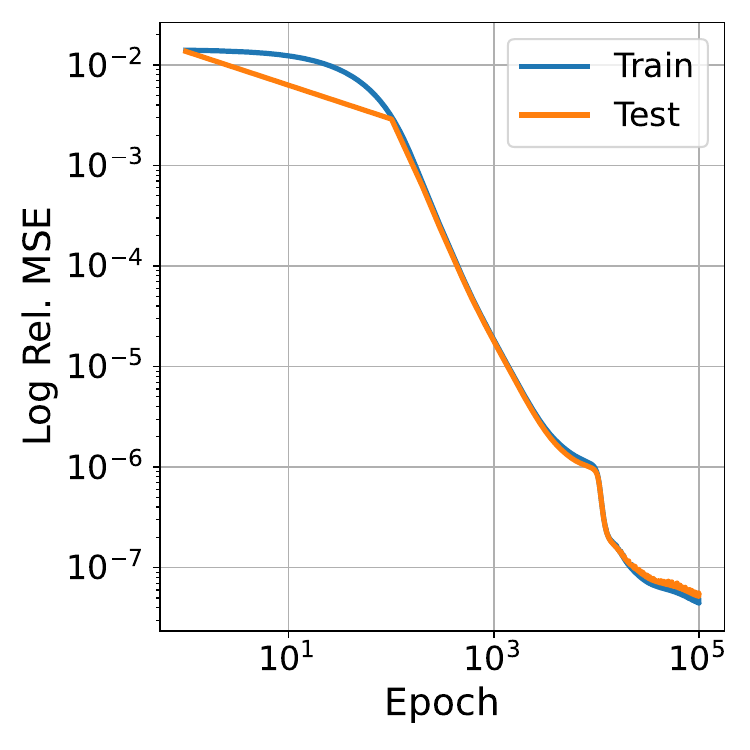}
    \caption{}
    %\label{subfig:one}
  \end{subfigure}
    \begin{subfigure}[b]{0.3\textwidth}
    \includegraphics[width=\textwidth]{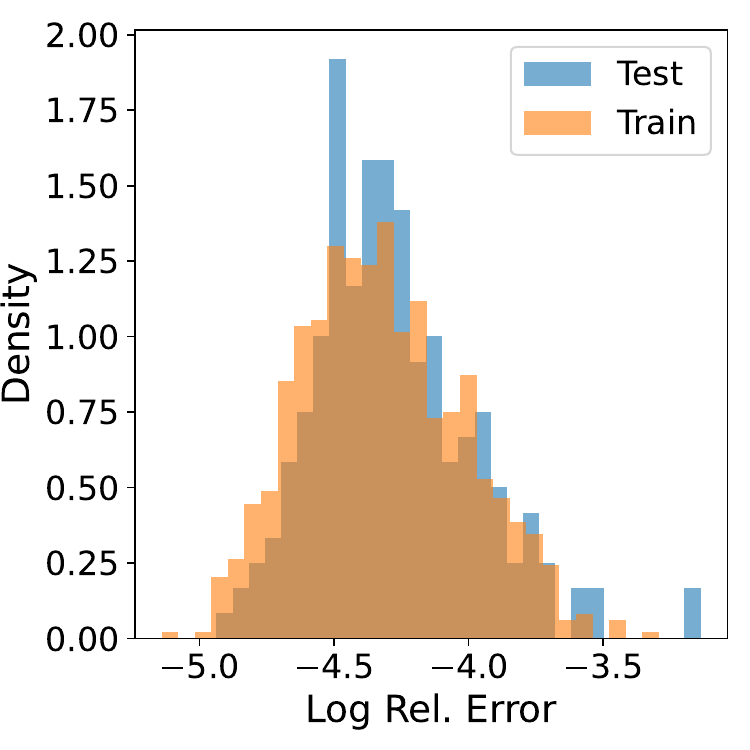}
    \caption{}
    %\label{subfig:one}
  \end{subfigure}
  \caption{2D Darcy's Equation on Unstructured Meshes with RINO: (a) Loss function during optimization iterations for operator learning. (b) Distribution of output function prediction errors for the training and testing datasets after training.
  }
  \label{fig:darcy2dMesh-brOnly-loss-out}
\end{figure}

The evolution of the loss function during the operator learning stage for the RINO case where we utilized the learned output basis dictionary is depicted in Figure~\ref{fig:darcy2dMesh-brOnly-loss-out}(a). The distribution of output function prediction errors after training is plotted in Figure~\ref{fig:darcy2dMesh-brOnly-loss-out}(b).

\begin{figure}[!ht]
  \centering
    \begin{subfigure}[b]{0.8\textwidth}
    \includegraphics[width=\textwidth]{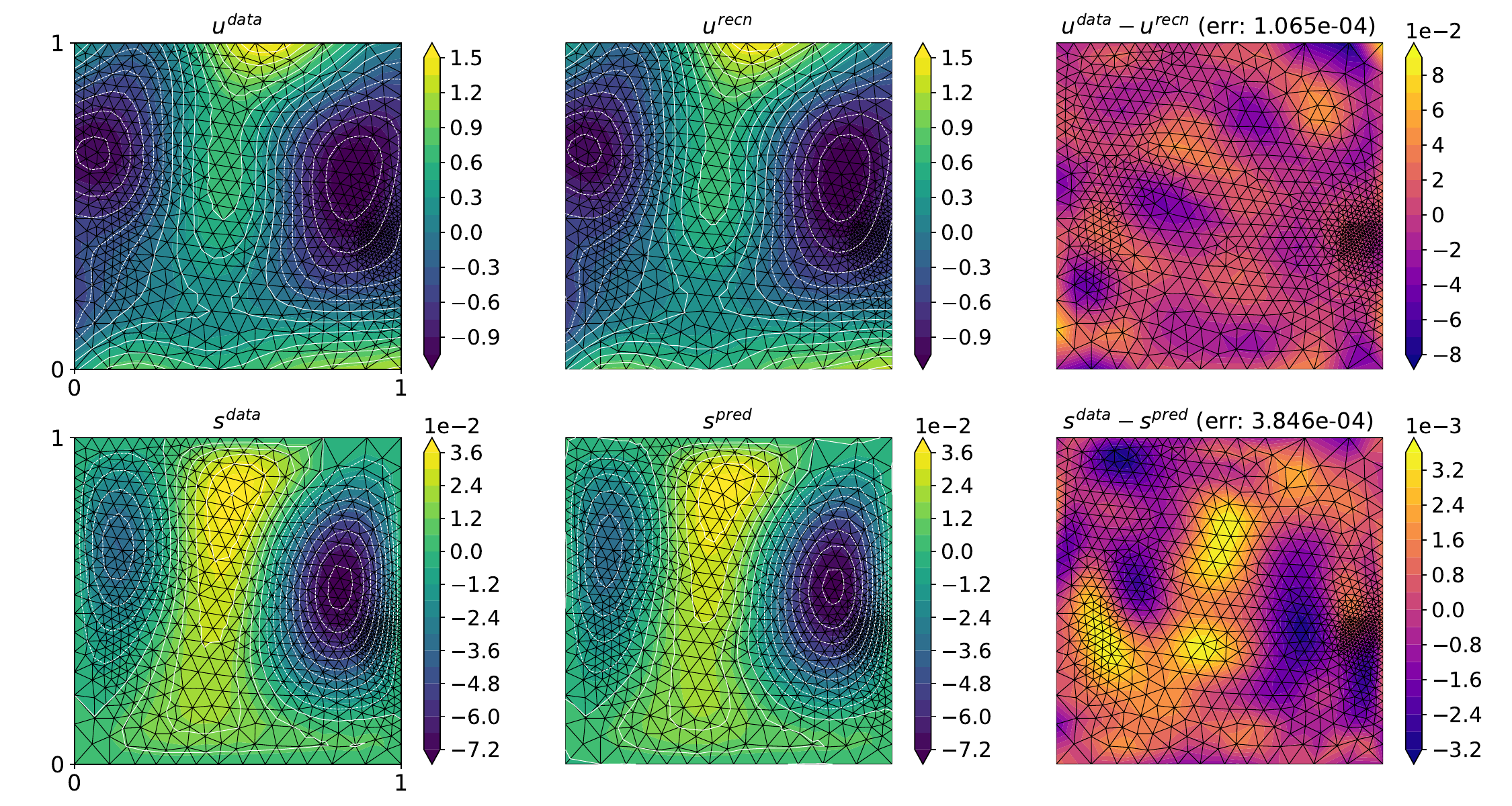}
    %\caption{}
    %\label{subfig:one}
  \end{subfigure}
  \caption{
    2D Nonlinear Darcy's Equation (Unstructured Mesh) with Predefined INR Output Basis: RINO prediction with the worst-case output function, $s(\boldsymbol{y})$, in the test dataset. The top left figure shows the queried input function data $u(\boldsymbol{x})$. The middle top figure shows the input field reconstruction via learned dictionary. The top right figure shows the error between the reconstructed input field and the ground truth. The bottom figures show the true output field $s(\boldsymbol{y})$, the RINO prediction, and its corresponding errors.
  }
  \label{fig:darcy2dMesh-nonlin-brOnly-pred-Worst}
\end{figure}

\begin{figure}[!ht]
  \centering
    \begin{subfigure}[b]{0.8\textwidth}
    \includegraphics[width=\textwidth]{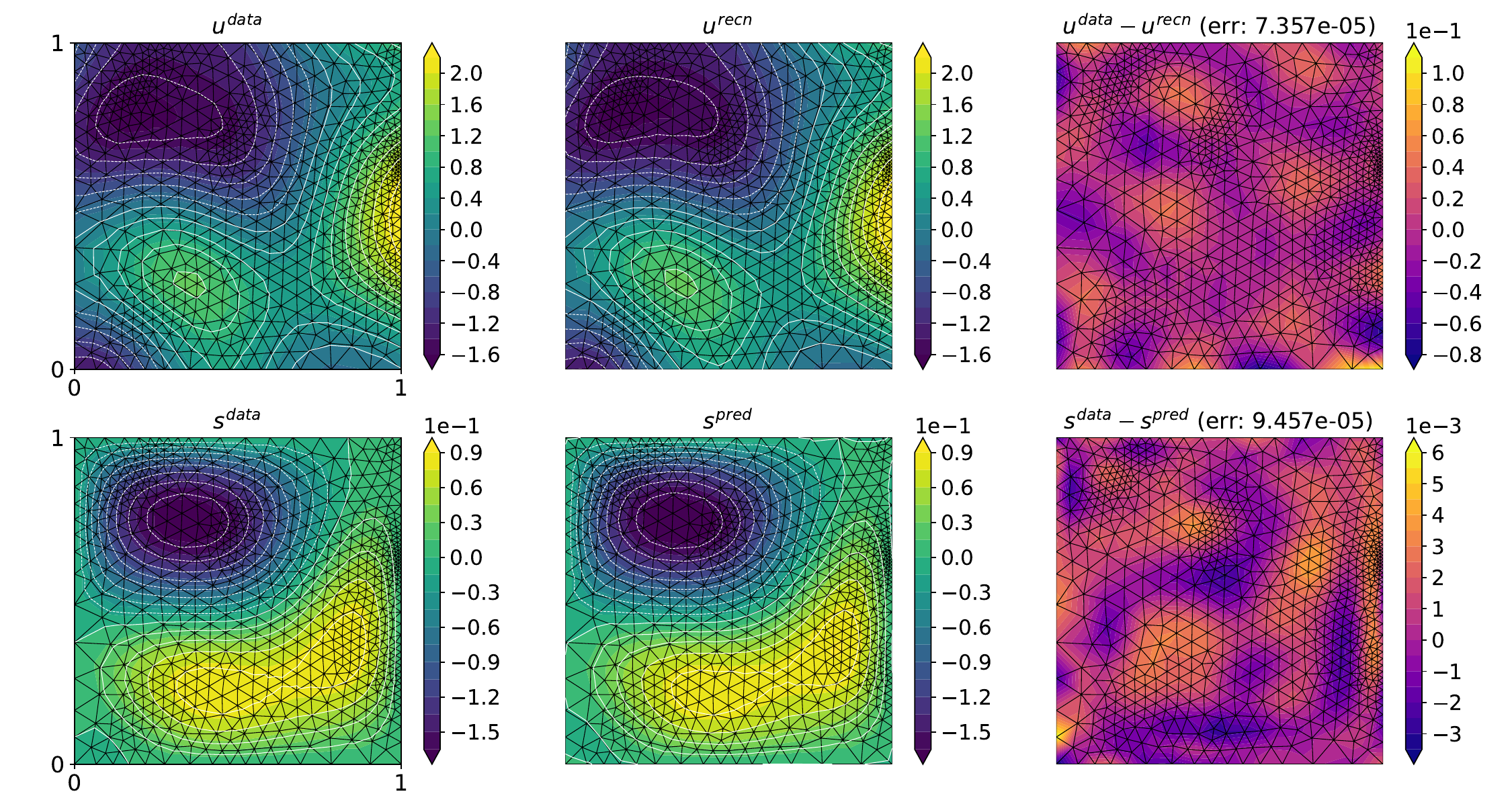}
    %\caption{}
    %\label{subfig:one}
  \end{subfigure}
  \caption{
    2D Nonlinear Darcy's Equation (Unstructured Mesh) with Predefined INR Output Basis: RINO results for the test sample with an output function prediction error in the 25th percentile of the error distribution.
    The top left figure shows the queried input function data $u(\boldsymbol{x})$. The middle top figure shows the input field reconstruction via learned dictionary. The top right figure shows the error between the reconstructed input field and the ground truth. The bottom figures show the true output field $s(\boldsymbol{y})$, the RINO prediction, and its corresponding errors.
  }
  \label{fig:darcy2dMesh-nonlin-brOnly-pred-q25}
\end{figure}

The worst-case output function prediction and the case corresponding to the 25th percentile of the error distribution from the unseen test data are shown in ~\cref{fig:darcy2dMesh-nonlin-brOnly-pred-Worst,fig:darcy2dMesh-nonlin-brOnly-pred-q25}, respectively. A successful training with satisfactory accuracy is confirmed by the results for the RINO case where we indirectly learned a mapping between the coefficients of two sets of weakly orthogonal basis functions: one corresponding to the input function data and the other to the output function data, which were identified separately and \textit{offline}.

From these results and the error plots presented for the case RI-DeepONet, we see that the RI-DeepONet and RINO results have comparable accuracy. However, the RINO has the distinct advantage of learning the operator in the compact embedding space such that the input and output bases do not have redundancies (due to their orthogonality). The result of this is that the neural operator itself is much more compact -- having far fewer parameters. Where the RI-DeepONet uses MLP with layers \texttt{[40, 50, 100]}, the RINO uses MLP with layers \texttt{[40, 40, 37]} to achieve comparable accuracy (see Appendix \ref{appx:nn-config}). Other potential advantages of RINO over RI-DeepONet, particularly in terms of improved generalization capability, are also discussed in Appendix~\ref{appx:ri-vs-ri}.

\subsection{Demonstration Example 2: Solid Mechanics}
\label{sec:solid-mech}

\begin{figure}[!ht]
  \centering
  \begin{subfigure}[b]{0.3\textwidth}
    \includegraphics[width=\textwidth]{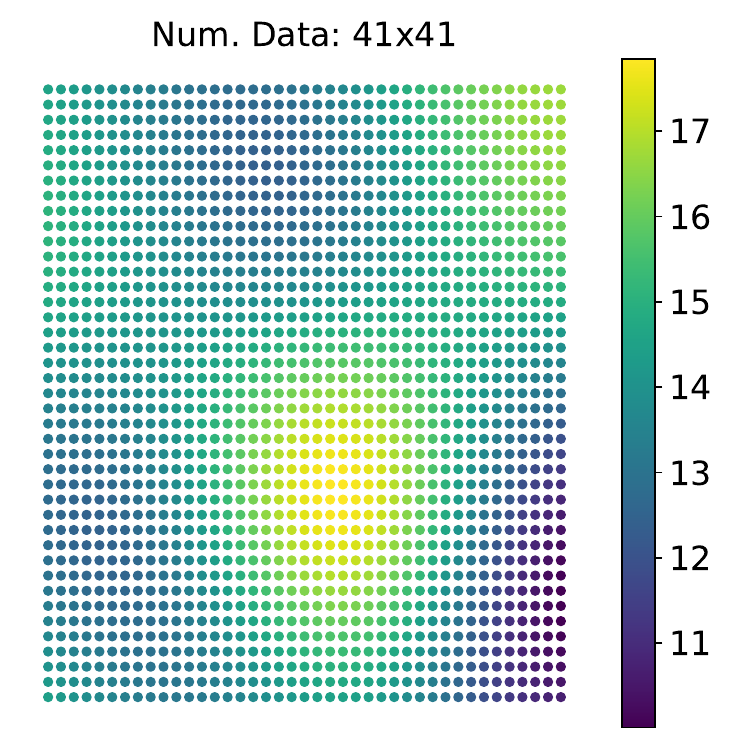}
    % \caption{}
    %\label{subfig:one}
  \end{subfigure}
  \begin{subfigure}[b]{0.3\textwidth}
    \includegraphics[width=\textwidth]{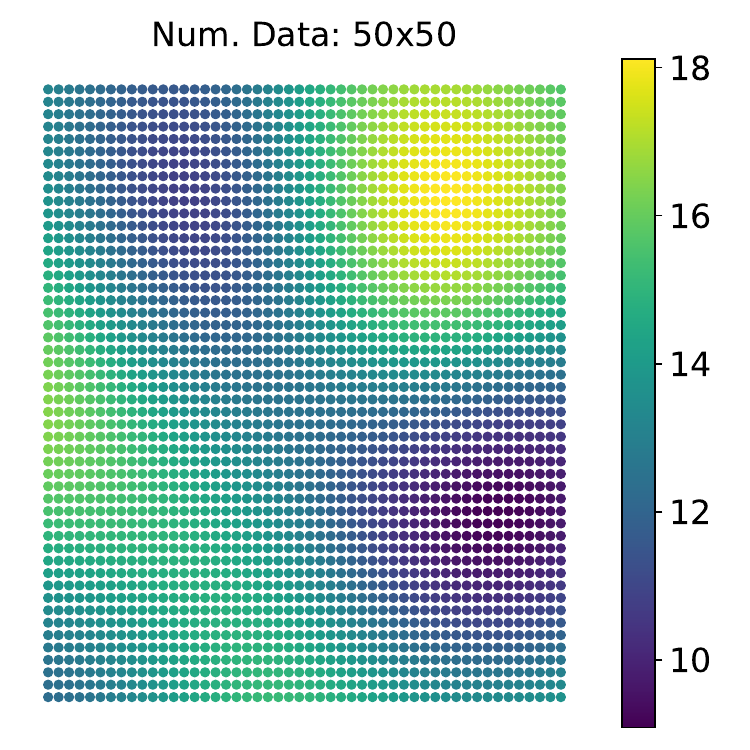}
    % \caption{}
    %\label{subfig:one}
  \end{subfigure}
  \begin{subfigure}[b]{0.3\textwidth}
    \includegraphics[width=\textwidth]{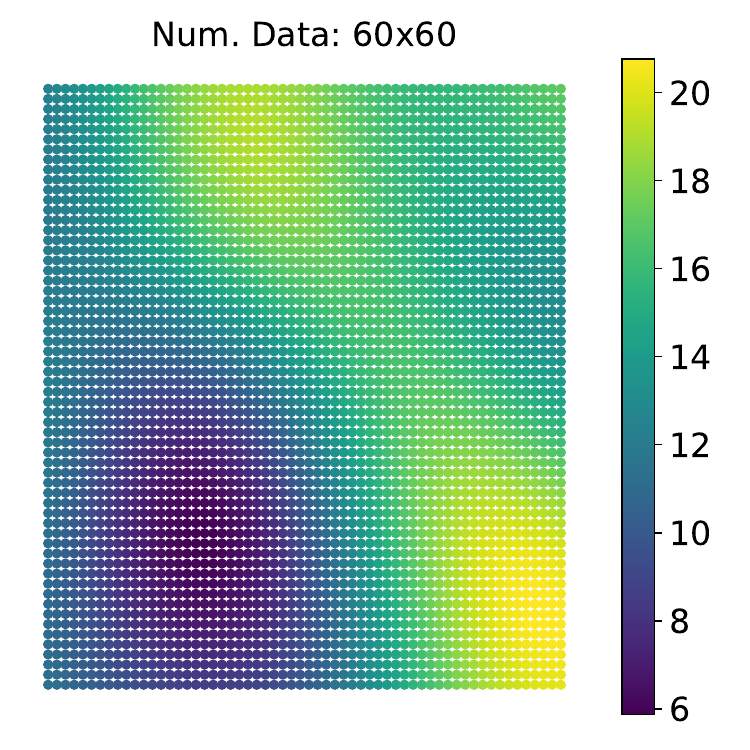}
    % \caption{}
    %\label{subfig:one}
  \end{subfigure}
  \caption{
  Solid Mechanics Model: Finite element nodal values of the elastic modulus field for three realizations, with varying mesh densities across the realizations. Units are in MPa.
  }
  \label{fig:solid-elasticMod}
\end{figure}

In the absence of body forces, the balance of linear momentum, combined with elastic constitutive laws, can be expressed as follows:
\begin{align}
    \frac{\partial s_{ij}}{\partial x_i} &= 0, \quad s_{ij} = C_{ijkl} (\boldsymbol{x}) \epsilon_{kl}, \quad
    C_{ijkl} (\boldsymbol{x}) = \lambda(\boldsymbol{x}) \delta_{ij}\delta_{kl} + \mu(\boldsymbol{x})\left(\delta_{ik}\delta_{jl} + \delta_{il}\delta_{jk}\right),
\end{align}
where $s_{ij}$ denotes symmetric stress tensor components, $\epsilon_{ij}$ denotes infinitesimal strain tensor components, $\mathbb{C}$ denotes elasticity tensor and Lame's first parameter $\lambda$ and shear modulus $\mu$ are related to Elastic modulus and Poisson's ratio. Poisson's ratio is set to a constant value of 0.3 throughout the domain. However, the elastic modulus modeled as a Gaussian random field with mean 15MPa, correlation length scale 2.5mm and standard deviation 0.25MPa, which varies spatially across the domain and differs among realizations. We consider a square plate with a side length of 1cm subjected to uniaxial loading along the vertical axis under plane stress conditions. The applied traction is 0.42MPa. We generate data on structured quadrilateral finite element meshes of varying densities, where the number of elements along each side of the square plate is a uniformly distributed random variable between 40 and 60. For illustration, three realizations are plotted in \cref{fig:solid-elasticMod}.

In this example, we aim to learn the stress fields corresponding to a given elastic modulus field. The proposed method is applied separately to each component of the stress tensor, namely $s_{11}$, $s_{12}$, and $s_{22}$. We apply the RINO method described in Section \ref{sec:OL-BrOnly}, where two separate sets of basis functions are learned: one for the input function and another for the output function. The basis functions are learned using the procedure outlined in Algorithm \ref{algo:DL-sbs} with network architecture is described in Appendix~\ref{appx:nn-config}.

As described earlier, once an appropriate set of basis functions for the input and output fields is identified, the operator learning task reduces to mapping the input embeddings (input basis coefficients) to the output embeddings (output basis coefficients). This mapping is parameterized using a MLP (see Appendix~\ref{appx:nn-config}).

\begin{figure}[!ht]
  \centering
  \begin{subfigure}[b]{0.8\textwidth}
    \includegraphics[width=\textwidth]{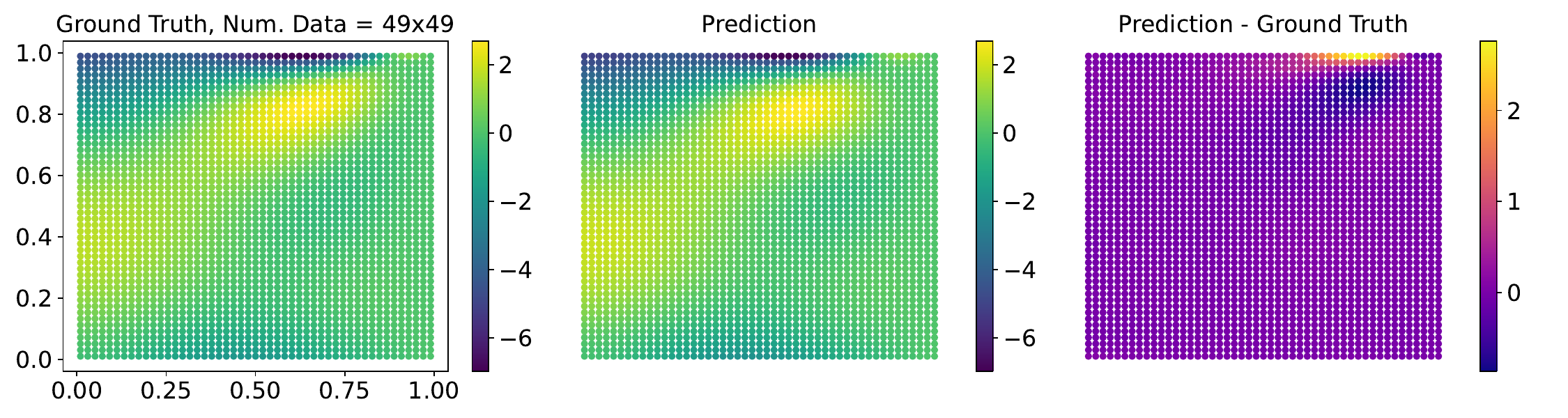}
    \caption{}
    %\label{subfig:one}
  \end{subfigure}
  \begin{subfigure}[b]{0.8\textwidth}
    \includegraphics[width=\textwidth]{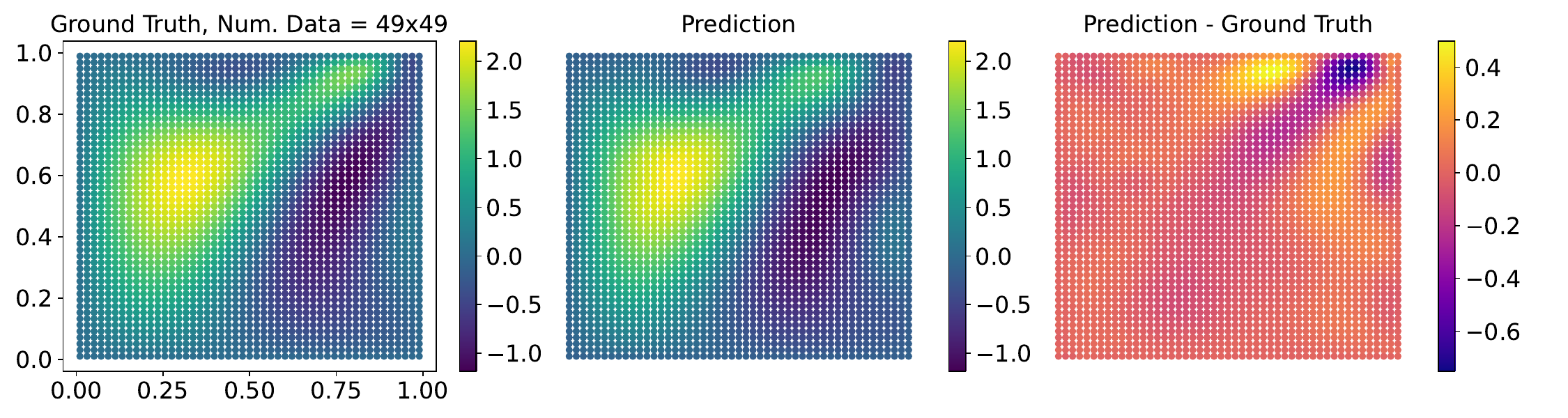}
    \caption{}
    %\label{subfig:one}
  \end{subfigure}
  \begin{subfigure}[b]{0.8\textwidth}
    \includegraphics[width=\textwidth]{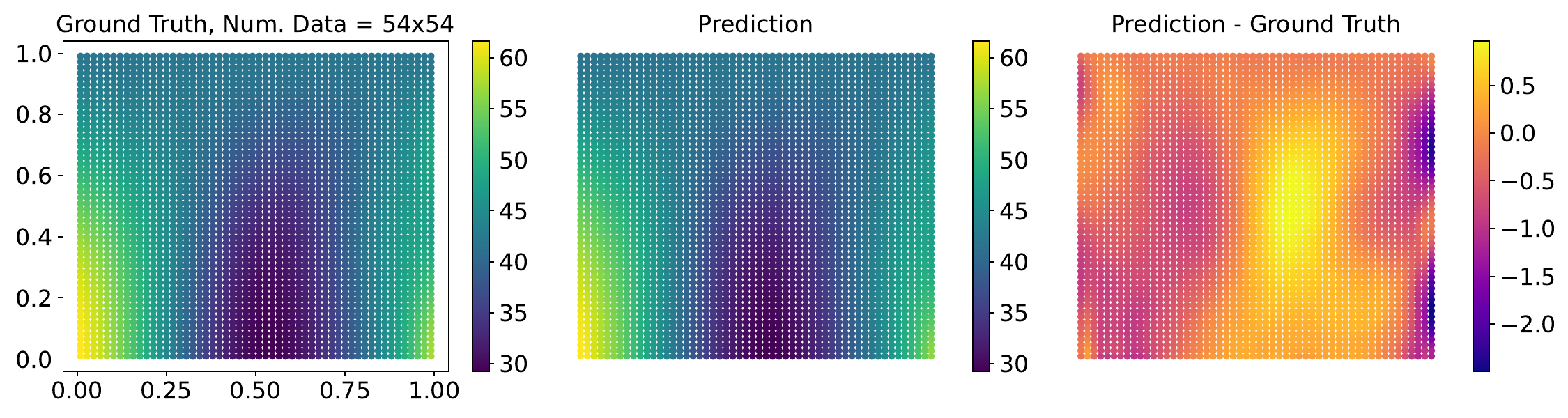}
    \caption{}
    %\label{subfig:one}
  \end{subfigure}
  \caption{
  Solid Mechanics Model: Worst case stress component predictions among test realizations: (a) $\sigma_{11}$, (b) $\sigma_{12}$, (c) $\sigma_{22}$.
  }
  \label{fig:solid-worst-test}
\end{figure}

\begin{figure}[!ht]
  \centering
  \begin{subfigure}[b]{0.25\textwidth}
    \includegraphics[width=\textwidth]{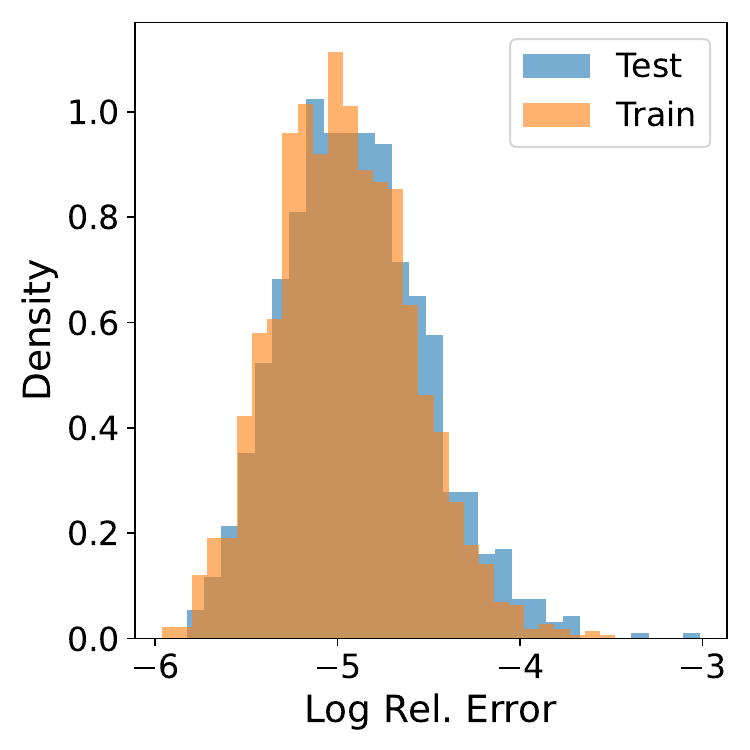}
    \caption{}
    %\label{subfig:one}
  \end{subfigure}
  \begin{subfigure}[b]{0.25\textwidth}
    \includegraphics[width=\textwidth]{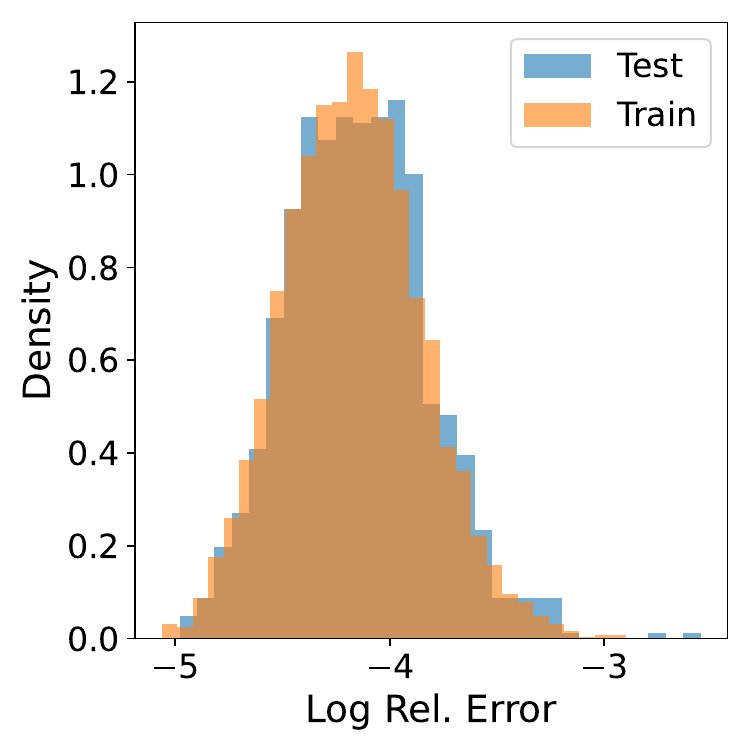}
    \caption{}
    %\label{subfig:one}
  \end{subfigure}
  \begin{subfigure}[b]{0.25\textwidth}
    \includegraphics[width=\textwidth]{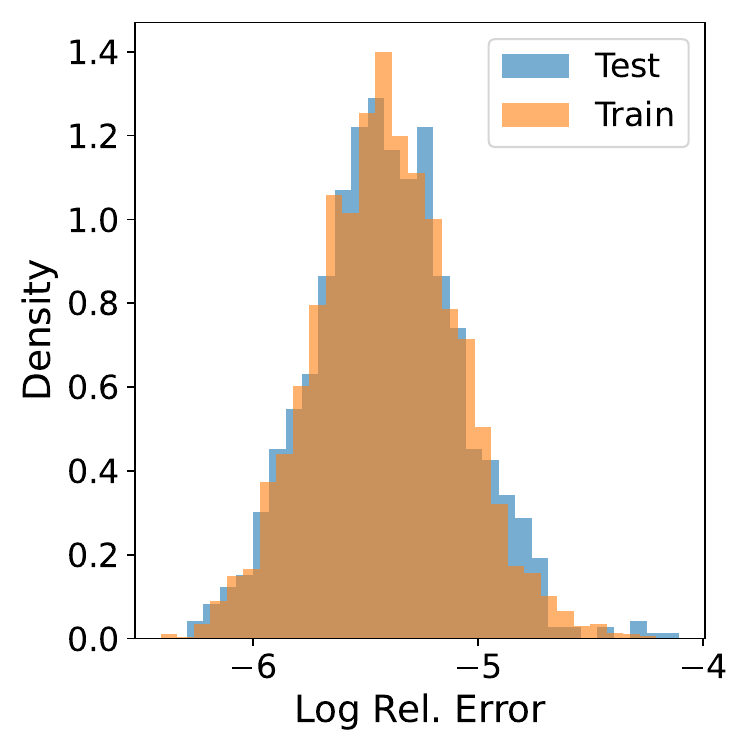}
    \caption{}
    %\label{subfig:one}
  \end{subfigure}
  \caption{
  Relative error distribution of stress component predictions among 4000 training and 1000 test realizations: (a) $\sigma_{11}$, (b) $\sigma_{12}$, (c) $\sigma_{22}$.
  }
  \label{fig:solid-err-dist}
\end{figure}

The worst-case relative error for stress component predictions on unseen test cases is plotted in \cref{fig:solid-worst-test}, demonstrating good accuracy. To further validate this, the relative mean square error across all training and test realizations is shown in \cref{fig:solid-err-dist}. The results suggest that not only is high accuracy achieved, but good generalization is also evident.

\section{Conclusion}
\label{sec:conclusion}
% interp
% extrapol

In this work, we propose a resolution-independent neural operator (RINO) that builds from the DeepONet formulation to directly work on (sufficiently rich) point clouds sampled from input and output functions without any underlying assumptions on the data structure. Specifically, we introduce a dictionary learning algorithm to identify appropriate continuous and fully differentiable basis functions for input and output function realizations. In the context of DeepONet, this allows us to construct an appropriate coordinate system (frame) to project input function point clouds onto, and use the resulting coordinates as the input function representation in the vanilla DeepONet; thus creating a resolution-independent DeepONet (RI-DeepONet). 
In fact, by applying the proposed dictionary learning algorithm separately to the input and output functions, we demonstrate that the operator learning task reduces to learning a mapping function between the coefficients of the input and output basis functions. This can be viewed as a new operator learning framework termed the RINO. Moreover, we show that for certain problems, the approximate operator in the constructed latent spaces significantly simplifies the operator learning optimization process -- e.g., a simple linear transformation between the embeddings.
The application of the proposed algorithm extends beyond the operator learning task and can be used for reconstruction tasks from incomplete or missing information, similar to GPOD. This algorithm can be considered a reconstruction method over function spaces rather than vector spaces, as is the case with GPOD or PCA.

We have empirically studied and demonstrated the resolution independence of the proposed operator learning scheme by solving several numerical experiments from the literature and newly designed ones. Different aspects of the algorithm have been discussed and particularly compared with GPOD. 

The introduced algorithm for finding a resolution-independent representation of the input function data has two additional advantages that ``come for free''. First, if there is an underlying lower-dimensional structure for the realizations, the algorithm can identify it and reduce the dimensionality of the input/output function data. This was demonstrated in all of the numerical examples. For instance, in one example, the data dimensionality was 100, and we achieved a relatively low reconstruction error using only 10 basis functions. This compact data representation can directly reduce the size of the operator network. Moreover, such reduced dimensions are more suited for uncertainty quantification and propagation. Second, the reconstruction error can naturally serve as a good estimate of out-of-sample or distribution indicators, which can be leveraged for data acquisition and experiment design.

\section{Acknowledgements}
We acknowledge the discussion with Professor George E. Karniadakis during the CRUNCH seminar, who brought random projection and ReNO \cite{bartolucci2024representation} to our attention, which improved the work.

This material is based upon work supported by the U.S. Department of Energy, Office of Science, Office of Advanced Scientific Computing Research, under Award Number DE-SC0024162.

\bibliographystyle{plain}
\bibliography{bibliography}  %%% Uncomment 
\section*{Appendix}
Supplementary results for the numerical examples are provided in \Cref{appdx:darcy1d,appdx:darcy2d,appdx:burger}. Appendix \ref{appx:nn-config} lists the hyperparameters used for each problem. An ablation study for Algorithm \ref{algo:DL}, comparing the RINO method with GPOD, is presented in Appendix \ref{appx:gpod}. Moreover, the use of random projections as an alternative for finding vector representations of input or output function data has been studied and compared to the proposed dictionary learning algorithm.
Additional potential advantages of RINO over RI-DeepONet in terms of generalization are examined in Appendix~\ref{appx:ri-vs-ri}.
Appendix \ref{appx:RFG} summarizes the formulation for Gaussian random field generation.

\appendix

\section{Nonlinear 1D Darcy's Equation}
\label{appdx:darcy1d}
% In this example, we increase the complexity of the previous problem by introducing a nonlinear operator. A variant of the nonlinear 1D Darcy's equation is given by the following form:
% \begin{equation}
%     \frac{ds}{dx}
%     \left(
%     -\kappa(s(x)) \frac{d s}{dx}
%     \right)
%      = u(x); \quad x\in[0, 1], 
% \end{equation}
% where the solution-dependent permeability is $\kappa(s(x)) = 0.2 + s^2(x)$ and the source term is a random field $u(x)\sim \mathcal{GP}$ with length scale $l=0.05$. Homogeneous Dirichlet boundary conditions $s=0$ are defined at the domain boundaries. The FEniCS finite element solver \cite{barrata2023dolfinx} is used to generate data by discretizing the domain into 50 uniformly spaced nodal points. Following the procedure described earlier, we set $M_{\text{min}} = 20$ and $M_{\text{max}} = 35$ to generate random point clouds for the training and testing cases. The sizes of the training and test datasets are 800 and 200, respectively.

\begin{figure}[!ht]
  \centering
  \begin{subfigure}[b]{0.3\textwidth}
    \includegraphics[width=\textwidth]{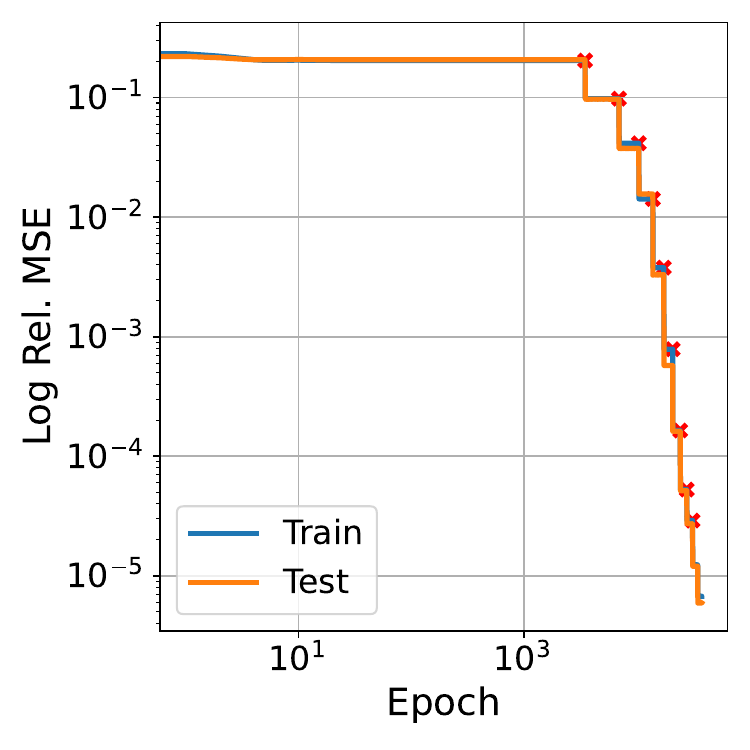}
    \caption{}
    %\label{subfig:one}
  \end{subfigure}
    \begin{subfigure}[b]{0.3\textwidth}
    \includegraphics[width=\textwidth]{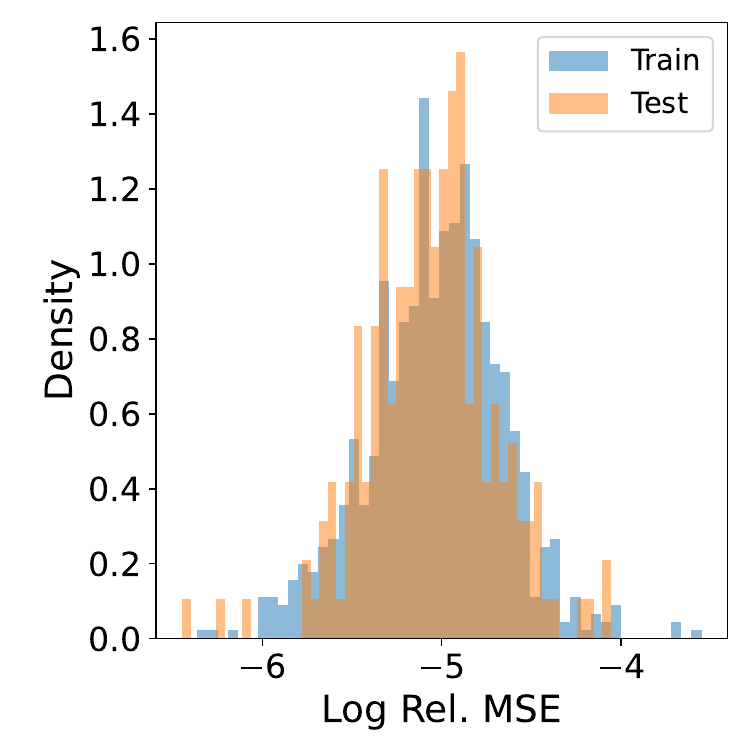}
    \caption{}
    %\label{subfig:one}
  \end{subfigure}
  \caption{1D Nonlinear Darcy Example: (a) Loss function during optimization iterations for dictionary learning of the input basis functions. (b) Distribution of reconstruction errors for the train and test datasets after training.
  % from rand 3
  }
  \label{fig:darcy-nonlin-loss-in}
\end{figure}

Figure~\ref{fig:darcy-nonlin-loss-in}(a) shows the reconstruction errors during the dictionary learning iterations. With a total of 10 learned basis functions, the data can be reconstructed with high accuracy, as shown by the distribution of errors in Figure~\ref{fig:darcy-nonlin-loss-in}(b) for both training and testing datasets. 

\begin{figure}[!ht]
  \centering
  \begin{subfigure}[b]{0.3\textwidth}
    \includegraphics[width=\textwidth]{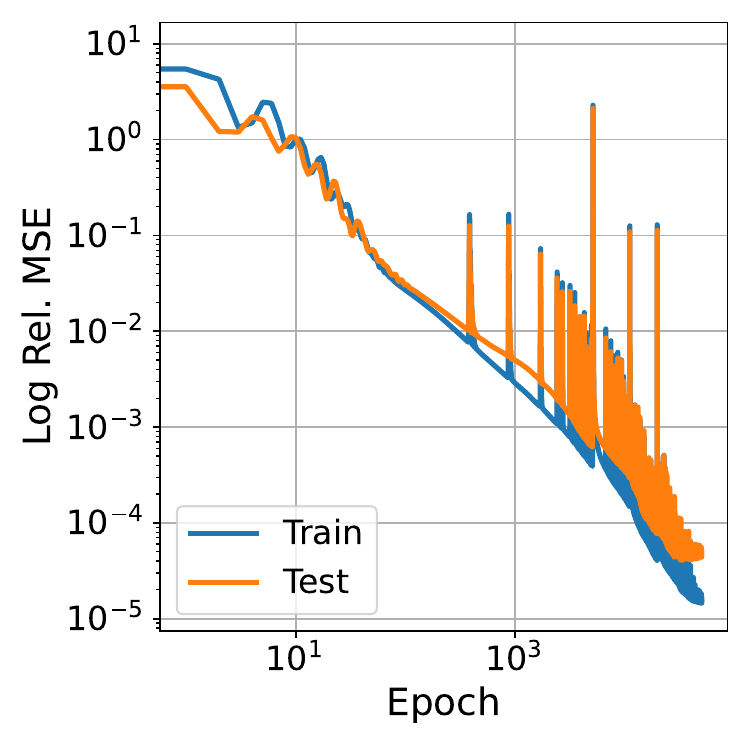}
    \caption{}
    %\label{subfig:one}
  \end{subfigure}
    \begin{subfigure}[b]{0.3\textwidth}
    \includegraphics[width=\textwidth]{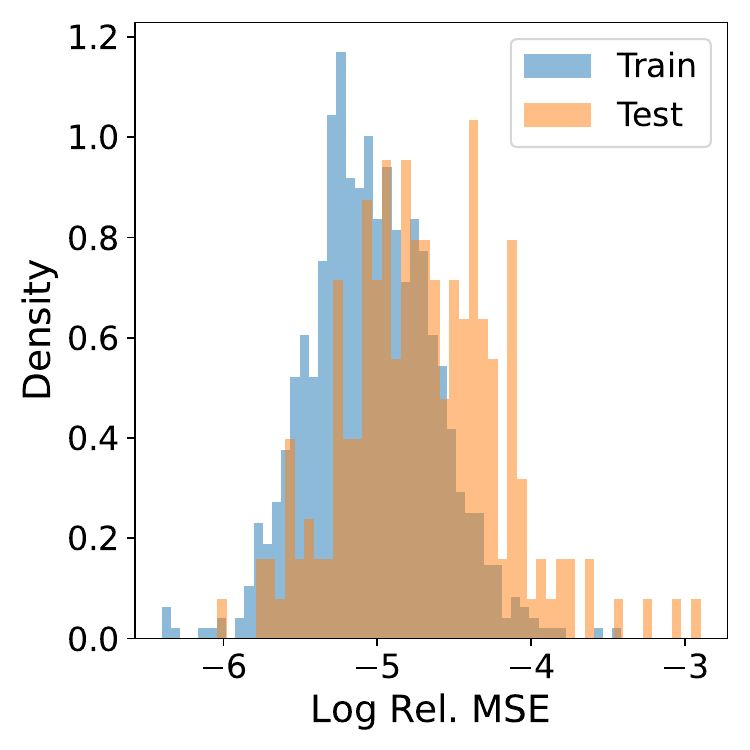}
    \caption{}
    %\label{subfig:one}
  \end{subfigure}
  \caption{1D Nonlinear Darcy Example: (a) Loss function during optimization iterations for operator learning. (b) Distribution of output function prediction errors for the training and testing datasets after training.
  % from rand 3
  }
  \label{fig:darcy-nonlin-loss-out}
\end{figure}

The history of the loss function during the operator learning stage is plotted in Figure~\ref{fig:darcy-nonlin-loss-out}(a). In Figure~\ref{fig:darcy-nonlin-loss-out}(b), the distribution of output function prediction errors after training is shown. The results suggest that the training is successful and the output function prediction via RI-DeepONet is very accurate.

\section{Nonlinear 2D Darcy's Equation}
\label{appdx:darcy2d}
% In this example, we extend the previous problem to a two-dimensional setting. The nonlinear Darcy's equation in two dimensions is given by:
% \begin{equation}
%     \nabla \cdot\left(-\kappa(s(\boldsymbol{x})) \nabla s\right)
%      = u(\boldsymbol{x}); \quad \boldsymbol{x}\in[0, 1]^2, 
% \end{equation}
% where the solution-dependent permeability is $\kappa(s(\boldsymbol{x})) = 0.2 + s^2(\boldsymbol{x})$ and the source term is a random field $u(\boldsymbol{x})\sim \mathcal{GP}$ with length scale $l=0.25$. Homogeneous Dirichlet boundary conditions $s=0$ are defined at the domain boundaries. The solution field is obtained via the finite element method. Input function random fields are discretized on a $20\times20$ uniform grid, while the triangulation of the finite element domain does not necessarily coincide with these input sensor point locations. Following the procedure described earlier, we set $M_{\text{min}} = 100$ and $M_{\text{max}} = 280$ to generate random point clouds for the training and testing cases. The sizes of the training and test datasets are 800 and 200, respectively.

\begin{figure}[!ht]
  \centering
  \begin{subfigure}[b]{0.3\textwidth}
    \includegraphics[width=\textwidth]{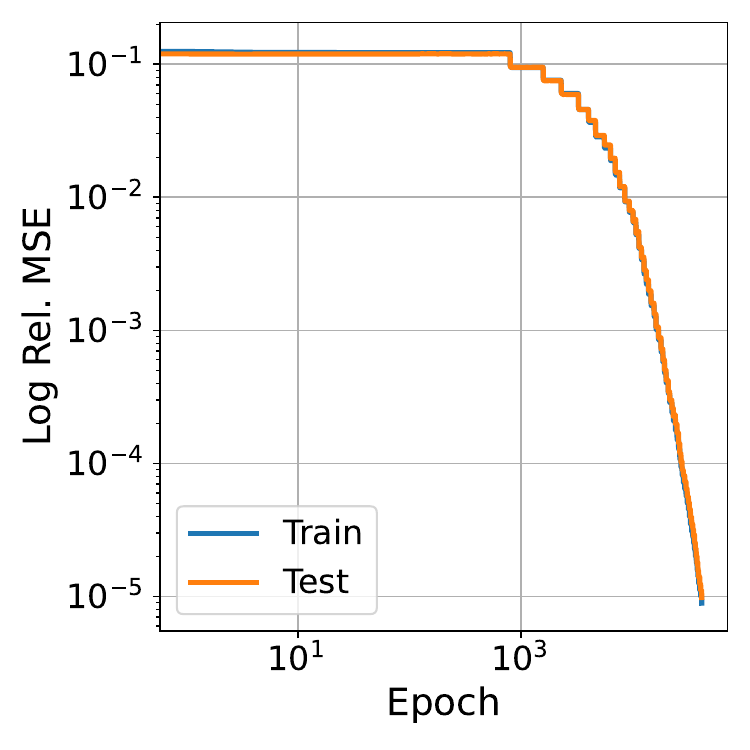}
    \caption{}
    %\label{subfig:one}
  \end{subfigure}
    \begin{subfigure}[b]{0.3\textwidth}
    \includegraphics[width=\textwidth]{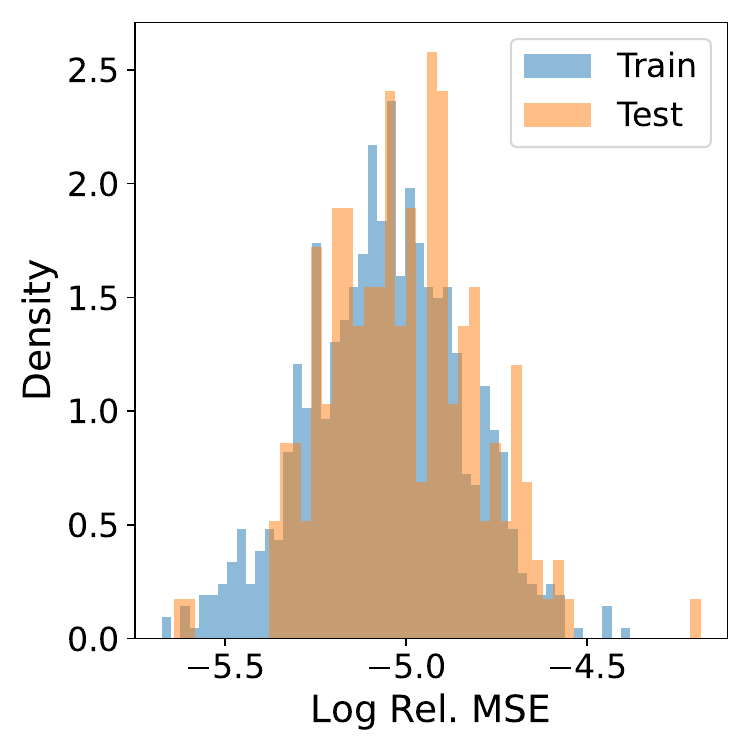}
    \caption{}
    %\label{subfig:one}
  \end{subfigure}
  \caption{2D Nonlinear Darcy Example: (a) Loss function during optimization iterations for dictionary learning of the input basis functions. (b) Distribution of reconstruction errors for the train and test datasets after training.
  % from rand 0
  }
  \label{fig:darcy2d-nonlin-loss-in}
\end{figure}

Figure~\ref{fig:darcy2d-nonlin-loss-in}(a) shows the reconstruction errors during the dictionary learning iterations. With a total of 57 learned basis functions, the data can be reconstructed with high accuracy, as shown by the distribution of errors in Figure~\ref{fig:darcy2d-nonlin-loss-in}(b) for both training and testing datasets. Notice that the learned dictionary allows us to represent data with 57 dimensions, which is much less than the ambient dimension of the discretized input function on a $20\times20$ grid. Therefore, we not only bypass the resolution dependence but also achieve a rich representation with lower dimension than the ambient data dimensionality. From a computational perspective, the input dimension for the branch network is 57, whereas using an MLP branch for the vanilla DeepONet requires 400 dimensions.

\begin{figure}[!ht]
  \centering
  \begin{subfigure}[b]{0.3\textwidth}
    \includegraphics[width=\textwidth]{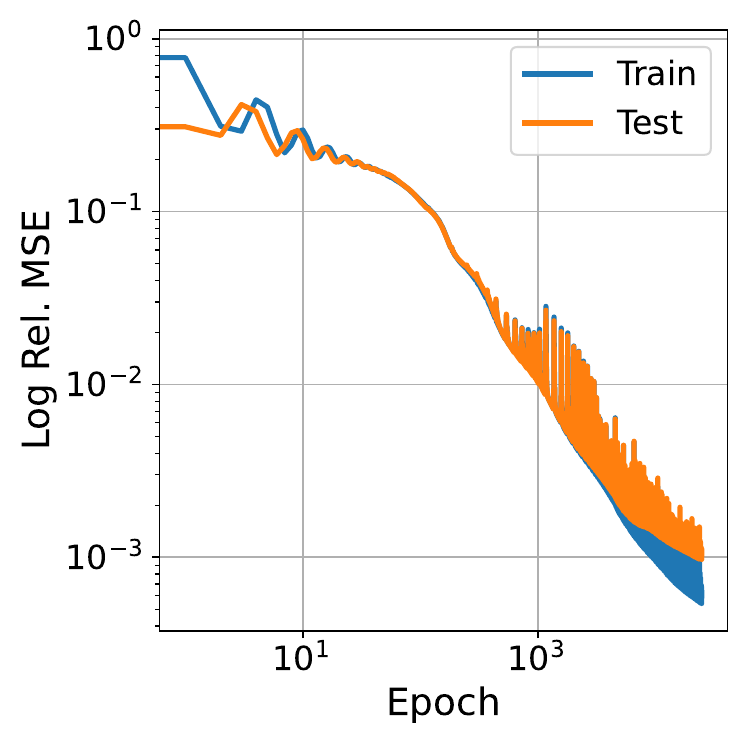}
    \caption{}
    %\label{subfig:one}
  \end{subfigure}
    \begin{subfigure}[b]{0.3\textwidth}
    \includegraphics[width=\textwidth]{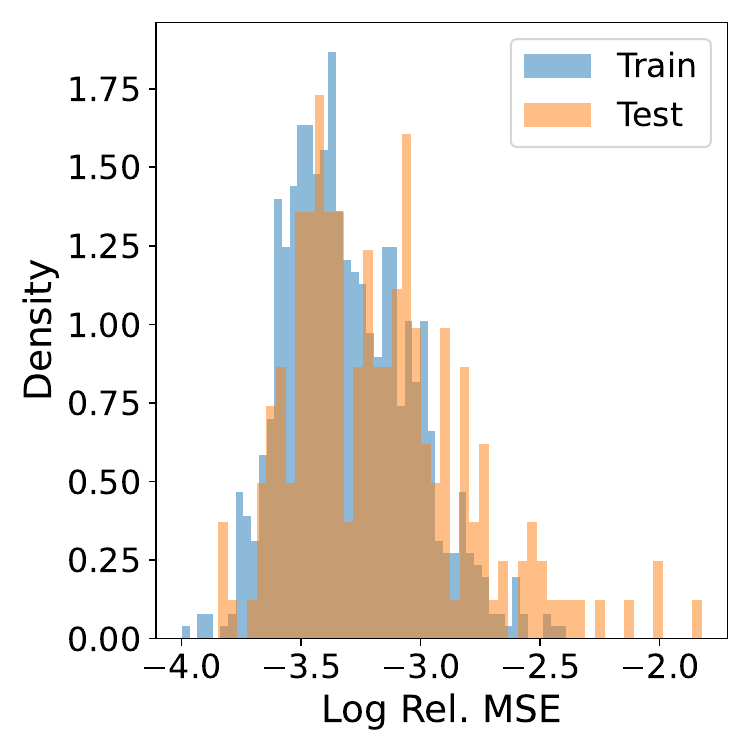}
    \caption{}
    %\label{subfig:one}
  \end{subfigure}
  \caption{2D Nonlinear Darcy Example: (a) Loss function during optimization iterations for operator learning. (b) Distribution of output function prediction errors for the training and testing datasets after training.
  % from rand 0
  }
  \label{fig:darcy2d-nonlin-loss-out}
\end{figure}

The history of the loss function during the operator learning stage is plotted in Figure~\ref{fig:darcy2d-nonlin-loss-out}(a). In Figure~\ref{fig:darcy2d-nonlin-loss-out}(b), the distribution of output function prediction errors after training is shown. 

\begin{figure}[!ht]
  \centering
    \begin{subfigure}[b]{0.75\textwidth}
    \includegraphics[width=\textwidth]{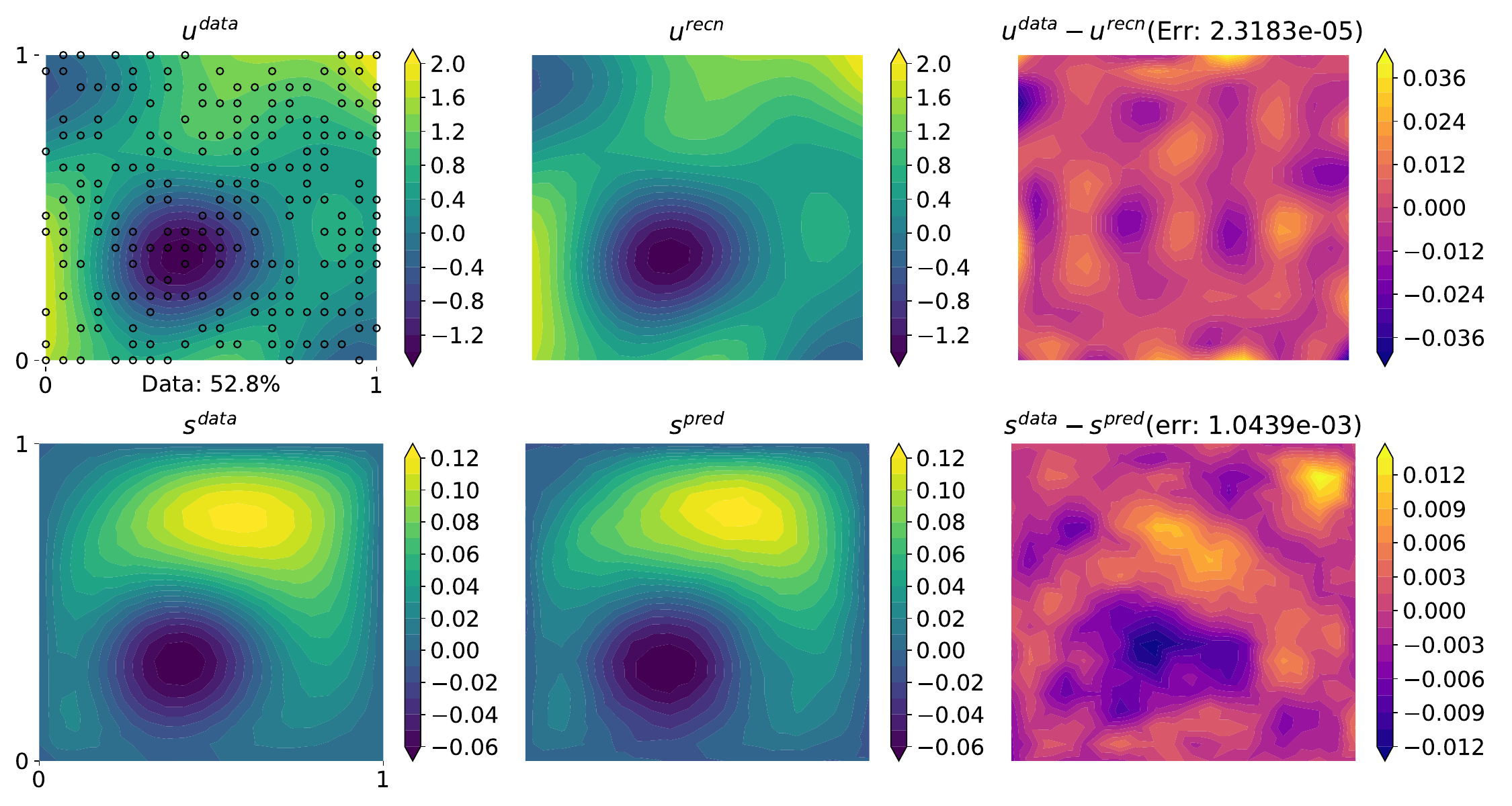}
    %\caption{}
    %\label{subfig:one}
  \end{subfigure}
  \caption{2D Nonlinear Darcy Example: The case with 75th percentile output function prediction error. The top left figure shows the queried input function data $u(\boldsymbol{x})$, with black circle symbols indicating the queried data. The middle top figure shows the input field reconstruction via the learned dictionary. The top right figure shows the error between the reconstructed input field and the ground truth. The bottom figures show the output field $s(\boldsymbol{y})$, the RI-DeepONet prediction, and its corresponding errors.
  % from rand 0
  }
  \label{fig:darcy2d-nonlin-pred-Q1}
\end{figure}

\begin{figure}[!ht]
  \centering
    \begin{subfigure}[b]{0.75\textwidth}
    \includegraphics[width=\textwidth]{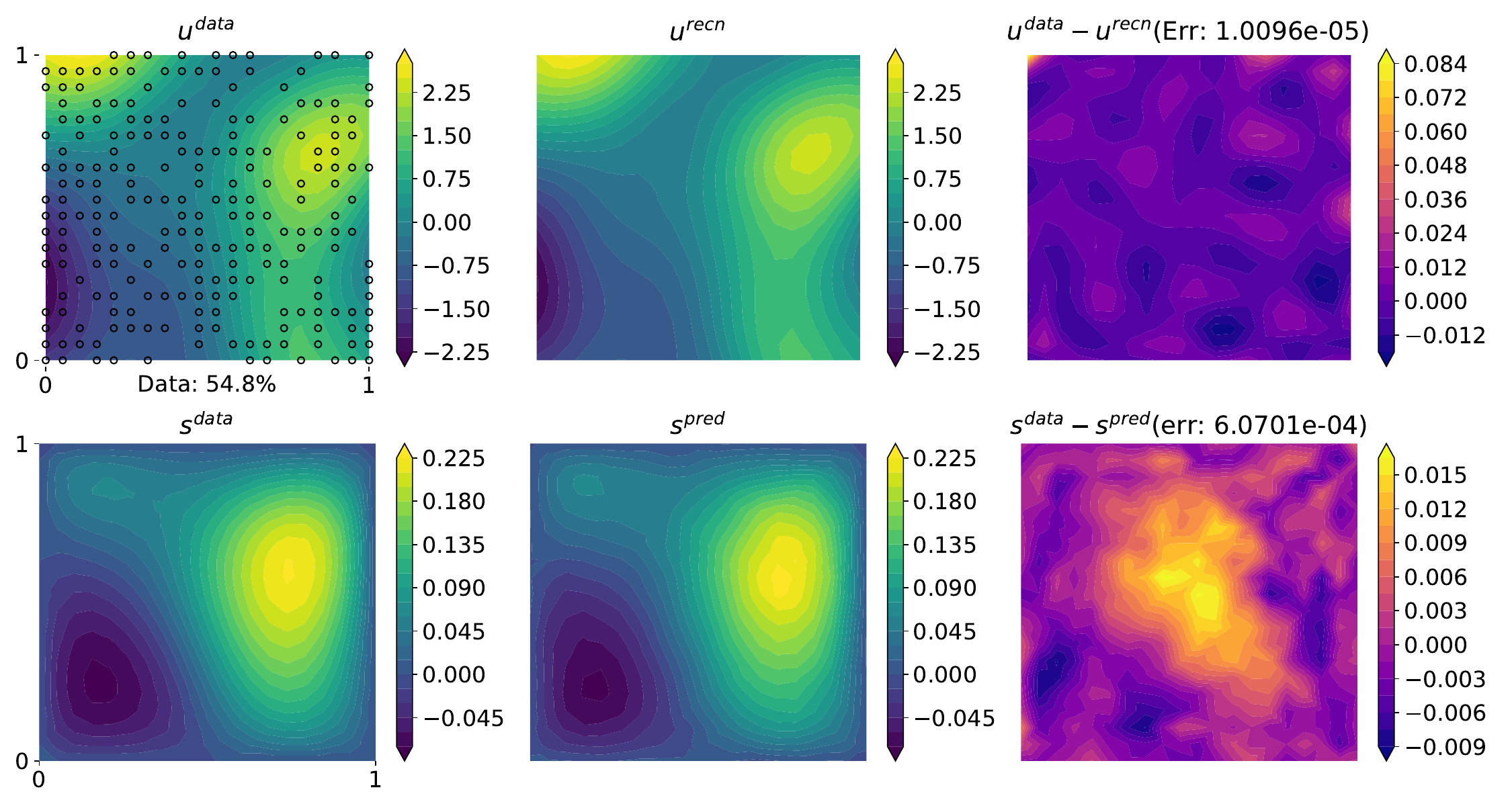}
    %\caption{}
    %\label{subfig:one}
  \end{subfigure}
  \caption{2D Nonlinear Darcy Example: The case with 50th percentile output function prediction error. The top left figure shows the queried input function data $u(\boldsymbol{x})$, with black circle symbols indicating the queried data. The middle top figure shows the input field reconstruction via the learned dictionary. The top right figure shows the error between the reconstructed input field and the ground truth. The bottom figures show the output field $s(\boldsymbol{y})$, the RI-DeepONet prediction, and its corresponding errors.
  % from rand 0
  }
  \label{fig:darcy2d-nonlin-pred-Q2}
\end{figure}

Two examples of the test predictions for the output field are shown in  \cref{fig:darcy2d-nonlin-pred-Q1,fig:darcy2d-nonlin-pred-Q2}, corresponding to the 75th percentile error and the median error, respectively. In each example, the queried point cloud input function data has different numbers of points and locations because the masked information varies among them.

\section{Nonlinear Burgers' Equation}
\label{appdx:burger}
\begin{figure}[!ht]
  \centering
  \begin{subfigure}[b]{0.3\textwidth}
    \includegraphics[width=\textwidth]{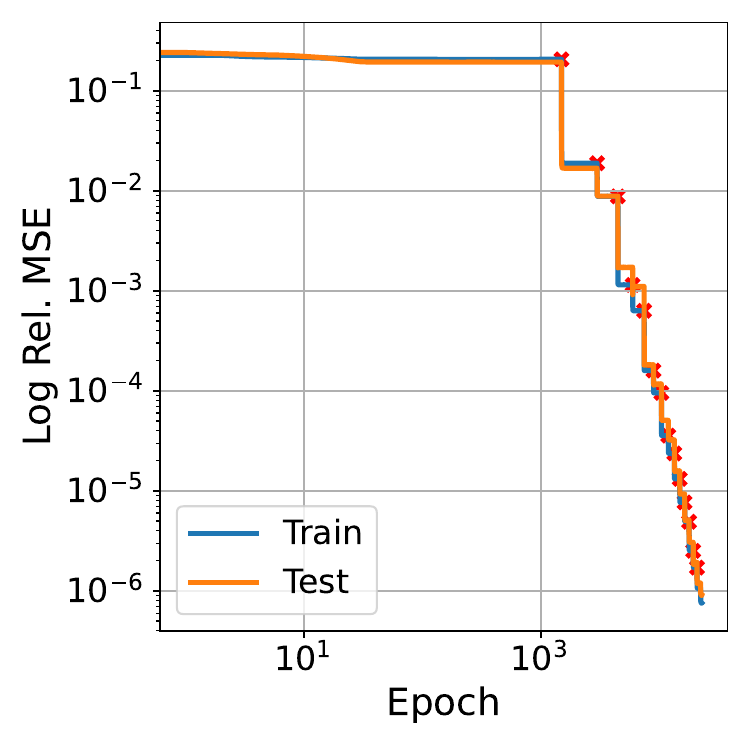}
    \caption{}
    %\label{subfig:one}
  \end{subfigure}
    \begin{subfigure}[b]{0.3\textwidth}
    \includegraphics[width=\textwidth]{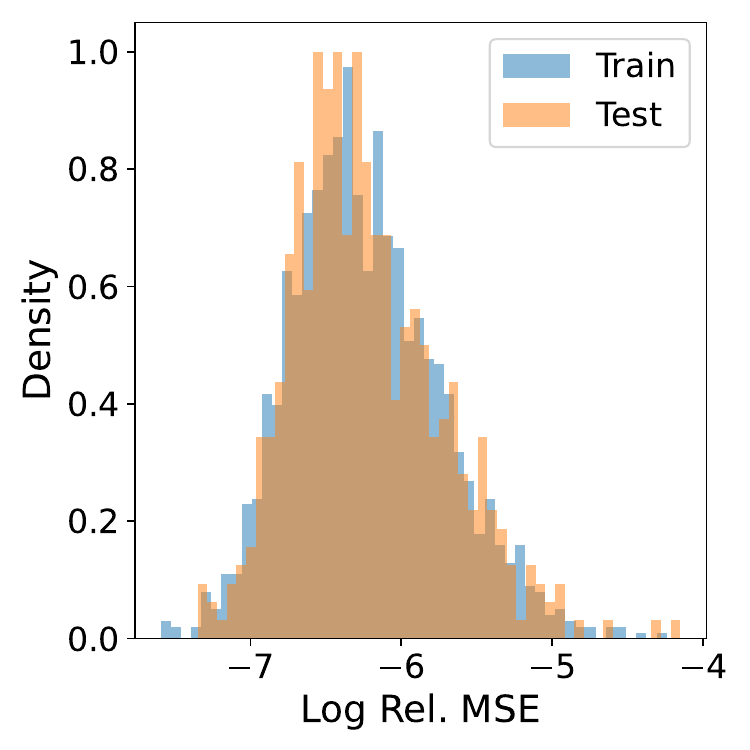}
    \caption{}
    %\label{subfig:one}
  \end{subfigure}
  \caption{Burgers' Equation: (a) Loss function during optimization iterations for dictionary learning of the input basis functions. (b) Distribution of reconstruction errors for the train and test datasets after training.
  % from rand 1
  }
  \label{fig:Burgers-loss-in}
\end{figure}

Figure~\ref{fig:Burgers-loss-in}(a) shows the reconstruction errors during the dictionary learning iterations. The distribution of reconstruction errors in Figure~\ref{fig:Burgers-loss-in}(b) for both training and testing datasets indicates that the learned dictionary, with a total of 16 basis functions, can satisfactorily represent the input function data.

\begin{figure}[!ht]
  \centering
  \begin{subfigure}[b]{0.3\textwidth}
    \includegraphics[width=\textwidth]{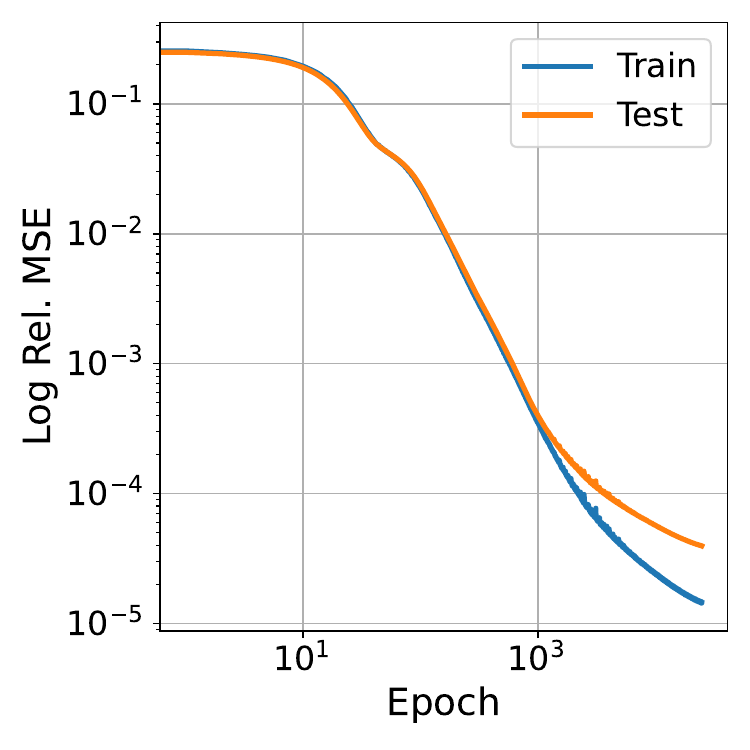}
    \caption{}
    %\label{subfig:one}
  \end{subfigure}
    \begin{subfigure}[b]{0.3\textwidth}
    \includegraphics[width=\textwidth]{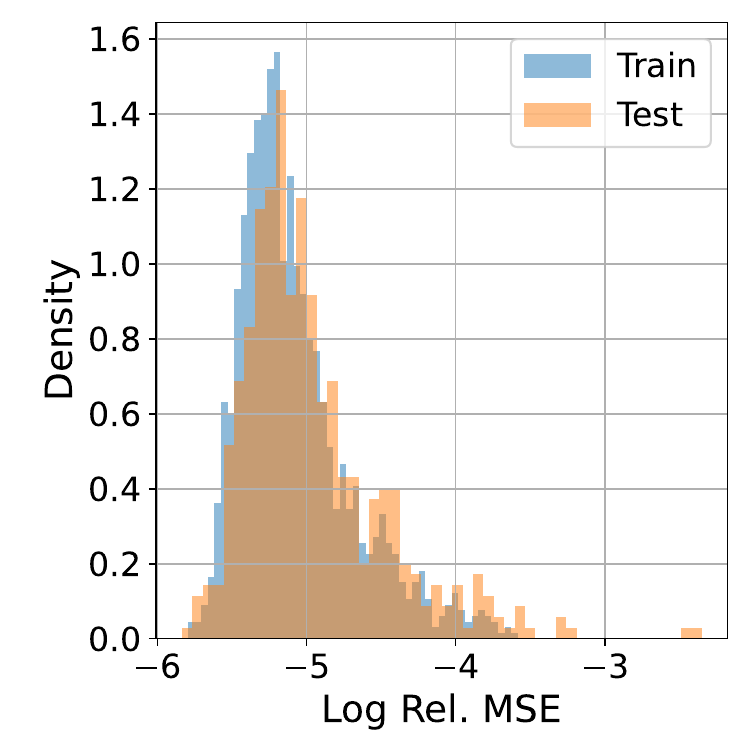}
    \caption{}
    %\label{subfig:one}
  \end{subfigure}
  \caption{Burgers' Equation: (a) Loss function during optimization iterations for operator learning. (b) Distribution of output function prediction errors for the training and testing datasets after training.
  % from rand 1 (last time was 0 mistake)
  }
  \label{fig:Burgers-loss-out}
\end{figure}

Figure~\ref{fig:Burgers-loss-out}(a) illustrates the evolution of the loss function during the operator learning stage. The distribution of output function prediction errors after training is depicted in Figure~\ref{fig:Burgers-loss-out}(b). These results indicate that the training was successful, leading to accurate output function predictions by RI-DeepONet.

\section{Nonlinear 2D Darcy's Equation on Unstructured Meshes}
\label{appdx:darcy2dMesh}

\begin{figure}[!ht]
  \centering
  \begin{subfigure}[b]{0.3\textwidth}
    \includegraphics[width=\textwidth]{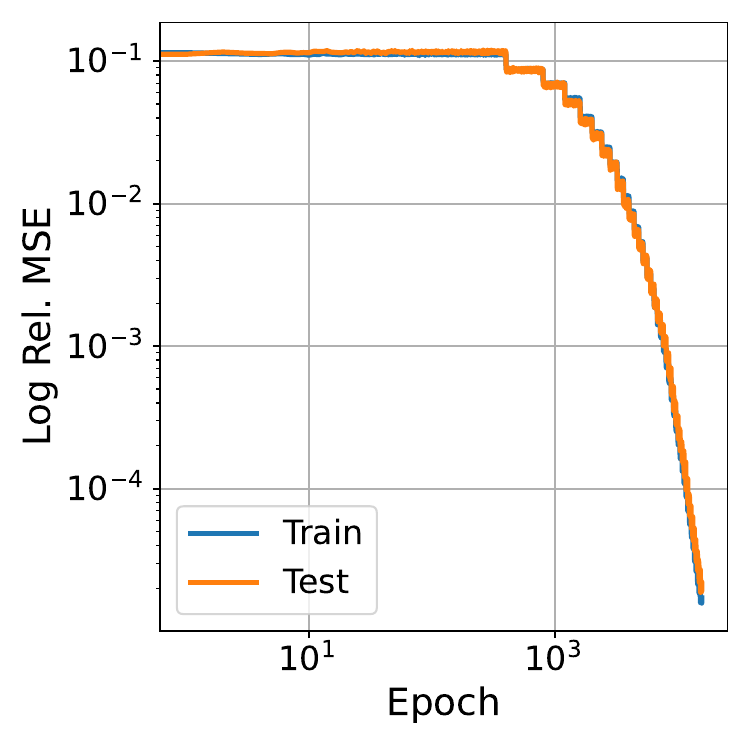}
    \caption{}
    %\label{subfig:one}
  \end{subfigure}
    \begin{subfigure}[b]{0.3\textwidth}
    \includegraphics[width=\textwidth]{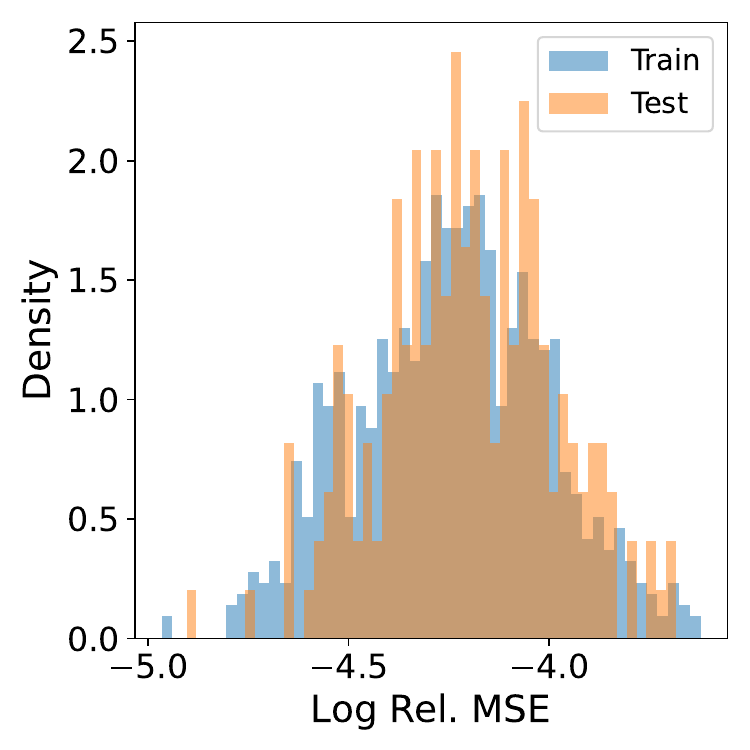}
    \caption{}
    %\label{subfig:one}
  \end{subfigure}
  \caption{2D Darcy's Equation on Unstructured Meshes: (a) Loss function during optimization iterations for dictionary learning of the input basis functions. (b) Distribution of reconstruction errors for the train and test datasets after training.
  % from rand 4
  }
  \label{fig:darcy2dMesh-loss-in}
\end{figure}

Figure~\ref{fig:darcy2dMesh-loss-in}(a) shows the reconstruction errors during the dictionary learning iterations for the input function point cloud data. The distribution of reconstruction errors in Figure~\ref{fig:darcy2dMesh-loss-in}(b) suggests that a satisfactory reconstruction error can be achieved by the learned dictionary with 40 basis functions.

\begin{figure}[!ht]
  \centering
    \includegraphics[width=0.3\textwidth]{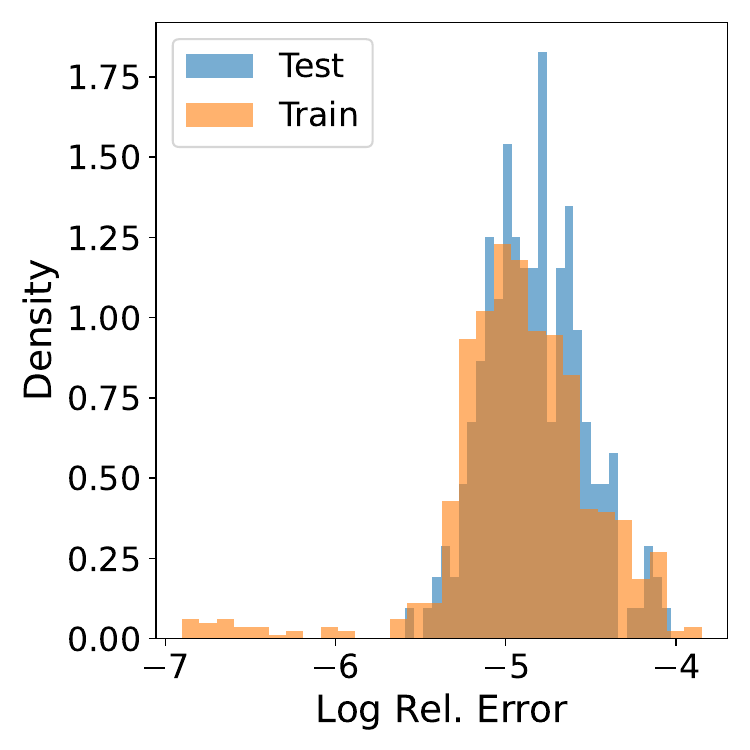}
  \caption{
  2D Darcy's Equation on Unstructured Meshes: Distribution of reconstruction errors for the training and test datasets after learning the dictionary of basis functions for the output function.
  }
  \label{fig:darcy2dMesh-nonlin-out-dl-loss}
\end{figure}

As discussed in the main text, the proposed RINO method can build the neural operator by learning approporiate basis functions for the output function data as well. Figure~\ref{fig:darcy2dMesh-nonlin-out-dl-loss} shows the reconstruction error distributions for the output function point cloud data, where Algorithm~\ref{algo:DL-sbs} was able to identify an appropriate dictionary of 37 basis functions that span the output solution fields.

\section{Network Architecture}
\label{appx:nn-config}
In this work, the neural network architectures and hyperparameters were chosen through manual trial and error and, therefore, are not optimized. As a result, the performance may improve by utilizing rigorous hyperparameter optimization methods, such as Bayesian optimization \cite{snoek2012practical}. In the following, ``2D Darcy\textsuperscript{*}'' corresponds to the numerical example where we solve the nonlinear 2D Darcy problem on different finite element meshes from one realization to another.

\subsection{Input basis functions}
The hyperparameters utilized for learning the dictionary of input basis functions are listed in Table~\ref{tab:hyper-dl-in}.
\begin{table}[!ht]
\centering
\caption{Hyperparameters of SIREN input basis functions used in the dictionary learning algorithm}
\label{tab:example}
\begin{tabular}{cccccccc}
\toprule
 & Antiderivative & 1D Darcy & 2D Darcy & Burger's & 2D Darcy\textsuperscript{*} & 2D Elasticity\\
\midrule
Number of hidden units  & $20$ & $30$ & $50$ & $30$ & $20$ & $100$\\
Number of hidden layers  & $2$ & $2$ & $3$ & $2$ & $2$ & $2$\\
Learning rate  & $1.66\times10^{-4}$ & $3\times10^{-4}$ & $6.6\times10^{-4}$ & $6.6\times10^{-4}$ & $4\times10^{-4}$ & $5\times10^{-4}$\\
$\lambda$  & $10^{-4}$ & $10^{-4}$ & $10^{-5}$ & $10^{-5}$ & $10^{-3}$ & $10^{-5}$\\
$\omega_0$  & $5$ & $10$ & $10$ & $10$ & $10$ & 20\\
\bottomrule
\label{tab:hyper-dl-in}
\end{tabular}
\end{table}

\subsection{Output basis functions}
The hyperparameters utilized for learning the dictionary of output basis functions are listed in Table~\ref{tab:hyper-dl-out}. In the case of 2D solid mechanics, $\omega_0$ is set to 35, 25, and 15 for stress components $s_{11}$, $s_{12}$, and $s_{22}$, respectively.
\begin{table}[!ht]
\centering
\caption{Hyperparameters of SIREN output basis functions used in the dictionary learning algorithm}
\label{tab:example}
\begin{tabular}{cccccccc}
\toprule
 & 2D Darcy\textsuperscript{*} & 2D Solid Mechanics\\
\midrule
Number of hidden units  & $20$ & $100$\\
Number of hidden layers  & $2$ & $2$\\
Learning rate  & $1\times10^{-3}$ & $5\times10^{-4}$\\
$\lambda$  & $10^{-4}$ & $10^{-5}$\\
$\omega_0$  & $10$ & 35, 25, 15\\
\bottomrule
\label{tab:hyper-dl-out}
\end{tabular}
\end{table}

\subsection{Operator learning}
The obtained embeddings are not normalized in any case except for the nonlinear 2D Darcy equation, where the sparse codes from the dictionary are linearly normalized to range between 0 and 1 prior to training. The learning rate of 0.001 is chosen for all operator learning tasks.

\textbf{Antiderivative}: RI-DeepONet is employed without a branch network; the sparse codes obtained from the dictionary are directly used as coefficients for the trunk basis, effectively making the branch network an identity operator. The trunk network is SIREN with two hidden layers of size 50 and $w_0=5$.

\textbf{Nonlinear 1D Darcy}: RI-DeepONet is employed with a branch network consisting of an MLP with two hidden layers, each of size 50, and \texttt{ReLU} activation, and a trunk network implemented as a SIREN with three hidden layers, each of size 50, and a frequency parameter $w_0=5$.

\textbf{Nonlinear 2D Darcy}: RI-DeepONet is employed with a branch network consisting of an MLP with two hidden layers of sizes 50 and 100, using \texttt{ReLU} activation, and a trunk network implemented as an MLP with three hidden layers of sizes 50, 50, and 100, also using \texttt{ReLU} activation.

\textbf{Nonlinear Burgers' equation}: RI-POD-DeepONet is employed with a branch network consisting of three hidden layers of sizes 100, 100, and 70 using \texttt{Tanh} activation.

\textbf{Nonlinear 2D Darcy}\textsuperscript{*}: RINO is employed, where the network $\mathcal{F}^{\text{ri}}$ is an MLP with a single hidden layer of size 40, using \texttt{Tanh} activation. The input and output embeddings have dimensionality 40 and 37, respectively.

\textbf{Solid Mechanics}: RINO is employed, where the networks $\mathcal{F}^{\text{ri}}$ are MLPs with three hidden layers and \texttt{Tanh} activation functions. The number of hidden units are 100, 50, and 50 for stress components $s_{11}$, $s_{12}$, and $s_{22}$ respectively. These MLPs are used to map input embedding of dimension 24 to output embeddings of dimensions 41, 29, and 14 for $s_{11}$, $s_{12}$, and $s_{22}$ respectively.

\section{Ablation Study: Comparison with Gappy Proper Orthogonal Decomposition and Random Projection
}
\label{appx:gpod}
% order of basis
% orthogonality of generative and identified basis
% comparison errors
In this example, we aim to empirically demonstrate that Algorithm~\ref{algo:DL} can effectively identify continuous (weakly) orthogonal basis functions that accurately represent point clouds sampled from a random function. We compare the proposed dictionary learning to GPOD, but we note first some important distinctions.
% To enable a comparison with the GPOD method, we configure the problem setup to be compatible with it. 
The proposed method shares certain similarities with GPOD, but offers greater flexibility. In particular, it can operate with point cloud data instead of regular grids and, more importantly, it finds continuous, fully differentiable basis functions rather than finite-dimensional basis vectors. Moreover, with GPOD, one cannot query an arbitrary sensor location that differs from the union of sensor locations used in the training data due to its discrete nature, while the proposed approach allows queries at any location.

\begin{figure}[!ht]
  \centering
  \begin{subfigure}[b]{0.3\textwidth}
    \includegraphics[width=\textwidth]{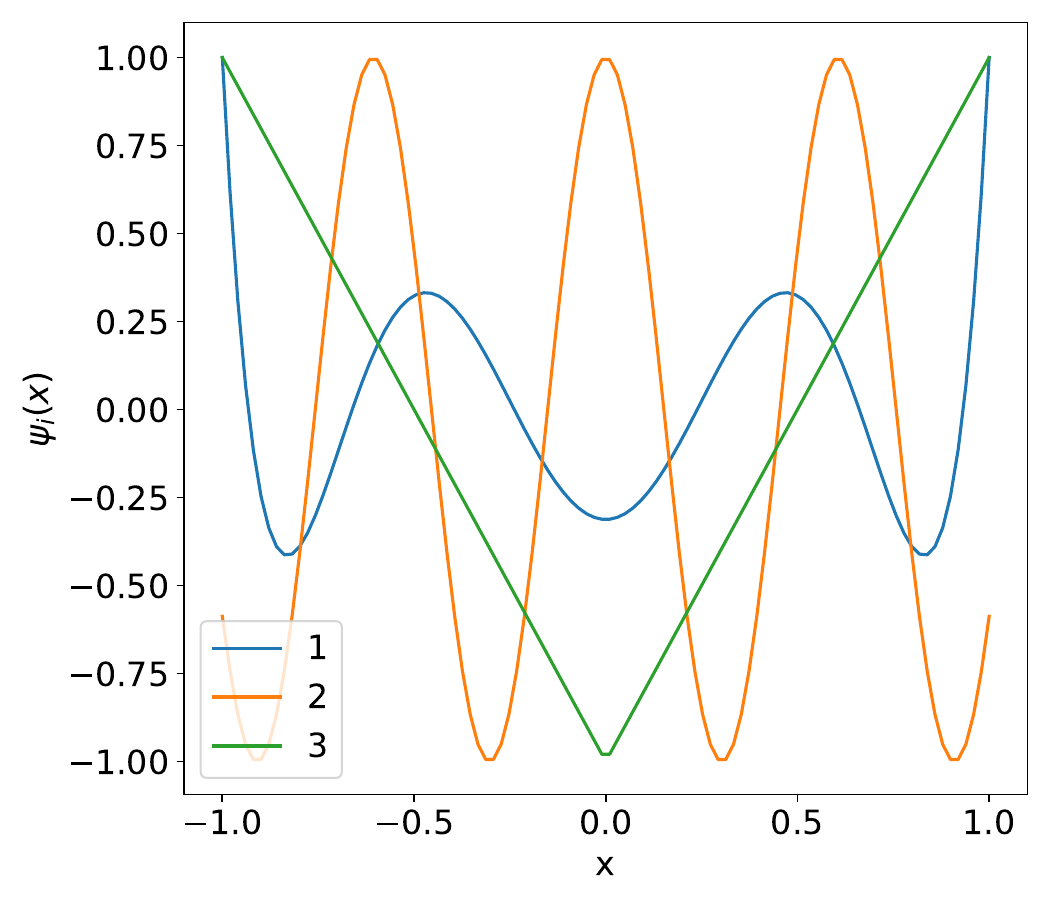}
    \caption{Arbitrary bases}
    %\label{subfig:one}
  \end{subfigure}
  \begin{subfigure}[b]{0.3\textwidth}
    \includegraphics[width=\textwidth]{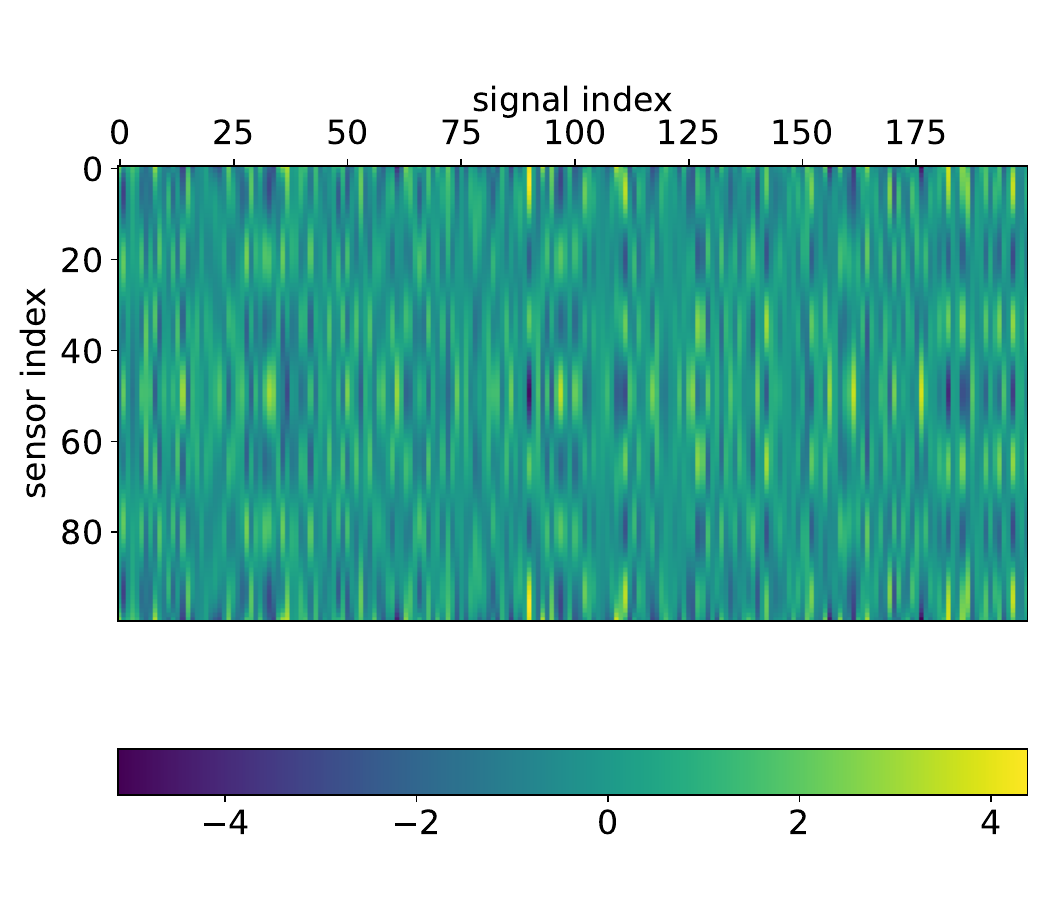}
    \caption{Complete data (training)}
    %\label{subfig:one}
  \end{subfigure}
    \begin{subfigure}[b]{0.3\textwidth}
    \includegraphics[width=\textwidth]{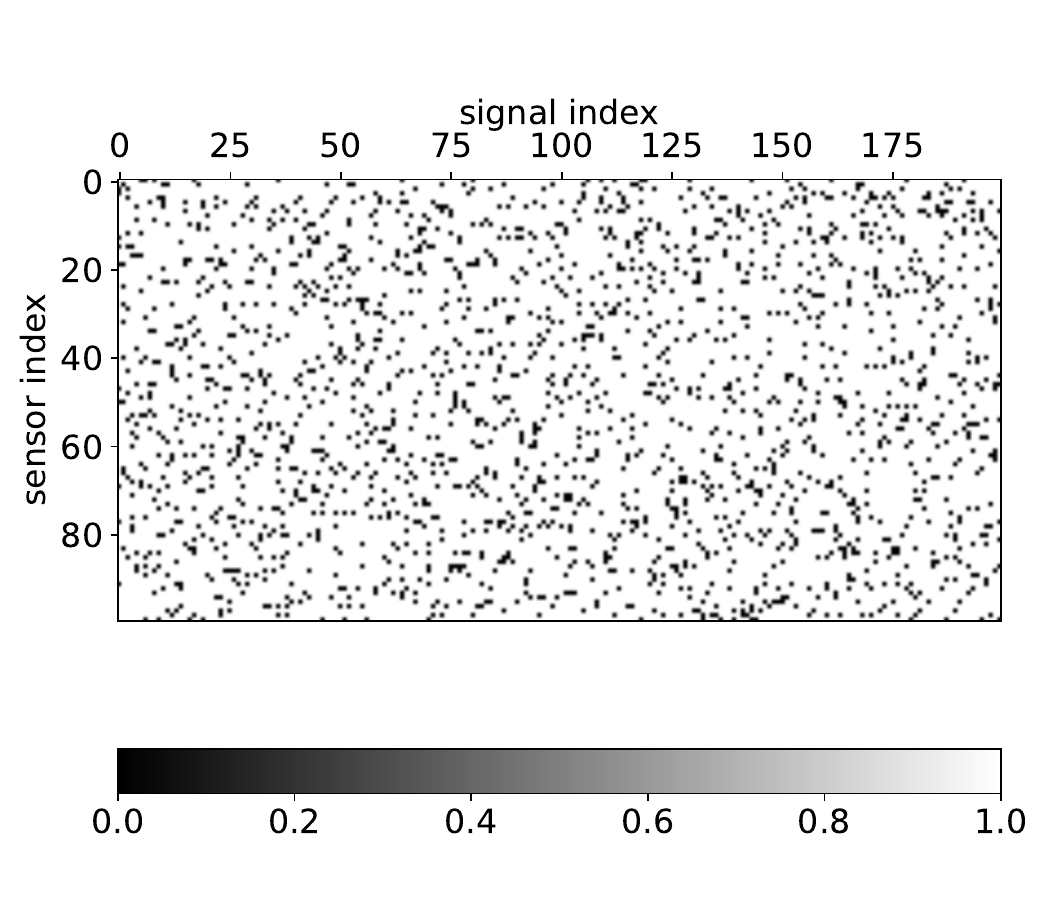}
    \caption{90\% mask matrix (training)}
    %\label{subfig:one}
  \end{subfigure}
  \caption{(a) Arbitrary basis functions used to generate random functions. (b) Complete training data set of 200 random functions generated from the arbitrary basis in (a), each sampled at 100 sensor locations. (c) Example of the mask matrix representing 90\% masked training data for each function, where white (value 1) indicates masked entries.}
  \label{fig:gpod-data}
\end{figure}

For the purpose of comparison, we generate several realizations of random functions $u(x)$ that share a common low-dimensional structure defined through the following expansion where the $i$-th realization is given by:
\begin{equation}
    u^{(i)}(x) = \sum_{l=1}^3 \alpha_l^{(i)} \psi_l(x), \quad x \in [0, 1],
\label{eq:gen-pod}
\end{equation}
the coefficients are assumed to be normally distributed, $\alpha_l^{(i)} \sim \mathcal{N}(0, 1)$, and the arbitrary basis functions are the 6th-order Legendre polynomial, $\psi_1(x) = P_6(x)$; the cosine function, $\psi_2(x) = \cos(3.3\pi x)$; and the absolute value function, $\psi_3(x) = 2 |x| - 1$, as depicted in Figure~\ref{fig:gpod-data}(a). The random functions are generated similarly to PCE, where deterministic functions with random coefficients are used to model stochastic processes.
Clearly, such random functions can be fully characterized by at most three modes, each corresponding to one of these functions.
We generate 200 realizations of this random function, where each realization is discretized at 100 uniform sensor locations to establish the “complete” training dataset, as plotted in the matrix in Figure~\ref{fig:gpod-data}(b). In this figure, each row represents a realization, each column corresponds to a sensor location, and the colors at the $(i, j)$ location indicate the value of the $i$-th realization at the $j$-th sensor location. The ``gappy'' version of the data is created by randomly masking $R\%$ of each realization. For example, in Figure~\ref{fig:gpod-data}(c), the mask matrix for masking $90\%$ of the information is presented.

\begin{figure}[!ht]
  \centering
  \begin{subfigure}[b]{0.25\textwidth}
    \includegraphics[width=\textwidth]{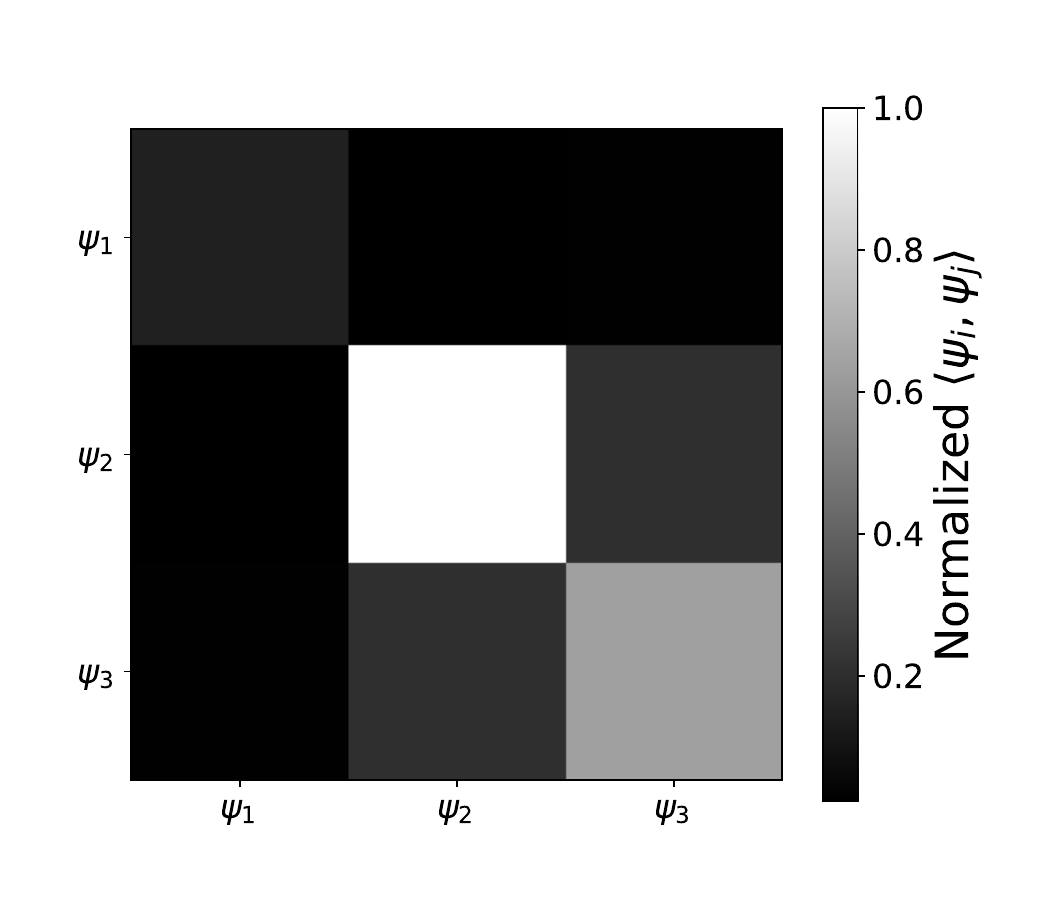}
    \caption{Data Model}
    %\label{subfig:one}
  \end{subfigure}
    \begin{subfigure}[b]{0.25\textwidth}
    \includegraphics[width=\textwidth]{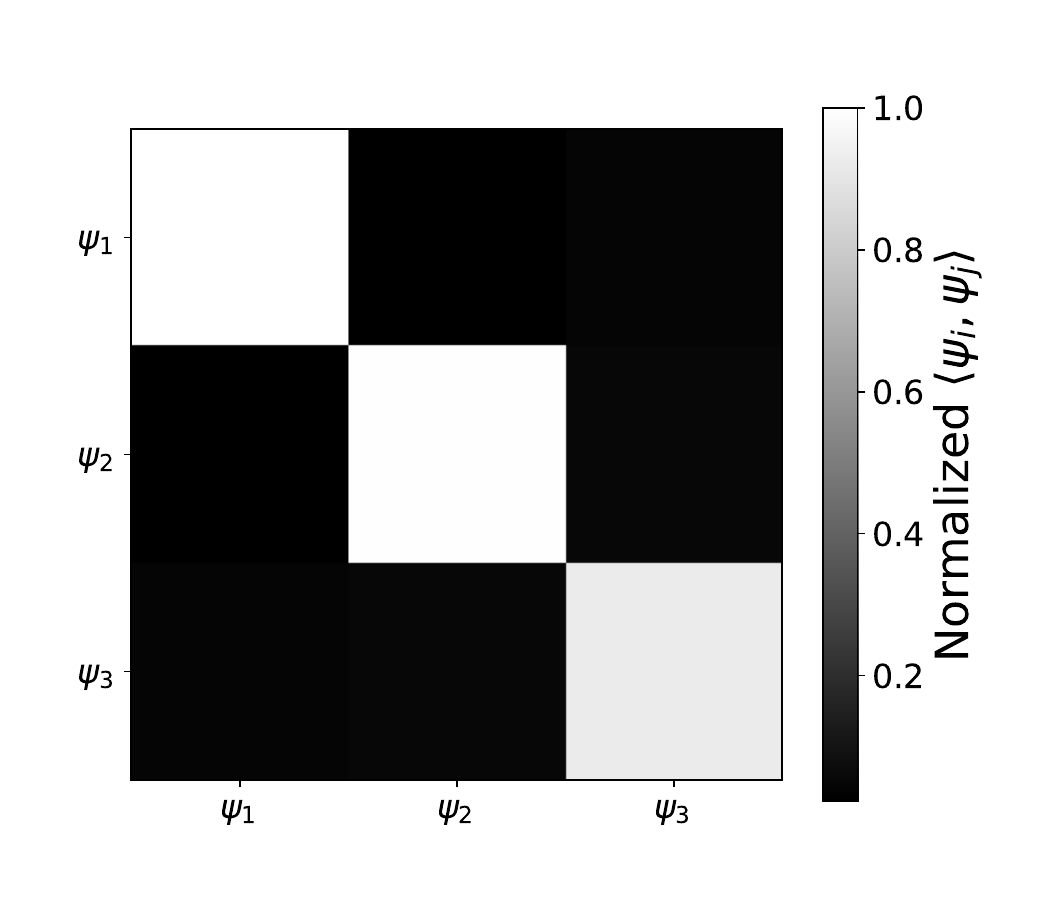}
    \caption{Gappy POD}
    %\label{subfig:one}
  \end{subfigure}
    \begin{subfigure}[b]{0.25\textwidth}
    \includegraphics[width=\textwidth]{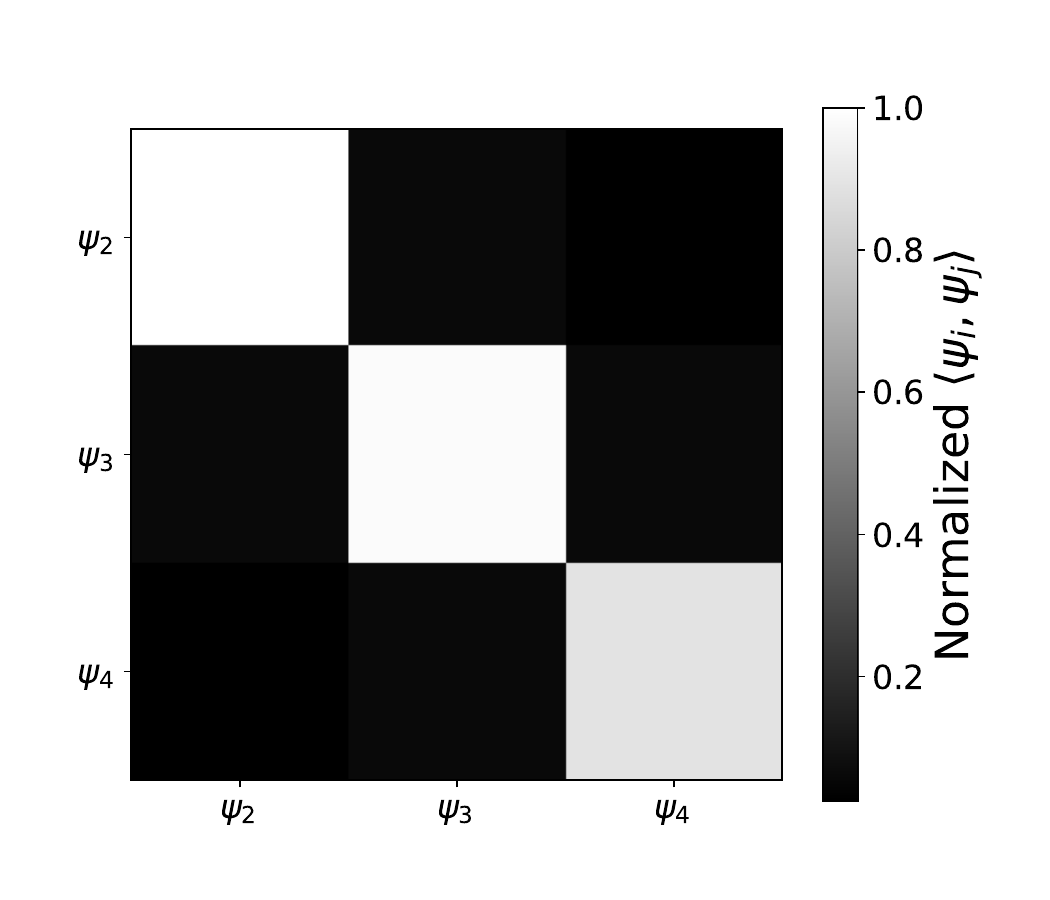}
    \caption{Algorithm~\ref{algo:DL}}
    %\label{subfig:one}
  \end{subfigure}
  \caption{Orthogonality check among basis functions: (a) Basis functions used to generate the random function data (see Eq.~\ref{eq:gen-pod}); (b) Basis vectors identified by the Gappy POD algorithm with 50\% of the data masked; and (c) Basis functions identified by the proposed dictionary learning algorithm with 50\% of the data masked.
  }
  \label{fig:gpod-orthog}
\end{figure}

Note that the arbitrary basis functions used to generate the random functions from Eq.~\eqref{eq:gen-pod}, are purposely chosen to not be orthogonal (see Figure~\ref{fig:gpod-orthog}(a)) and exhibit a multiscale nature, as they have different length scales. We seek to investigate whether the proposed dictionary learning algorithm can find orthogonal basis functions, how similar these basis functions are to those found by GPOD, and how closely they resemble the non-orthogonal basis functions used to generate the data. It is well-known that the basis \textit{vectors} learned by PCA-based algorithms, such as GPOD, will be orthogonal.

\begin{figure}[!ht]
  \centering
    \includegraphics[width=\textwidth]{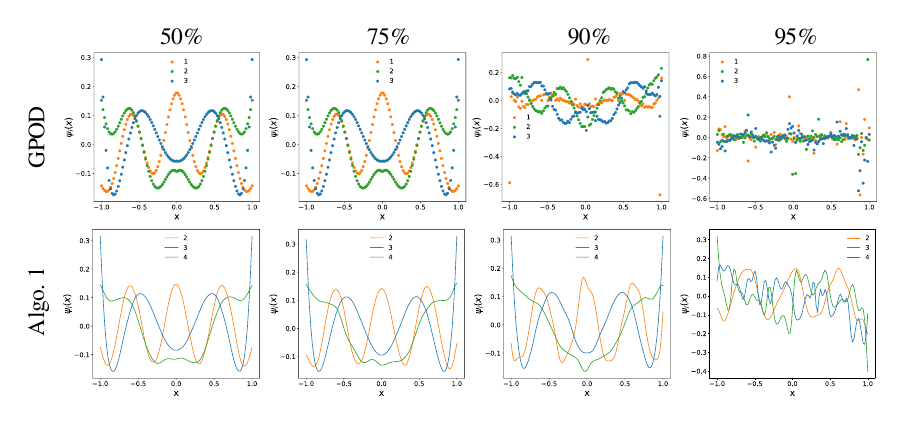}
  \caption{
  Comparison of identified bases using Gappy POD and Algorithm~\ref{algo:DL} at different levels of masked data. The percentage indicates the extent of data masking.
  }
  \label{fig:gpod-basis-all}
\end{figure}

In Figure~\ref{fig:gpod-basis-all}, the discovered bases from GPOD and the proposed dictionary learning method are plotted for various levels of masked data, ranging from $50\%$ to $95\%$. With sufficient unmasked data (e.g., $R=50\%,75\%$), the first two modes (orange and blue curves) from both GPOD and dictionary learning closely resemble the cosine function and the Legendre function, which were initially weakly orthogonal in the generative process, as shown in  Figure~\ref{fig:gpod-orthog}(a). However, the third mode (green curve) differs from the absolute value function, which was the non-orthogonal basis in the generative process, see Figure~\ref{fig:gpod-orthog}(a). From the GPOD perspective, this is expected since it represents data using orthogonal bases. Interestingly, the third basis function from the proposed dictionary learning closely resembles the third basis from GPOD. Furthermore, the order of the bases in the proposed dictionary learning approach matches the order in GPOD. These similarities occur even though we only weakly enforce the orthogonality constraint and do not enforce ordering based on data variance, as is the case with GPOD. The (weak) orthogonality of the bases in the proposed method can be confirmed by the inner product calculation shown in Figure~\ref{fig:gpod-orthog}(c).

Figure~\ref{fig:gpod-basis-all} further shows that, up to a certain level of data masking (75\%), the GPOD results are consistent with dictionary learning and Table~\ref{tab:gpod-all-err} shows that it can achieve very small reconstruction error. However, at higher levels of masking (e.g., $R\ge90\%$), its accuracy drops significantly and the basis vectors become noisy. The proposed dictionary learning method, on the other hand, still produces smooth, orthogonal basis functions with low reconstruction error. In fact, even in the extreme case of $R=95\%$, the basis functions remain smooth and the accuracy, though, diminished is much better than GPOD, which has extremely noisy basis vectors.   
% can still operate with reduced accuracy under similar conditions. We hypothesize that this is due to our method's inductive bias towards smoothness in the basis functions, as they are parameterized by SIREN, whereas GPOD does not incorporate such an assumption. 
% Nonetheless, our method also becomes less effective when the data is extremely sparse, as evidenced by the case with $R=95\%$.

The reconstruction errors at various level of masking percentages are listed in Table~\ref{tab:gpod-all-err} for both GPOD and the proposed dictionary learning. Note that the GPOD is a deterministic algorithm and, hence, there is no variability in its prediction accuracy. However, the proposed dictionary learning algorithm relies on INRs as basis functions, where different randomly initialized parameters for the neural network result in slightly different predictions. Therefore, we repeated the training five times, each with different initial neural network parameters. As the results suggest, for small masking percentage the accuracy of GPOD in this problem is better than Algorithm~\ref{algo:DL} and is resilient to the amount of masked data up to a certain limit, after which it suddenly fails catastrophically. On the other hand, the proposed scheme is still very accurate for small masking percentage and gradually reduces accuracy as the amount of masked data increases. Importantly, it continues to perform well under high masking rates where GPOD fails. This abrupt reduction in accuracy might be undesirable, particularly for data acquisition tasks where a well-behaved continuous error feedback is important.

\begin{table}
\centering
\begin{tabular}{@{}lccc@{}}
\toprule
     & $R\%$ & Train Error  & Test Error (mean $\pm$ std) \\ 
\midrule
 \textbf{GPOD}\\
    & 25  & 1.407e-08 $\pm$ 0 & 3.601e-08 $\pm$ 0\\
    & 50 & 2.670e-08 $\pm$ 0  & 4.253e-08 $\pm$ 0  \\
    & 75 & 2.360e-08 $\pm$ 0  & 2.360e-08 $\pm$ 0 \\
    & 90 & 2.390e+02 $\pm$ 0  & 2.474e+01 $\pm$ 0 \\
    & 95 & 2.357e+03 $\pm$ 0  & 8.262e+02 $\pm$ 0 \\
  \midrule
     \textbf{Algorithm~\ref{algo:DL}}\\
    & 25 & 3.733e-05 $\pm$ 3.273e-06 & 3.914e-05 $\pm$ 3.403e-06 \\
    & 50 & 9.291e-05 $\pm$ 9.549e-06 & 9.678e-05 $\pm$ 1.012e-05 \\
    & 75 & 6.263e-04 $\pm$ 4.066e-05 & 6.502e-04 $\pm$ 4.340e-05  \\
    & 90 & 2.609e-03 $\pm$ 1.358e-04 & 3.302e-03 $\pm$ 2.357e-04  \\
    & 95 & 1.423e-02 $\pm$ 2.625e-03 & 3.991e-02 $\pm$ 3.489e-03  \\
  \bottomrule                          
\end{tabular}
\caption{Reconstruction errors for various percentages of randomly masked data. The same masking percentage $R$ is used for both training and testing.}
\label{tab:gpod-all-err}
\end{table}

\begin{figure}[!ht]
  \centering
    \includegraphics[width=0.7\textwidth]{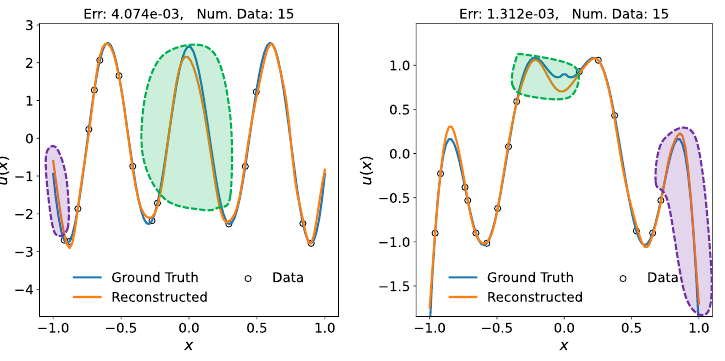}
  \caption{
  Reconstruction generalization capability of the proposed dictionary learning method, demonstrated on two queried samples, with the dictionary trained on only 15\% of the data. Purple areas highlight regions of extrapolation and green areas show regions where the model interpolates well despite lacking data elucidating features in these regions.
  }
  \label{fig:abl1-recont}
\end{figure}

The most important feature of the proposed dictionary learning is that we learn smooth, continuous orthogonal basis functions from all realizations, which improves generalization in predictions. In Figure~\ref{fig:abl1-recont}, we demonstrate the generalization ability of the proposed method. The shaded regions in green and purple represent two types of generalization: interpolation and extrapolation of in-distribution functions. If one attempts per-sample curve fitting, there is no guarantee that such fitting will capture important features of the underlying ground truth function, as fluctuations in these regions exceed the data frequency for interpolation; according to the Nyquist sampling theorem, the sampling rate should be at least twice the frequency of changes. Particularly under this scenario, the purple region can be considered pure extrapolation, which is known to be challenging without additional assumptions. The reason the proposed reconstruction method performs so well is that it identifies common structures in the data set by considering \textbf{all samples (realizations)}, not just individual samples, and utilizes basis functions that harness this structure. Moreover, although the available data in these samples are highly asymmetric, the reconstruction is almost symmetric. This symmetry is due to the identified symmetric basis functions, which are one of the main characteristics of the generated data, as the generative basis functions are symmetric. This property cannot be achieved through per-sample curve fitting. Therefore, we do not advocate for methods that attempt per-sample curve fitting and then align the samples by interpolation at fixed sensor locations.

Given that the advantages of the proposed dictionary learning method derive from the continuous nature of the basis functions, one may consider assigning the basis arbitrarily. For example, we've already mentioned the possiblity of using orthogonal polynomials. Another options is to define the basis using random smooth functions, such as those parameterized by a shallow neural network \citep{schmidt1992feed,rahimi2007random,rahimi2008uniform,gorban2016approximation} -- a method referred to as random projection. These functions can naturally impose smoothness as inductive biases and can remain resolution-independent during the reconstruction and projection steps. 

\begin{figure}[!ht]
  \centering
  \begin{subfigure}[b]{0.3\textwidth}
    \includegraphics[width=\textwidth]{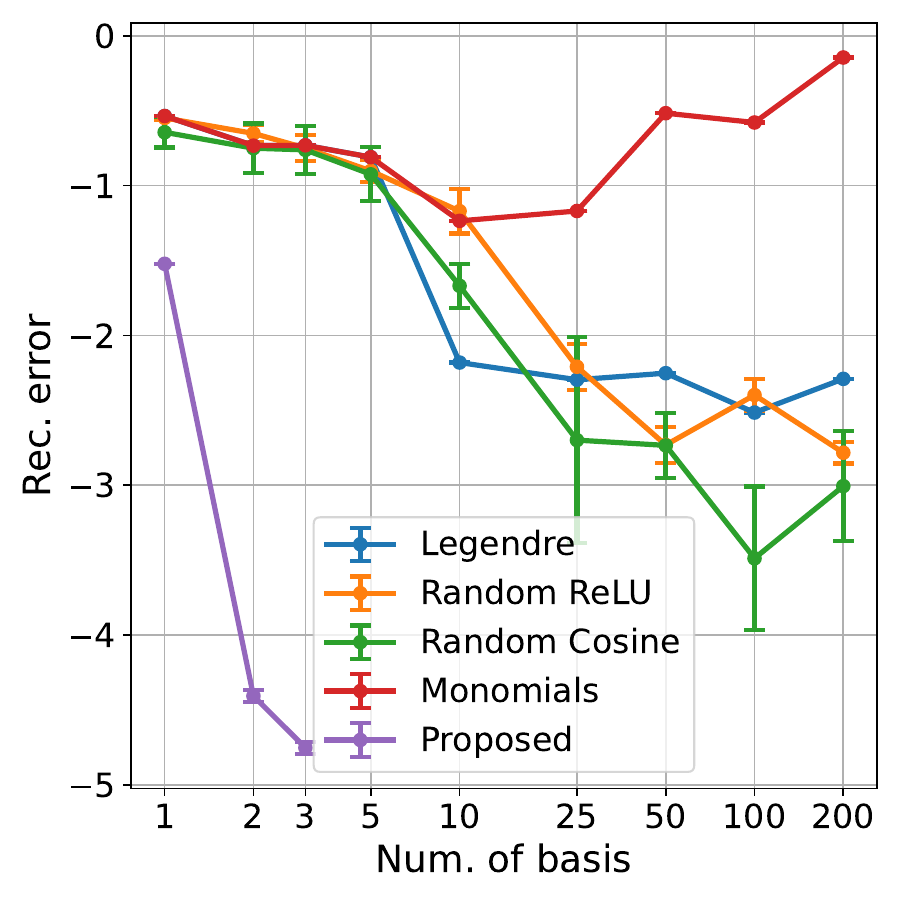}% 75 means unmasked
    \caption{25\% Masked}
    %\label{subfig:one}
  \end{subfigure}
    \begin{subfigure}[b]{0.3\textwidth}
    \includegraphics[width=\textwidth]{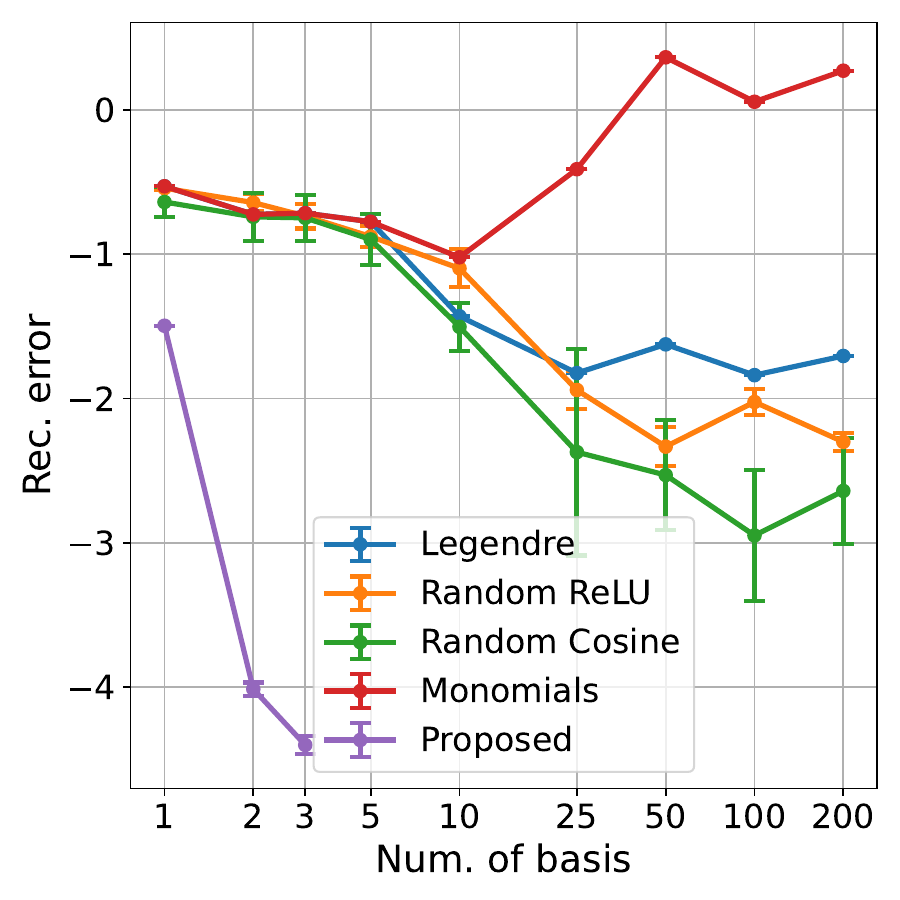}
    \caption{50\% Masked}
    %\label{subfig:one}
  \end{subfigure}
    \begin{subfigure}[b]{0.3\textwidth}
    \includegraphics[width=\textwidth]{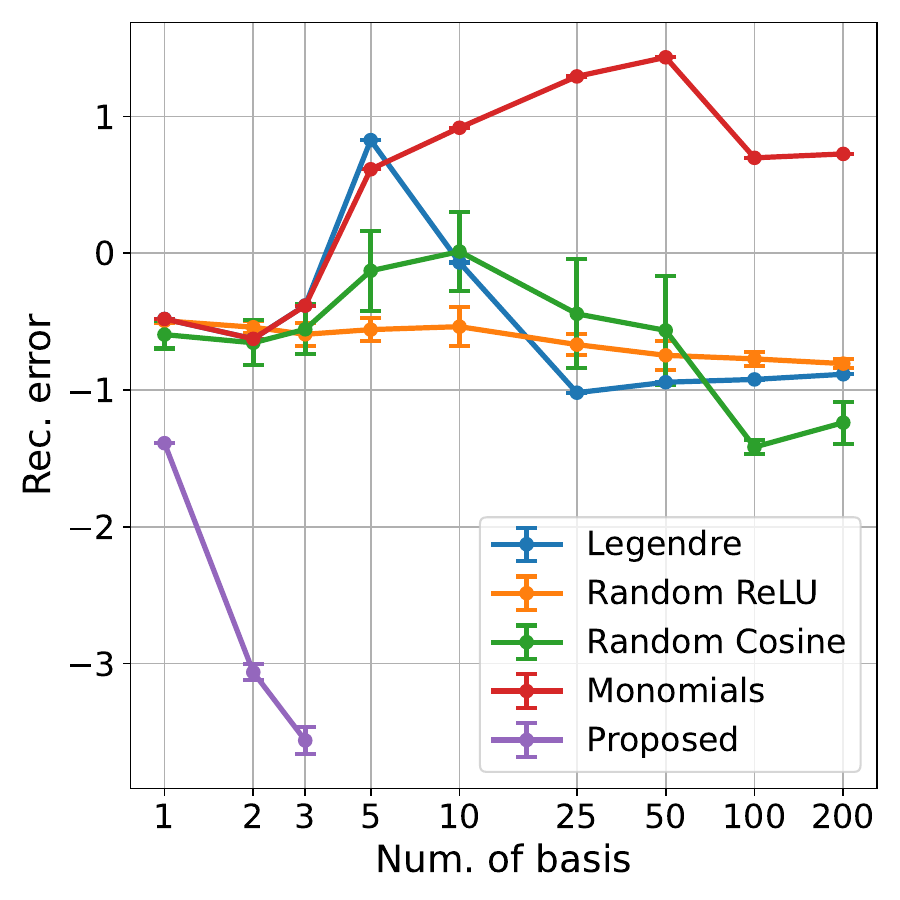}
    \caption{85\% Masked}
    %\label{subfig:one}
  \end{subfigure}
  \caption{Comparison of reconstruction accuracy on test data using four distinct sets of basis functions: Random projections with shallow ReLU networks  (orange), Random projections with cosine functions (green), Monomials (red), Legendre Polynomials (blue), and the proposed dictionary learning (purple). Each set is evaluated for five independent trials with randomly chosen parameters for stochastic methods. For deterministic basis functions (monomial and Legendre), no randomness is involved and a single trial is performed. As the results demonstrate, even as data sparsity increases, the proposed dictionary learning method consistently outperforms alternatives. Notably, dictionary learning achieves lower reconstruction error with significantly fewer basis functions, needing only 3 basis functions compared to hundreds in other methods.}
  \label{fig:abl-rand-basis}
\end{figure}

Here we aim to benchmark and compare the performance of the proposed dictionary learning method to different methods that employ arbitrary pre-defined continuous basis functions using the pedagogical example described above (Eq.~\eqref{eq:gen-pod}). To this end, we consider four cases: 1) random shallow ReLU networks, where each basis is a ReLU network with 20 hidden neurons randomly initialized; 2) classical cosine basis functions with random frequency and phase shifts as follows:
\begin{equation}
    \psi_i(x) = \cos(w_i x + b_i); w_i\sim \mathcal{N}(0, \sigma=2\pi), b_i \sim \text{Uniform}[0, 2\pi],
\end{equation}
for the $i$-th cosine basis functions; 3) the classical monomial basis functions; and 4) Legendre polynomial basis functions (deterministic). In Figure~\ref{fig:abl-rand-basis}, we compare the reconstruction accuracy using these different sets of basis functions for an increasing number of basis functions and different degrees of random data masking.

When the data density is sufficiently high, as shown in Figure~\ref{fig:abl-rand-basis}(a), the random ReLU, random cosine, and Legendre bases deliver satisfactory performance. However, they appear to require two orders of magnitude more basis functions than the proposed dictionary learning method, which presents challenges for downstream learning tasks, as operations in embedding space of dimension 3 are much simpler than operations in an embedding space of 100 or more dimensions. This is due to both computational efficiency and the fact that the amount of data required for such learning tasks may scale exponentially with dimensionality.
As data sparsity increases, as shown in Figure~\ref{fig:abl-rand-basis}(b and c), the performance gap between the proposed dictionary learning method and others grows more pronounced. For instance, with 85\% of the data masked in Figure~\ref{fig:abl-rand-basis}(c), the performance of the other methods becomes unacceptable. This decline is due to the limited data, making these other methods prone to overfitting.%, leading to good performance at interpolation but poor performance at extrapolation.
%RINO on test: 1.356e-03 1.807e-04
%Best Cosine (100) on test: 3.825e-02, 4.520e-03
%Best Relu (200) on test: 1.560e-01, 1.119e-02

In all cases, monomial basis functions did not perform well, regardless of the number of chosen basis elements or the amount of data available. This may be due to the fact that monomial bases of type $\psi_n(x) = x^n$ lack the appropriate inductive biases needed to effectively learn the data, particularly when the data exhibits sinusoidal behavior. This observation underscores the importance of selecting a sufficiently flexible set of basis functions, which is often not obvious at first. Therefore, an approach that can automatically identify the right set of bases in a data-driven manner is crucial, and this is one of the key contributions of this paper.

\begin{figure}[!ht]
  \centering
  \begin{subfigure}[b]{0.7\textwidth}
    \includegraphics[width=\textwidth]{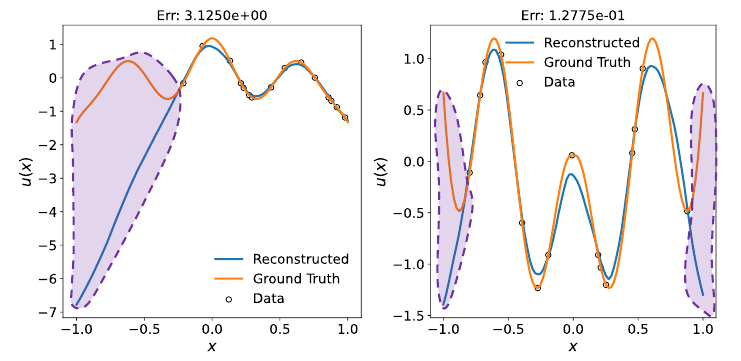}% 75 means unmasked
    \caption{Random Projection via ReLU Basis function}
    %\label{subfig:one}
  \end{subfigure}
    \begin{subfigure}[b]{0.7\textwidth}
    \includegraphics[width=\textwidth]{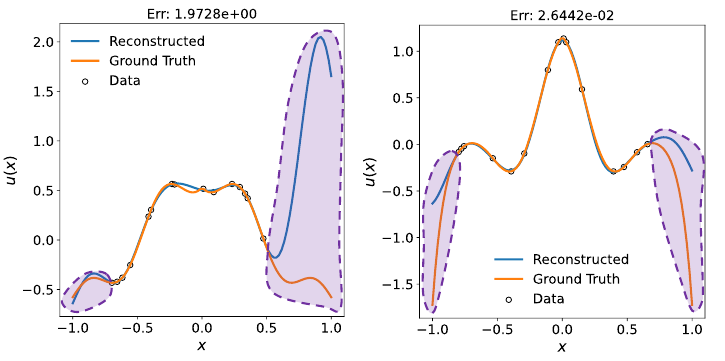}
    \caption{Random Projection via Cosine Basis Functions}
    %\label{subfig:one}
  \end{subfigure}
  \caption{Reconstruction curves for the worst case (left pane) and the 25th quantile (right pane) in terms of relative reconstruction error in the case where 85\% of the data was masked.}
  \label{fig:abl-rand-proj-rect}
\end{figure}

To further analyze the performance of random bases in the case where the data has the highest sparsity (i.e., 85\% masked data), we plotted the reconstruction curves for the worst case and the 25th quintile in terms of relative errors in Figure \ref{fig:abl-rand-proj-rect}. As expected (and as consistent with the literature \cite{rahimi2007random,gorban2016approximation}), the random basis functions perform well in the interpolation regime but poorly in the extrapolation regime beyond the data support (as observed in the purple-shaded areas in these figures). Moreover, the underlying generative process of the data is symmetric at $x=0$, see Figure \ref{fig:gpod-data}(a). However, there is no guarantee that this symmetry can be recovered using random basis functions. As already discussed, these issues do not arise in the proposed dictionary learning method, as we learn appropriate basis functions by considering the common correlation structure across all realizations, rather than merely interpolating for each individual realization.

\begin{figure}[!ht]
  \centering
    \begin{subfigure}[b]{0.7\textwidth}
    \includegraphics[width=\textwidth]{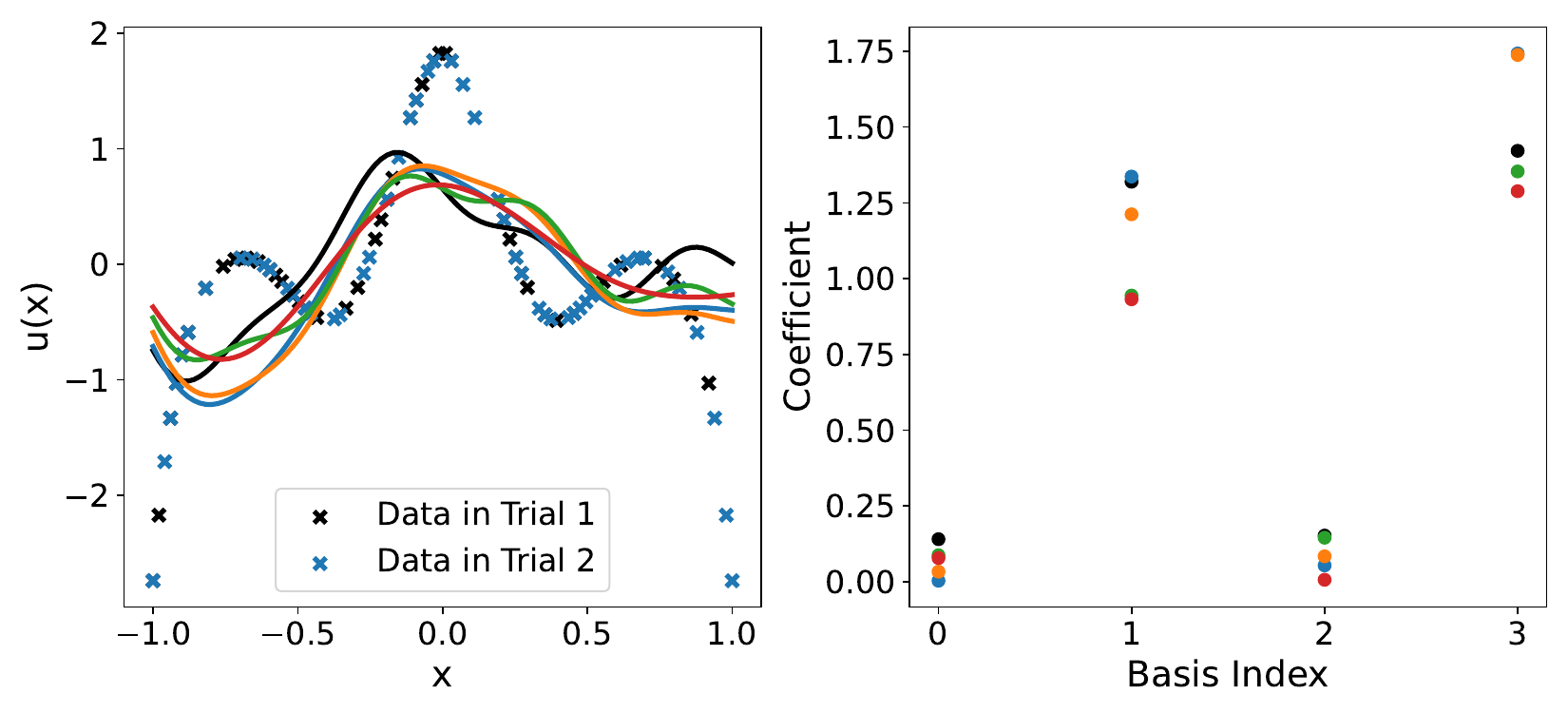}
    \caption{Random Basis}
    %\label{subfig:one}
  \end{subfigure}
    \begin{subfigure}[b]{0.7\textwidth}
    \includegraphics[width=\textwidth]{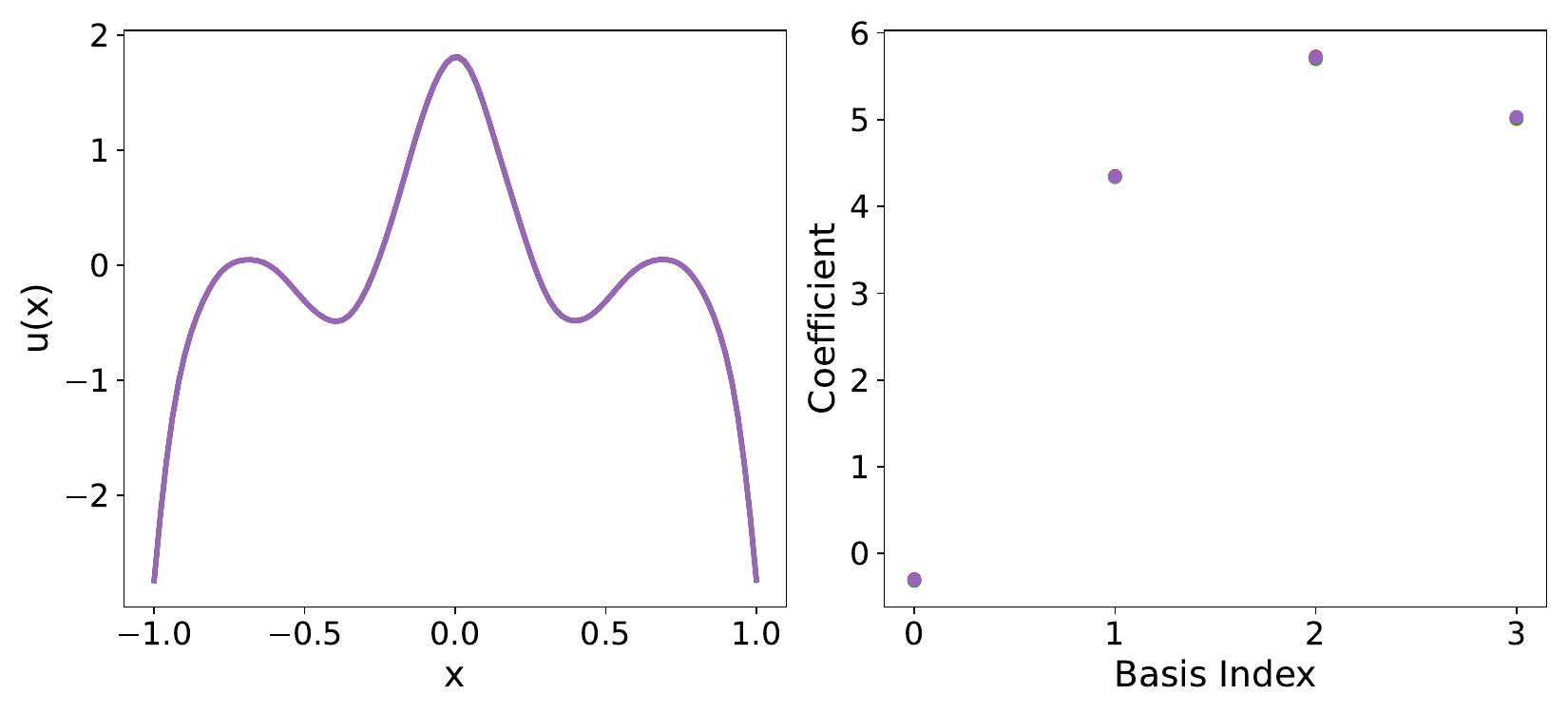}
    \caption{Optimal Basis}
    %\label{subfig:one}
  \end{subfigure}
  \caption{Consistency check of the embeddings obtained from projections via (a) random basis and (b) optimal basis sets. The right column display the coefficients corresponding to each basis function, while the left column show the reconstructed full fields using the identified coefficients. Different colors represent distinct training datasets, where the number of data points remains the same but their locations are chosen randomly and differ from one dataset to another. For illustration purposes, the data from the first two of the five trials is shown in the top-left panel using black and blue ``x'' symbols, with their corresponding reconstructions represented by black and blue solid lines, respectively.
  }
  \label{fig:abl-embd-consistcy-rand-proj}
\end{figure}

\paragraph{Spanning versus Parametrization}
As we have shown, arbitrary \textbf{linear} (with respect to the basis functions) projection and reconstruction using linearly independent random basis functions often requires exponentially more basis functions than projections using an optimal orthogonal basis set.
%Although this is aligned with the linear theory \cite{gorban2016approximation}, it contrasts with findings from manifold learning via random projections, which suggest that a maximum of $2n+1$ random basis functions may suffice to recover the manifold [NEEDS YANNIS ????], where $n$ is the intrinsic dimensionality of the data manifold; in this case, $n= 3$. In fact, one can recover the data manifold using a maximum of $2n+1$ random basis functions, but not by spanning the space. Instead, this can be achieved via nonlinear parameterization, which maps the projected solution on the hyperplane back to the target manifold.
Although this is aligned with the linear theory \cite{gorban2016approximation}, it does not take advantage of the fact that, with established embedding theorems, one can {\em embed} an $n$-dimensional manifold with at most $2n+1$ generic observables \citep{whitney1936differentiable,whitney1944self} (in the present case, the data components result in $2n+1$ random basis functions). 
These projections do not {\em span} the intrinsically $n$-dimesional data; yet, the projections of the data in any additional random basis functions can be obtained as (potentially nonlinear) {\em functions} of the first $2n+1$ random basis functions.
We say that the $2n+1$ basis functions {\em parametrize} the data, but do not {\em span} them. The approximation of the data can now be achieved via nonlinear parameterization, which maps the projected solution on the hyperplane back to the target manifold. 
This approach shares similarities with the post-processing Galerkin method \citep{koronaki2024nonlinear,garcia1998postprocessing}. However, such nonlinear parameterization may have some caveats: (1) it is less interpretable than the proposed spanning scheme; (2) the number of basis functions is not known a \textit{priori}—only an upper bound is provided—while the proposed approach adaptively identifies the number of basis functions according to a tolerance; (3) The embedding space constructed via random projection is highly sensitive to sampling, which undermines the uniqueness of the signal representation. To elaborate further on the last point, we conducted a simple test as follows. We pick one signal realization from the test data set and aim to reconstruct the signal with three linearly independent random cosine basis functions, as mentioned earlier. To assess the sensitivity of the embeddings, we consider five cases where half of the points are masked in each case. However, the locations of these masked points are randomly selected and vary from one case to another. The embeddings obtained from random projection, shown by the coefficients in the right pane of Figure~\ref{fig:abl-embd-consistcy-rand-proj}(a), are clearly dependent on the location of the samples. The left column shows the reconstructed signal (solid lines with different colors), where we observe high variance due to this sample location dependence. In contrast, the projection via optimal orthogonal basis functions, as shown in Figure~\ref{fig:abl-embd-consistcy-rand-proj}(b),is consistent across cases and independent of the sample locations.

%However, this direction requires a separate study and goes beyond the scope of the current work.

\section{Ablation Study: Comparison between RI-DeepONet and RINO}
\label{appx:ri-vs-ri}
Here, we aim to demonstrate that while RI-DeepONet or DeepONet do not necessarily identify the most compact or orthogonal set of basis functions to span the output field, RINO effectively achieves this. The example used here is the antiderivative dataset introduced by Lu Lu et. al \cite{lu2021learning}.

Since the focus is on the impact of output basis functions on prediction quality, no data is masked in this study. Instead, full, clean data is utilized across all cases. The same input embedding is employed for all scenarios, determined through the proposed dictionary learning approach applied to the input function data.

\begin{figure}[!ht]
  \centering
  \begin{subfigure}[b]{0.4\textwidth}
    \includegraphics[width=\textwidth]{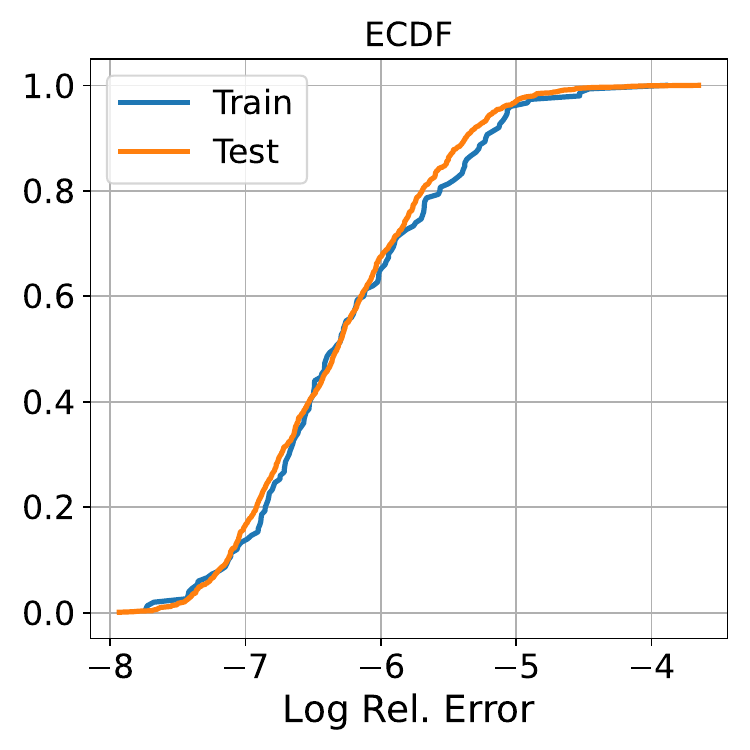}%
    % \caption{x}
    %\label{subfig:one}
  \end{subfigure}
    \begin{subfigure}[b]{0.4\textwidth}
    \includegraphics[width=\textwidth]{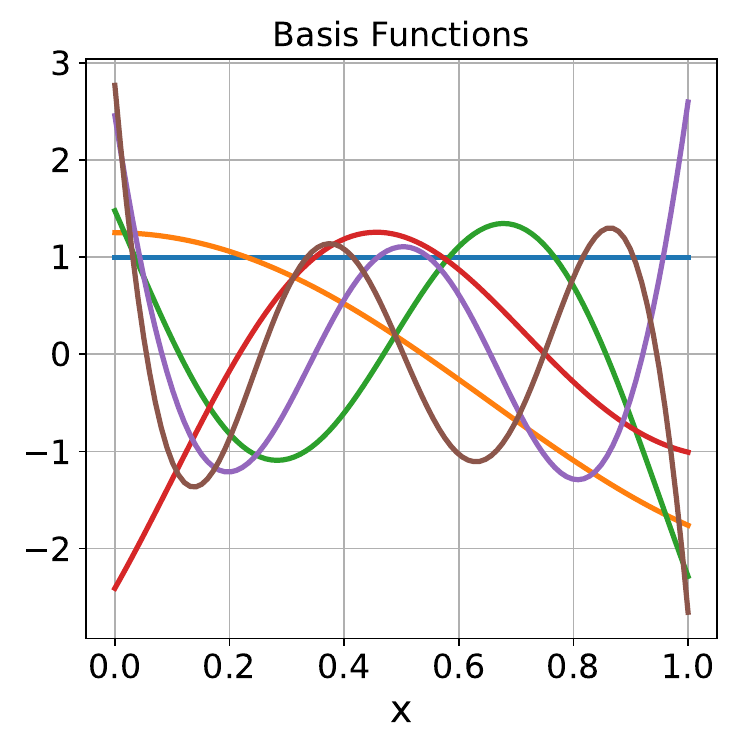}
    % \caption{x}
    %\label{subfig:one}
  \end{subfigure}
  \caption{RINO ablation: (left) ecdf plots for the learned operator, (right) the learned output basis functions.}
  \label{fig:abl2-rino}
\end{figure}

RINO identifies that only five output basis functions are sufficient to span the output functions effectively. Each of these basis functions is implemented as a SIREN with a single hidden layer, 10 hidden units, and a frequency parameter of $w_0=5$.A key advantage of RINO is its ability not only to identify these basis functions but also to determine the number required—an unknown hyperparameter in both DeepONet and RI-DeepONet. In this case, a linear transformation between the input and output embeddings proved sufficient and was determined analytically. The strong generalization capability of RINO is evident from the empirical cumulative distribution functions (ECDFs) of training and test relative errors, plotted on a logarithmic scale in Figure~\ref{fig:abl2-rino}.

In the case of RI-DeepONet experiments, the branch network is a vanilla MLP with a single hidden layer of size 50 and an \texttt{ELU} activation function. We consider two types of trunk networks: one is a vanilla MLP with a single hidden layer of size 50, and the other is a SIREN with a single hidden layer of size 50 and a frequency parameter of 5. Since RINO has already established that the output function space can be effectively spanned by only five basis functions, we test three cases where the sizes of the last layers of the branch and trunk networks are set to 5, 10, and 20.

\begin{figure}[!ht]
  \centering
  \begin{subfigure}[b]{0.32\textwidth}
    \includegraphics[width=\textwidth]{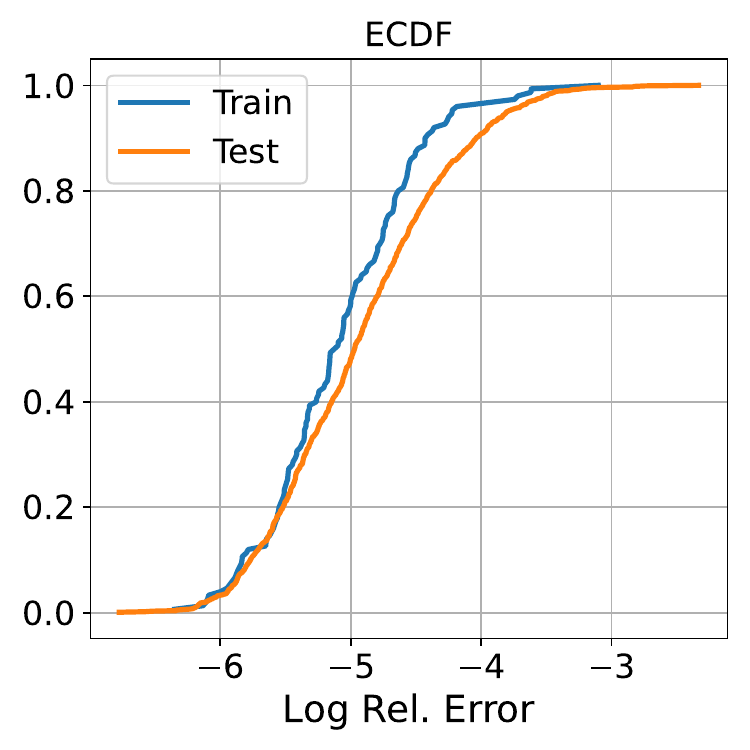}%
    \caption{Vanilla MLP 5}
    %\label{subfig:one}
  \end{subfigure}
    \begin{subfigure}[b]{0.32\textwidth}
    \includegraphics[width=\textwidth]{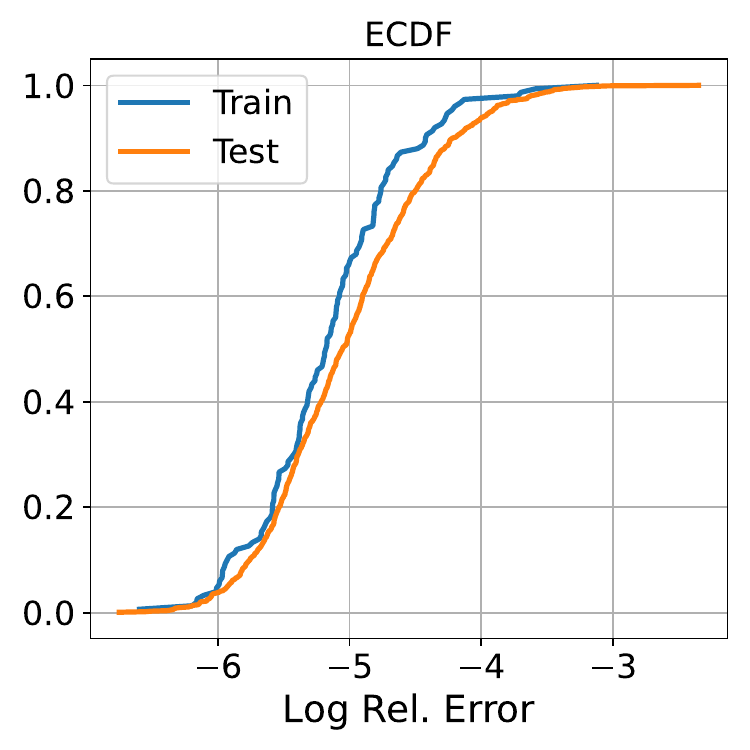}
    \caption{Vanilla MLP 10}
    %\label{subfig:one}
  \end{subfigure}
    \begin{subfigure}[b]{0.32\textwidth}
    \includegraphics[width=\textwidth]{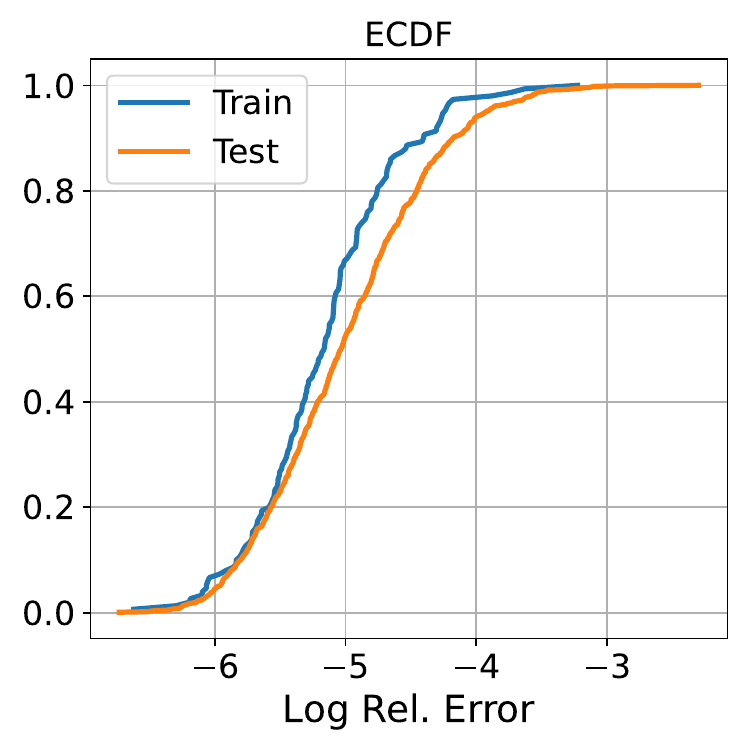}
    \caption{Vanilla MLP 20}
    %\label{subfig:one}
  \end{subfigure}
  %%%%%%%%%%%%%%%%%%%%%%%%%%%%%%%%%
  \begin{subfigure}[b]{0.32\textwidth}
    \includegraphics[width=\textwidth]{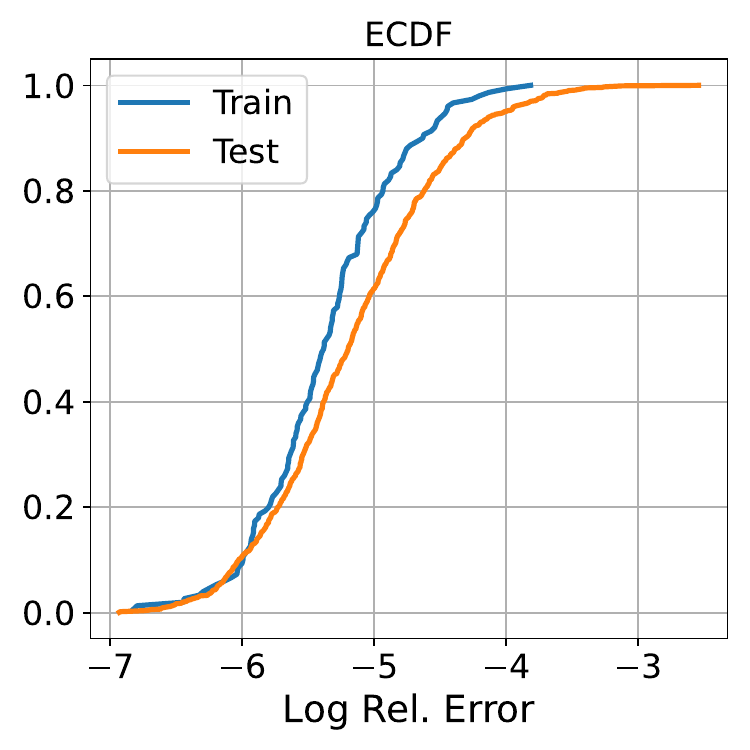}%
    \caption{SIREN 5}
    %\label{subfig:one}
  \end{subfigure}
    \begin{subfigure}[b]{0.32\textwidth}
    \includegraphics[width=\textwidth]{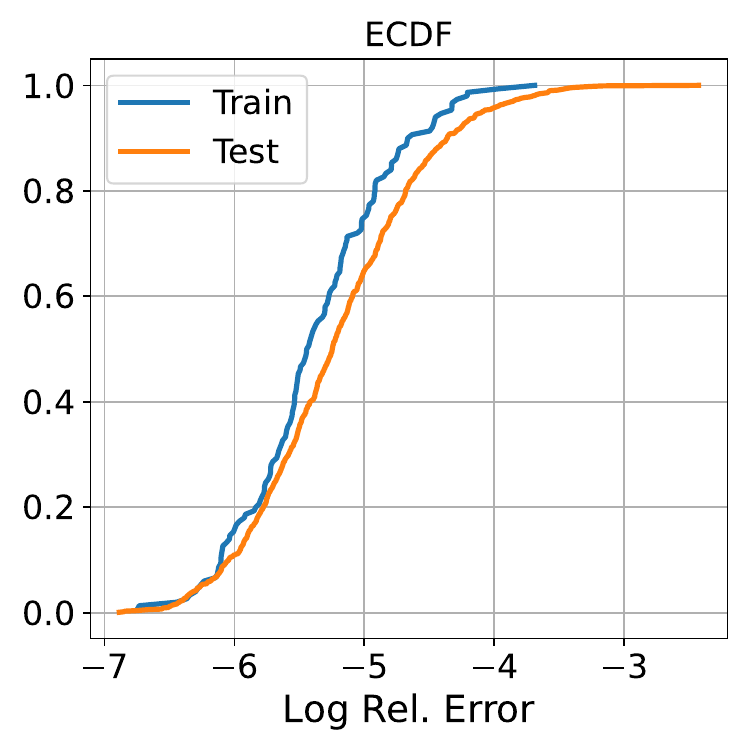}
    \caption{SIREN 10}
    %\label{subfig:one}
  \end{subfigure}
    \begin{subfigure}[b]{0.32\textwidth}
    \includegraphics[width=\textwidth]{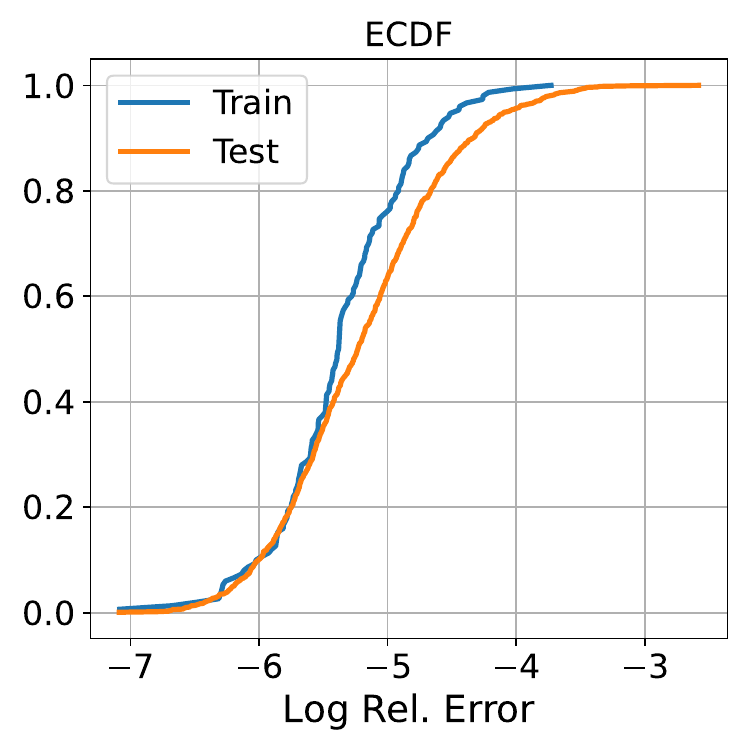}
    \caption{SIREN 20}
    %\label{subfig:one}
  \end{subfigure}
  \caption{RI-DeepONet ablation: ECDF plots for six different architectures. The branch net remains the same across all cases, while the trunk net differs in the number of hidden units (i.e., the number of output basis functions) and its architecture. The top row uses an MLP as the trunk net, and the bottom row uses a SIREN. Numbers 5, 10, and 20 indicate the number of hidden units in the trunk net.}
  \label{fig:abl2-ri-don-ecdf}
\end{figure}

The ECDF plots for these RI-DeepONet six cases are presented in Figure~\ref{fig:abl2-ri-don-ecdf}. 
Comparing these ECDFs with the RINO ECDF, one can observe two key points. First, the generalization gap in RINO is consistently much tighter than in RI-DeepONet, as evidenced by the closer alignment of training and test ECDFs in RINO. Second, the worst-case test prediction error in RINO is consistently better than in RI-DeepONet. We hypothesize that this improvement can be attributed to the superior quality of the output basis functions learned by RINO. These basis functions align closely with the data behavior and, importantly, are the most compact, containing no unnecessary information due to their orthogonality. A similar generalization effect, attributed to identifying the correct basis functions, was also observed in Appendix~\ref{appx:gpod}, though from the perspective of input functions.

\begin{figure}[!ht]
  \centering
  \begin{subfigure}[b]{0.32\textwidth}
    \includegraphics[width=\textwidth]{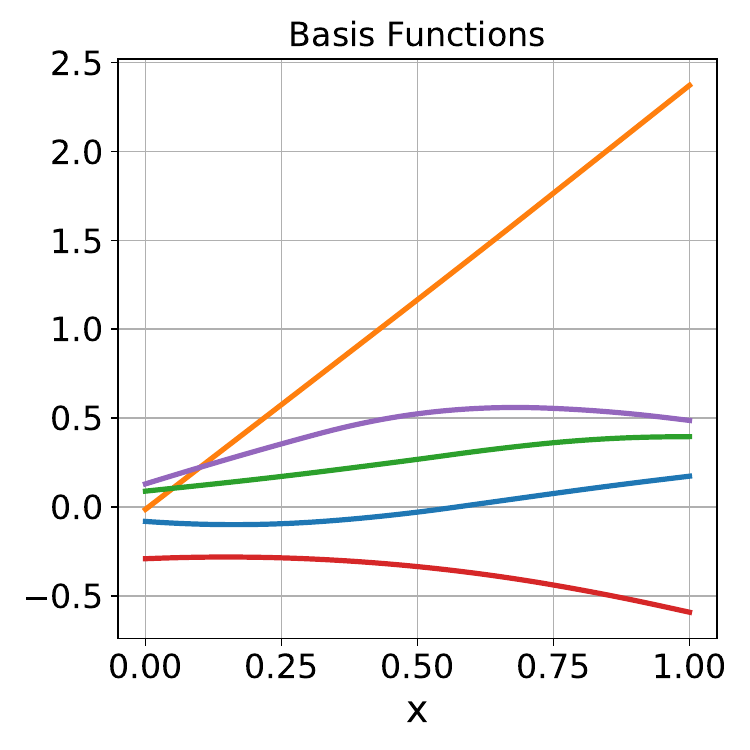}%
    \caption{Vanilla MLP 5}
    %\label{subfig:one}
  \end{subfigure}
    \begin{subfigure}[b]{0.32\textwidth}
    \includegraphics[width=\textwidth]{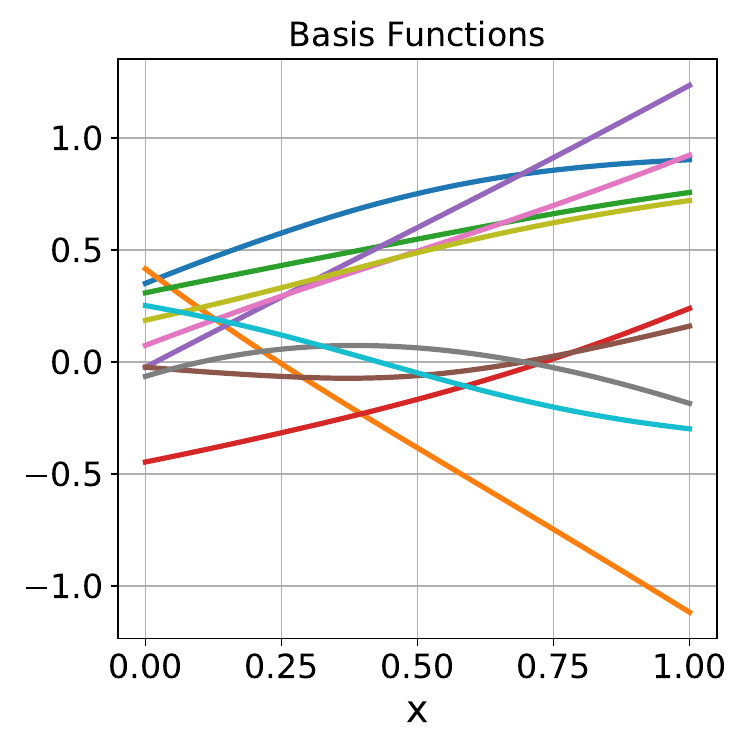}
    \caption{Vanilla MLP 10}
    %\label{subfig:one}
  \end{subfigure}
    \begin{subfigure}[b]{0.32\textwidth}
    \includegraphics[width=\textwidth]{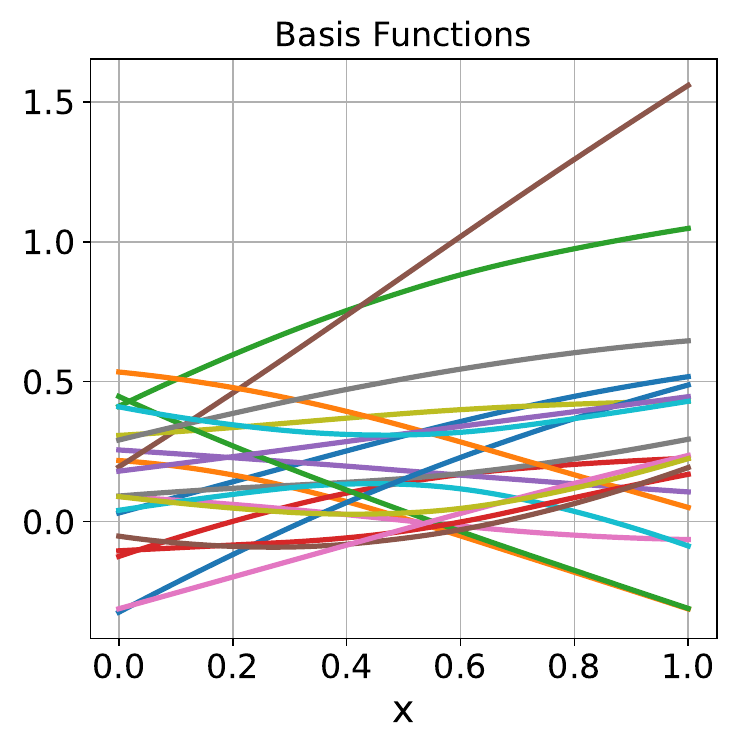}
    \caption{Vanilla MLP 20}
    %\label{subfig:one}
  \end{subfigure}
  %%%%%%%%%%%%%%%%%%%%%%%%%%%%%%%%%
  \begin{subfigure}[b]{0.32\textwidth}
    \includegraphics[width=\textwidth]{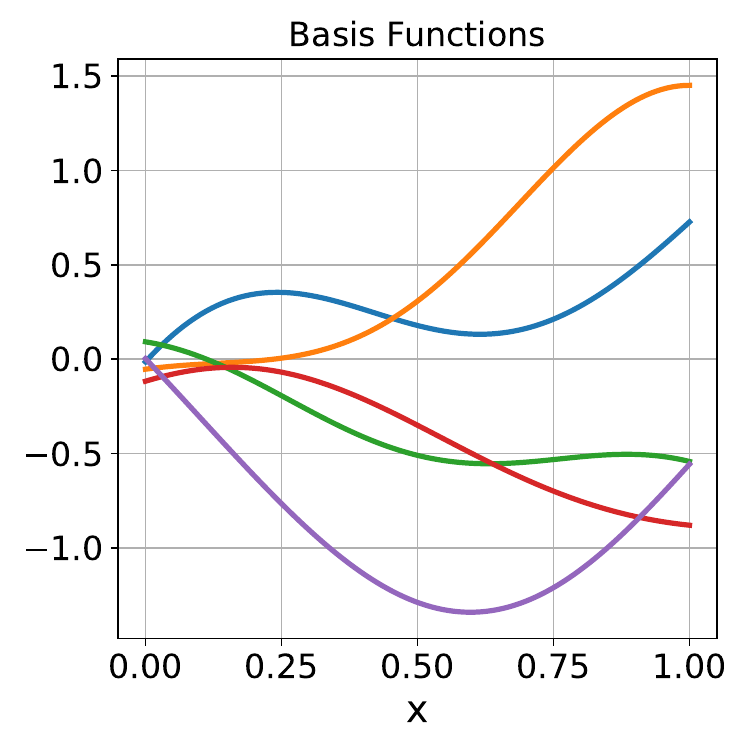}%
    \caption{SIREN 5}
    %\label{subfig:one}
  \end{subfigure}
    \begin{subfigure}[b]{0.32\textwidth}
    \includegraphics[width=\textwidth]{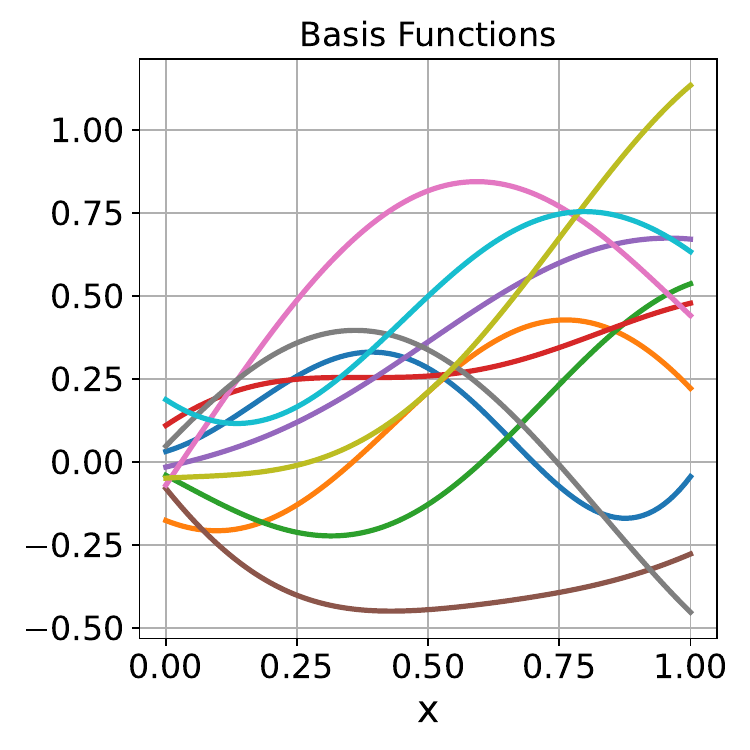}
    \caption{SIREN 10}
    %\label{subfig:one}
  \end{subfigure}
    \begin{subfigure}[b]{0.32\textwidth}
    \includegraphics[width=\textwidth]{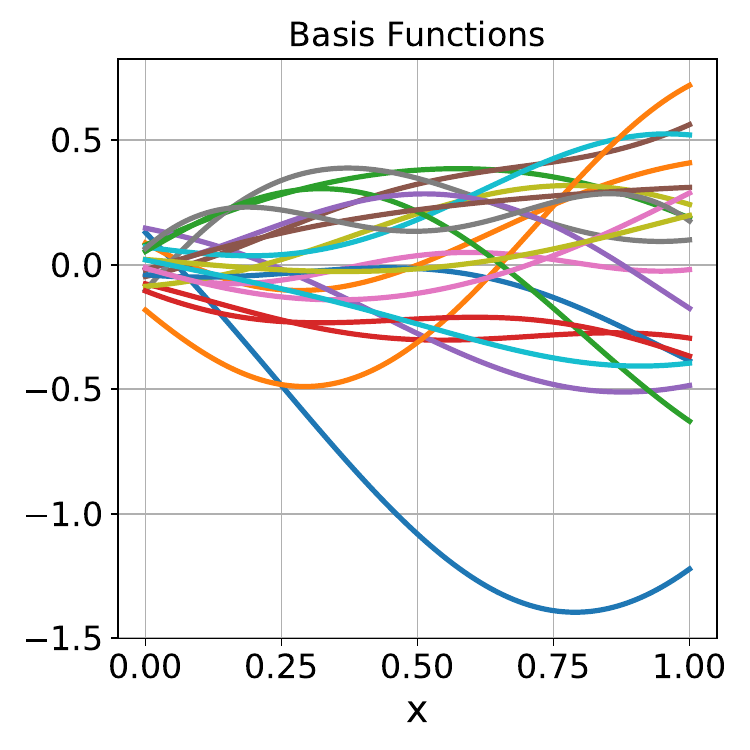}
    \caption{SIREN 20}
    %\label{subfig:one}
  \end{subfigure}
  \caption{RI-DeepONet ablation: learned trunk basis functions for six different architectures. The branch net remains the same across all cases, while the trunk net differs in the number of hidden units (i.e., the number of output basis functions) and its architecture. The top row uses an MLP as the trunk net, and the bottom row uses a SIREN. Numbers 5, 10, and 20 indicate the number of hidden units in the trunk net.}
  \label{fig:abl2-ri-don-basis}
\end{figure}

To further examine the aforementioned hypothesis, we plot the trunk basis functions learned by RI-DeepONet in Figure~\ref{fig:abl2-ri-don-basis}. It is evident from these plots that the learned basis functions do not guarantee compactness or orthogonality. As a result, some basis functions may exhibit linear dependence or high correlation, introducing redundancy in the approximation. This redundancy reduces the efficiency of the representation and may lead to overfitting, as the model focuses on memorizing training data instead of accurately capturing the underlying data structure.

\begin{figure}[!ht]
  \centering
  \begin{subfigure}[b]{0.32\textwidth}
    \includegraphics[width=\textwidth]{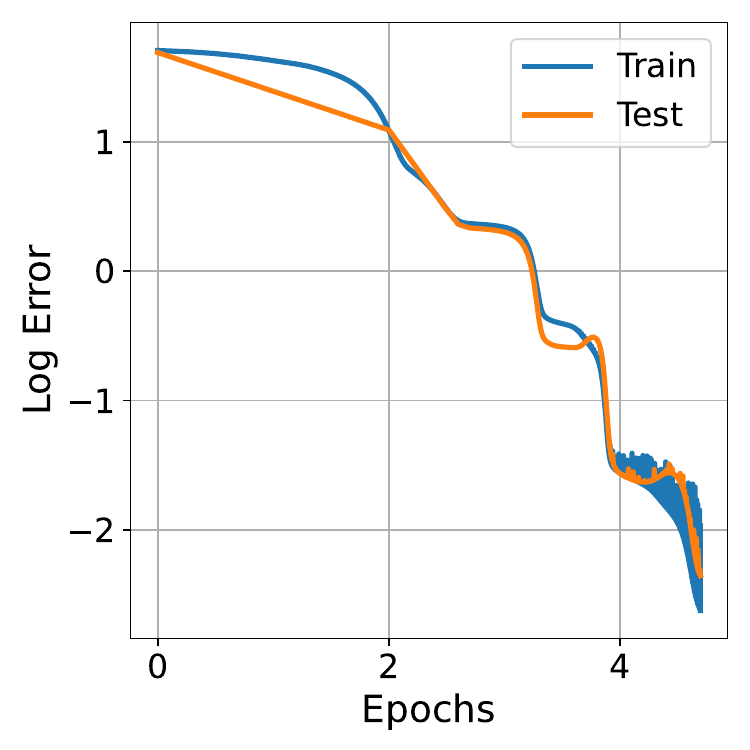}%
    \caption{Vanilla MLP 5}
    %\label{subfig:one}
  \end{subfigure}
    \begin{subfigure}[b]{0.32\textwidth}
    \includegraphics[width=\textwidth]{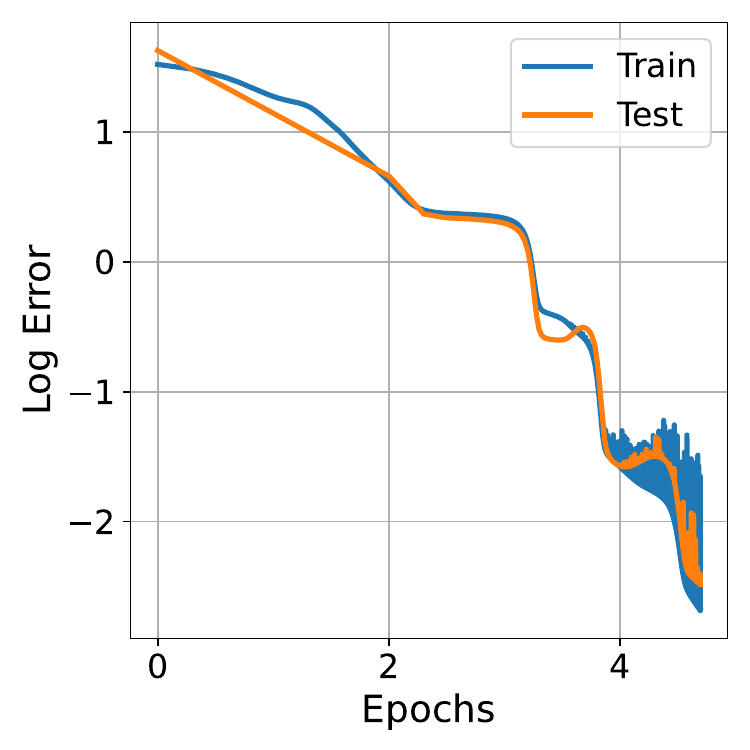}
    \caption{Vanilla MLP 10}
    %\label{subfig:one}
  \end{subfigure}
    \begin{subfigure}[b]{0.32\textwidth}
    \includegraphics[width=\textwidth]{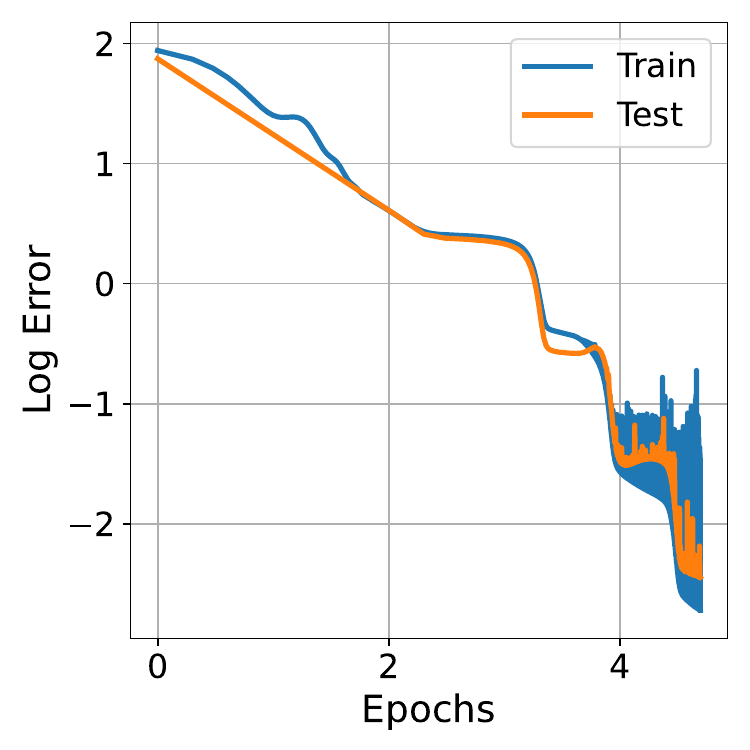}
    \caption{Vanilla MLP 20}
    %\label{subfig:one}
  \end{subfigure}
  %%%%%%%%%%%%%%%%%%%%%%%%%%%%%%%%%
  \begin{subfigure}[b]{0.32\textwidth}
    \includegraphics[width=\textwidth]{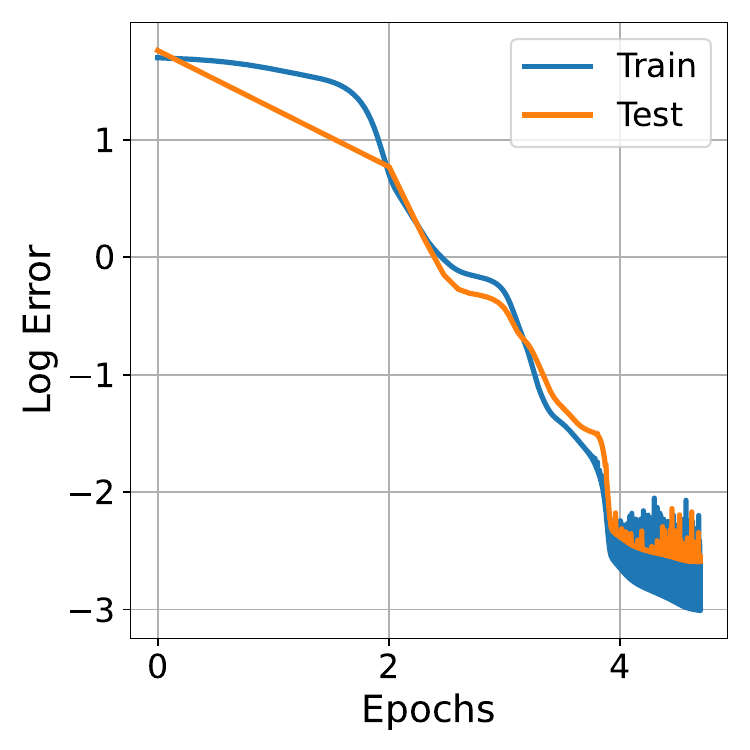}%
    \caption{SIREN 5}
    %\label{subfig:one}
  \end{subfigure}
    \begin{subfigure}[b]{0.32\textwidth}
    \includegraphics[width=\textwidth]{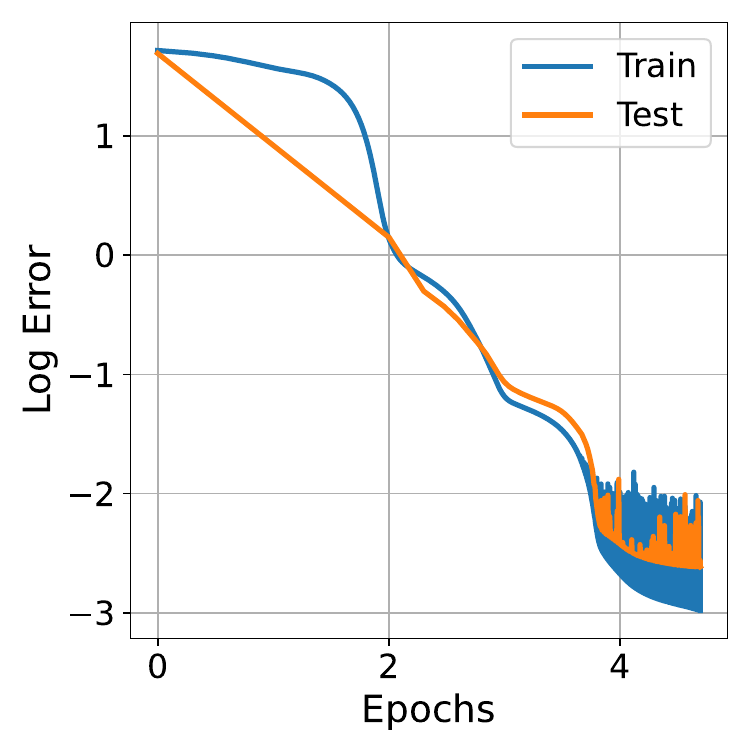}
    \caption{SIREN 10}
    %\label{subfig:one}
  \end{subfigure}
    \begin{subfigure}[b]{0.32\textwidth}
    \includegraphics[width=\textwidth]{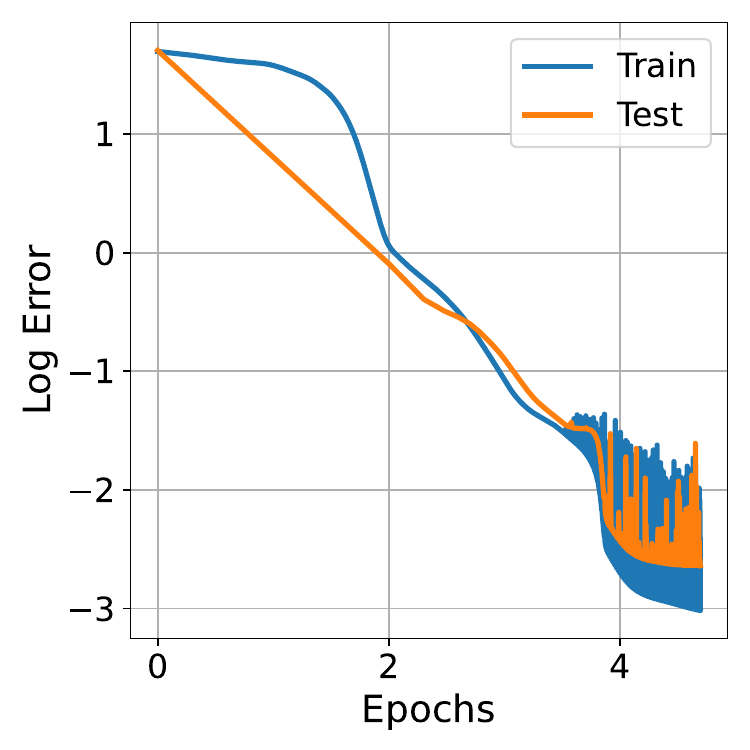}
    \caption{SIREN 20}
    %\label{subfig:one}
  \end{subfigure}
  \caption{RI-DeepONet ablation: operator learning loss functions for six different architectures. The branch net remains the same across all cases, while the trunk net differs in the number of hidden units (i.e., the number of output basis functions) and its architecture. The top row uses an MLP as the trunk net, and the bottom row uses a SIREN. Numbers 5, 10, and 20 indicate the number of hidden units in the trunk net.}
  \label{fig:abl2-ri-don-loss}
\end{figure}

The operator learning loss functions for RI-DeepONet are shown in Figure~\ref{fig:abl2-ri-don-loss} to ensure that the training has not been extended unnecessarily, which could inadvertently lead to overfitting. As the results indicate, the training process was healthy, with no significant evidence of overfitting observed in the training data. All these observations support our hypothesis that the structure of the learned basis functions may influence accuracy, robustness, generalization, and interpretability. Consequently, algorithms that ensure the learning of well-structured basis functions with such guarantees are desirable.

\section{Random Field Generation}
\label{appx:RFG}
Similar to the strategy utilized in \cite{lu2021learning,wang2021learning}, the random but correlated input functions $u(\boldsymbol{x})$ for data generation are sampled from a Gaussian random process $u(\boldsymbol{x}) \sim \mathcal{GP}(0, \text{Cov}(\boldsymbol{x}_1, \boldsymbol{x}_2; l))$ with zero mean and a covariance function defined the radial basis function (RBF) with an isotropic kernel as follows:
\begin{equation}
    \text{Cov}(\boldsymbol{x}_1, \boldsymbol{x}_2; l) = \exp
    \left(
        -\frac{\|\boldsymbol{x}_1 - \boldsymbol{x}_2\|_{\mathcal{X}}^2}{2l^2}
    \right),
\end{equation}
where $\boldsymbol{x}_1, \boldsymbol{x}_2\in\mathcal{X}$ and $\|\boldsymbol{x}_1 - \boldsymbol{x}_2\|_{\mathcal{X}}$ induces a norm associated with the domain $\mathcal{X}$. Here, in the 1D and 2D cases, the chosen norm is the $L_2$-norm.

\end{document}